\documentclass{article}

\usepackage{times}
\usepackage[pdftex]{graphicx}
\usepackage{amsmath}
\usepackage{amssymb}
\usepackage{authblk}

\usepackage{fmt}
\usepackage{mcr}
\usepackage{pkg}
\usepackage{thmE}
\usepackage{algE}
\usepackage[pagebackref=true,breaklinks=true,letterpaper=true,bookmarks=false]{hyperref}
\usepackage[margin=1in]{geometry}
 
\begin{document}

%%%%%%%%% TITLE
\title{Computing Valid $p$-values for Image Segmentation by Selective Inference}

%% omit the following lines up until the closing ``}''.
%% Additional authors and addresses can be added with ``\and'',
%% just like the second author.
%% To save space, use either the email address or home page, not both
%\and
\author[1~]{Kosuke Tanizaki}
\affil[1,2,4,5]{Nagoya Institute of Technology}

\author[2]{Noriaki Hashimoto}

\author[3]{Yu Inatsu}
\affil[3,5]{RIKEN Center for Advanced Intelligence Project}

\author[4]{Hidekata Hontani}

\author[5]{Ichiro Takeuchi}
\affil[5]{National Institute for Material Science\protect\\{\it tanizaki.k.mllab.nit@gmail.com\\~yu.inatsu@riken.jp\protect\\ \{hashimoto.noriaki,hontani,takeuchi.ichiro\}@nitech.ac.jp}}

\date{}
\maketitle
%\thispagestyle{empty}

%%%%%%%%% ABSTRACT
\begin{abstract}
Image segmentation is one of the most fundamental tasks of computer vision.
In many practical applications, it is essential to properly evaluate the reliability of individual segmentation results.
In this study, we propose a novel framework for determining the statistical significance of segmentation results in the form of $p$-values. 
Specifically, we utilize a statistical hypothesis test for determining the difference between the object region and the background region. 
This problem is challenging because the difference can be deceptively large (called segmentation bias) due to the adaptation of the segmentation algorithm to the data. 
To overcome this difficulty, we introduce a statistical approach called selective inference, and develop a framework for computing valid $p$-values in which segmentation bias is properly accounted for.
Although the proposed framework is potentially applicable to various segmentation algorithms, here we focus on graph-cut- and threshold-based segmentation algorithms, and develop two specific methods for computing valid $p$-values for the segmentation results obtained by these algorithms.
We prove the theoretical validity of these two methods and demonstrate their practicality by applying them to the segmentation of medical images.

\end{abstract}

%%%%%%%%%%%%%%%%%%%%%%%%%%%%%%%%%
\section{Introduction}
\label{sec:introduction}
Image segmentation is one of the most fundamental tasks in computer vision.
Many segmentation algorithms have been proposed and applied to various problems, such as the binarization of document images \cite{sauvola2000segtxt,papavassiliou2010segtxt} and the detection of abnormal regions in medical images~\cite{li2012segCT,li2015segCT}.
Segmentation algorithms are usually formulated as a problem of optimizing a certain loss function.
For example, in threshold (TH)-based segmentation algorithms~\cite{otsu1979global,white1983local}, the loss functions are defined based on similarity within a given region and dissimilarity between different regions.
In graph cut (GC)-based segmentation algorithms~\cite{boykov2001gc,boykov2006gc}, the loss functions are defined based on the similarity of adjacent pixels in a given region and dissimilarity of adjacent pixels at the boundaries.
Depending on the problem and the properties of the target images, an appropriate segmentation algorithm must be selected.

In many practical non-trivial applications, there may be the risk of obtaining incorrect segmentation results.
In practical problems in which segmentation results are used for decision-making or as a component of a larger system, it is essential to properly evaluate their reliability.
For example, when segmentation results are used in a computer-aided diagnosis system, it should be possible to appropriately quantify the risk of the obtained individual segmentation result being incorrect.
Although the expected proportion of the overall false positive findings can be evaluated, (e.g., by receiver operating characteristic curve analysis), it is difficult to quantitatively evaluate the reliability of individual segmentation results.

In this study, we propose a novel framework called Post-Segmentation Inference (PSegI) for determining the statistical significance of individual segmentation results in the form of $p$-values.
For simplicity, we focus only on segmentation problems in which an image is divided into an object region and a background region.
To quantify the reliability of individual segmentation results, we utilize a statistical hypothesis test for determining the difference in the average pixel intensities between the two regions (see \eq{eq:hypothesis_testing} in \S2).
If the difference is sufficiently large and the probability of observing such a large difference is sufficiently small in a null image (i.e., one that contains no specific objects), it indicates that the segmentation result is statistically significant.
The $p$-value of the statistical hypothesis test can be used as a quantitative reliability metric of individual segmentation results; i.e., if the $p$-value is sufficiently small, it implies that a segmentation result is reliable.

\begin{figure*}[t]
 \begin{center}
  \begin{minipage}[t]{0.20\hsize}
   \includegraphics[width = \linewidth]{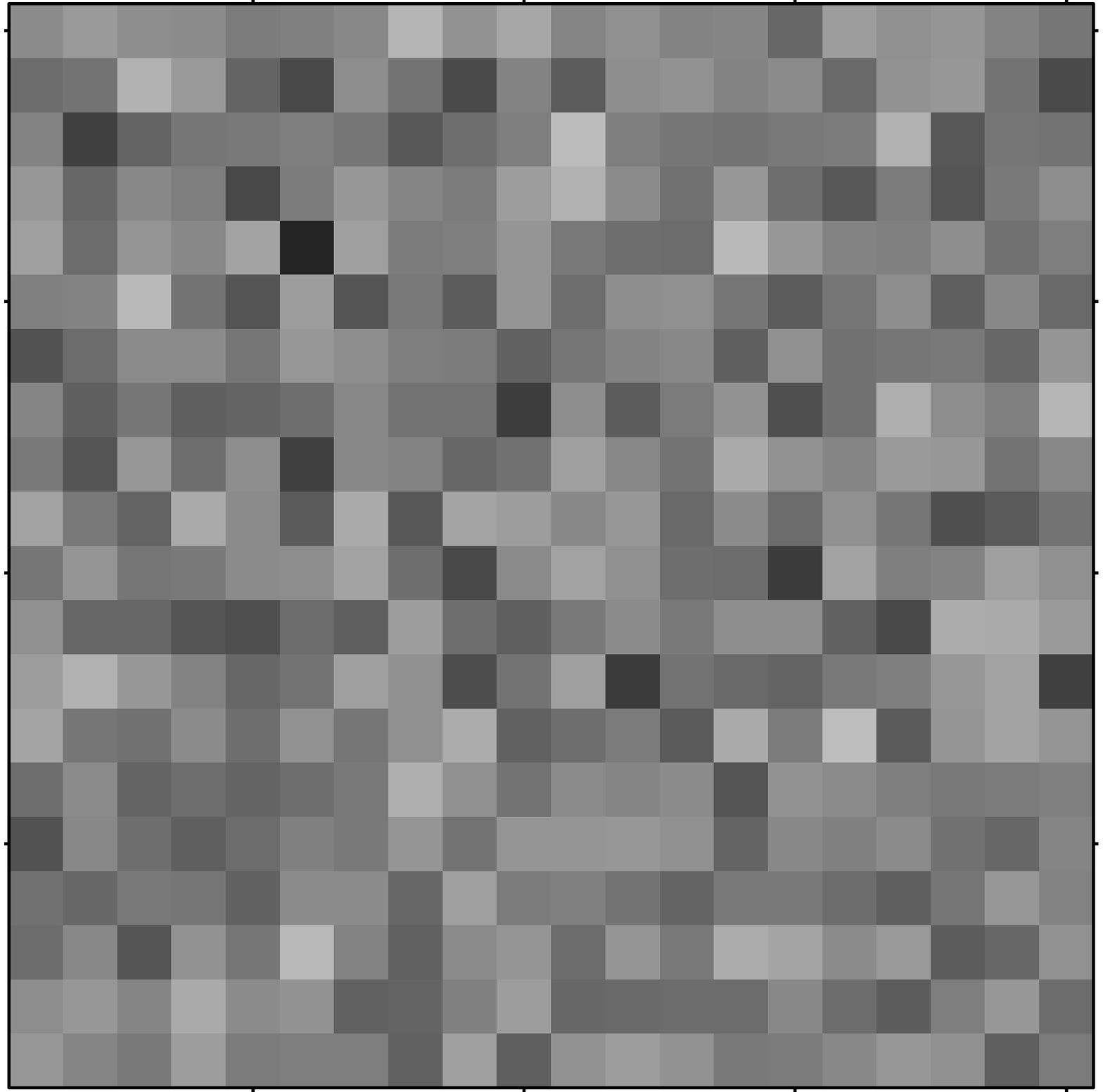}
   \subcaption{Original image} \label{fig: image}
  \end{minipage}
  \begin{minipage}[t]{0.20\hsize}
   \includegraphics[width =\linewidth]{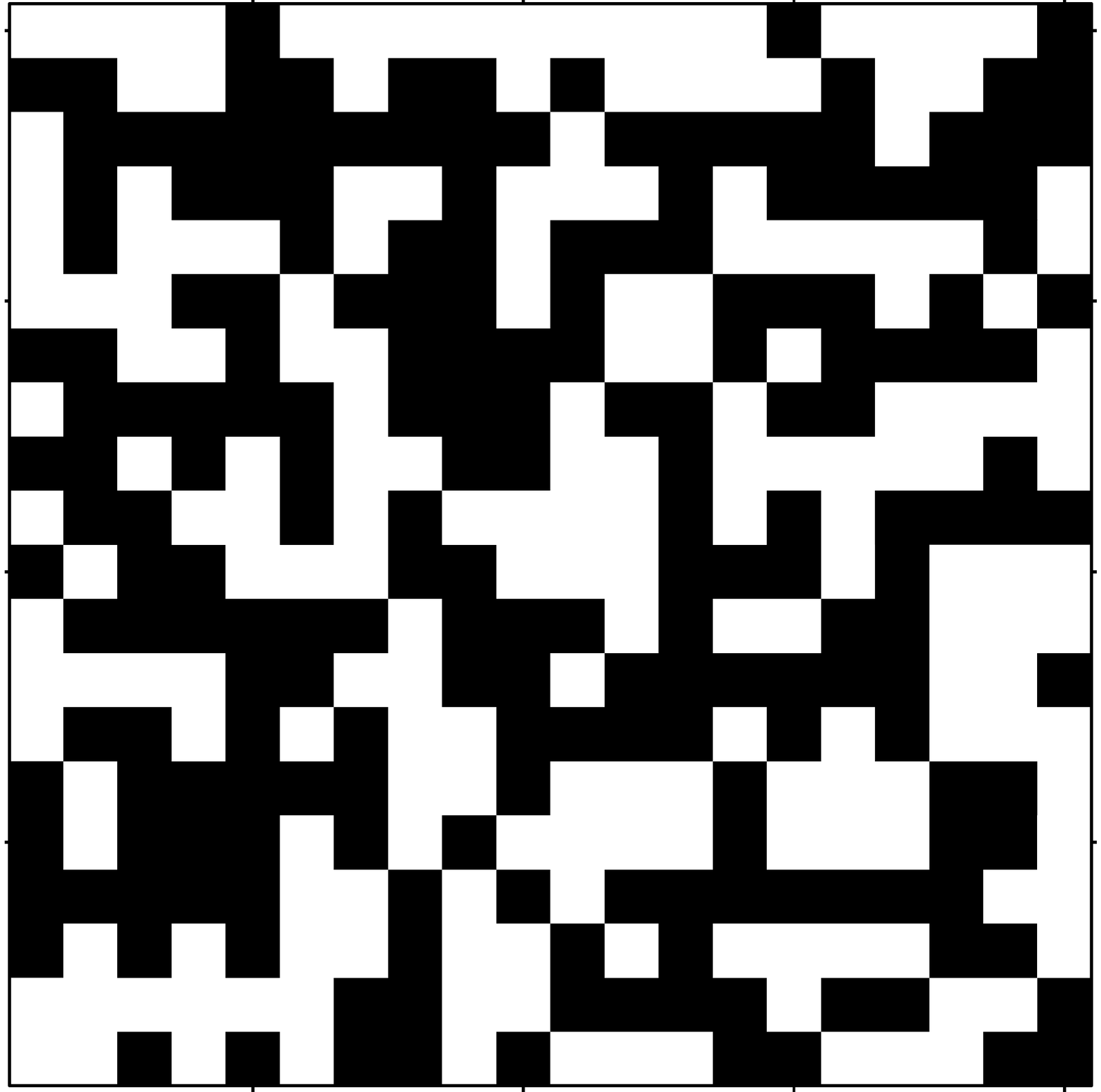}
   \subcaption{Segmentation result \\ }\label{fig: result}
  \end{minipage}
  \begin{minipage}[t]{0.25\hsize}
   \includegraphics[width = \linewidth]{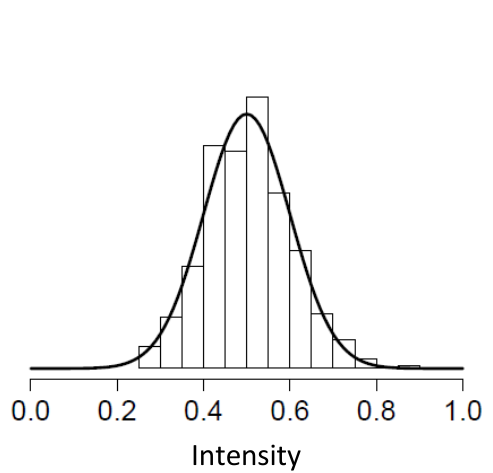}
   \subcaption{Pixel distribution}\label{fig: dist}
  \end{minipage}
  \begin{minipage}[t]{0.25\hsize}
   \includegraphics[width = \linewidth]{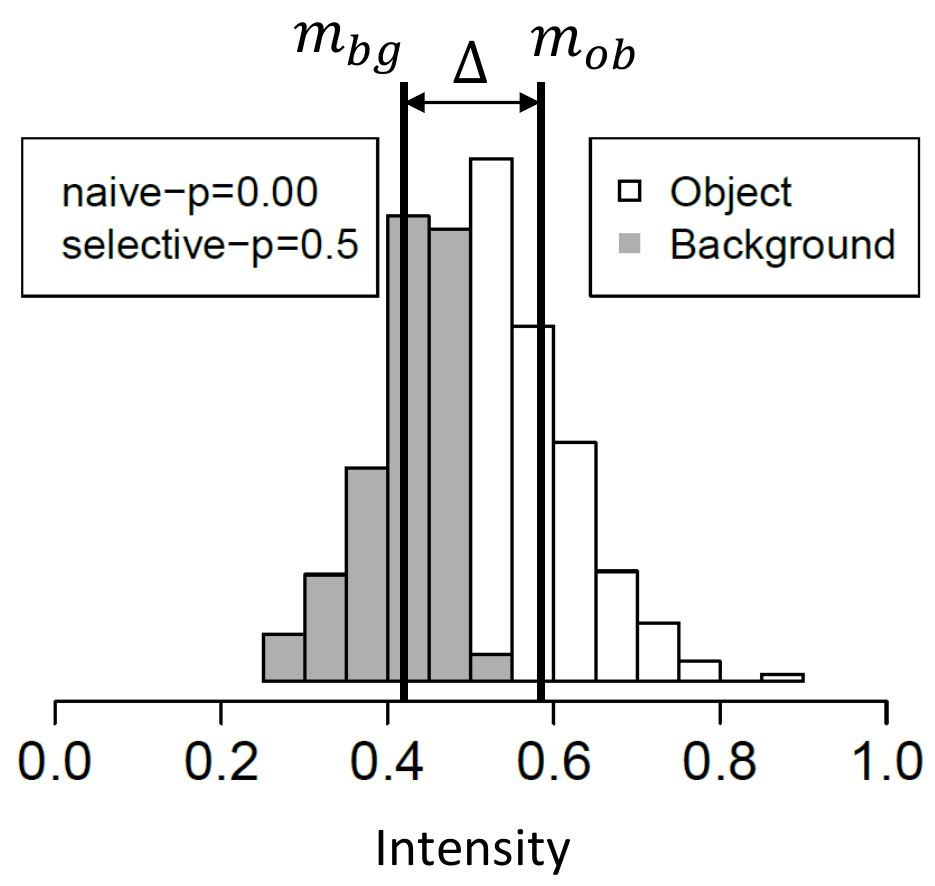}
   \subcaption{Ob.\ and Bg.\ histograms}\label{fig: dist_seg}
  \end{minipage}
  %
  %  \begin{minipage}[b]{0.4\hsize}
  %   \includegraphics[width = \linewidth]{image.pdf}
  %   \subcaption{Original image} \label{fig: image}
  %  \end{minipage}
  %  \begin{minipage}[b]{0.4\hsize}
  %   \includegraphics[width =\linewidth]{result.pdf}
  %   \subcaption{Segmentation result \\ }\label{fig: result}
  %  \end{minipage}\\
  %  \begin{minipage}[b]{0.45\hsize}
  %   \includegraphics[width = \linewidth]{dist.pdf}
  %   \subcaption{Pixel distribution}\label{fig: dist}
  %  \end{minipage}
  %  \begin{minipage}[b]{0.45\hsize}
  %   \includegraphics[width = \linewidth]{dist_seg.pdf}
  %   \subcaption{Ob. and Bg. distributions}\label{fig: dist_seg}
  %  \end{minipage}
  \caption{
  Schematic illustration of segmentation bias that arises when the statistical significance of the difference between the object and background regions obtained with a segmentation algorithm is tested. 
  (a) Randomly generated image with $n=400$ pixels from $N(0.5, 0.1^2)$.
  (b) Segmentation result obtained with the local threshold-based segmentation algorithm in \cite{white1983local}. 
  (c) Distribution and histogram of pixel intensities. 
  (d) Histograms of pixel intensities in the object region (white) and the background region (gray). 
  Note that even for an image that contains no specific objects, the pixel intensities of the object and background regions are clearly different. 
  Thus, if we naively compute the statistical significance of the difference, the $p$-value (naive-$p$ in (d)) would be very small, indicating that it cannot be used for properly evaluating the reliability of the segmentation result. 
  In this paper, we present a novel framework for computing valid $p$-values (selective-$p$ in (d)), which properly account for segmentation bias. 
  }\label{fig: bias}
 \end{center}
\end{figure*}

Although this problem seems fairly simple, computing a valid $p$-value for the above statistical hypothesis test is challenging because the difference in
the average pixel intensities between the object and background regions
can be deceptively
large even in a null image that contains no specific objects
since the segmentation algorithm divides the image into two regions and thus the difference is enlarged by adapting to the data. 
We refer to this deceptive difference for a segmentation result as segmentation bias.
It can be interpreted that
segmentation bias arises because the image pixel data are used twice:
once for dividing the object and background regions with a segmentation algorithm,
and again for testing the difference in the average intensities between the two regions.
Such data analysis is often referred to as double-dipping \cite{kriegeskorte2009circular} in statistics,
and it has been recognized that naively computed $p$-values in double-dipping data analyses are highly biased.
Figure \ref{fig: bias}
illustrates
segmentation bias
in a simple simulation.

In the proposed PSegI framework,
we overcome this difficulty
by introducing a recently developed statistical approach called selective inference (SI).
SI has been mainly studied for the statistical analysis of linear model coefficients after feature selection, which is interpreted as an example of double-dipping \cite{fithian2014optimal,taylor2014exact,taylor2015statistical,loftus2015selective,lee2016,tibshirani2016exact,suzumura2017selective,tian2018selective,taylor2018post} \footnote{
The main idea of the SI approach was first developed in \cite{lee2016} for computing the $p$-values of the coefficients of LASSO~\cite{tibshirani1996regression}.
This problem is interpreted as an instance of double-dipping data analysis since the training set is used twice: once for selecting features with $L_1$ penalized fitting, and again for testing the statistical significances of the coefficients of the selected features.
}.
Our paper has three main contributions.
First, we propose the PSegI framework in which the problem of quantifying the reliability of individual segmentation results is formulated as an SI problem, making the framework potentially applicable to a wide range of existing segmentation algorithms.
Second, we specifically study
the GC-based segmentation algorithm~\cite{boykov2001gc,boykov2006gc}
and
the TH-based segmentation algorithm~\cite{white1983local}
as examples,
and develop two specific PSegI methods,
called
PSegI-GC
and
PSegI-TH,
for computing valid $p$-values for the segmentation results obtained with these two respective segmentation algorithms.
Finally, we apply the PSegI-GC and PSegI-TH methods to medical images to demonstrate their efficacy.

\paragraph{Related work.}
A variety of image segmentation algorithms with different loss functions have been developed for computer vision by incorporating various properties of the target images~\cite{liu2007survey,elnakib2011survey,zhao2017survey}.
The performance of a segmentation algorithm is usually measured based on a human-annotated ground-truth dataset.
One of the most commonly used evaluation criteria for segmentation algorithms is the area under the curve (AUC).
Unfortunately, criteria such as AUC cannot be used to quantify the reliability of individual segmentation results. 
The segmentation problem can also be viewed as a two-class classification problem that classifies pixels into object and background classes.
Many two-class classification algorithms can provide some level of confidence that a given pixel belongs to the object or the background, e.g., by estimating the posterior probabilities~\cite{kendall2017uncertainties,hershkovitch2018model,qin2011image,grady2006random}.
Although confidence measures can be used to assess the relative reliability of a given pixel, they do not quantify the statistical significance of the segmentation result.
%
%The proposed PSegI framework has the following characteristics compared with existing methods:
%%
%\begin{itemize}
%
% \item It enables the evaluation of the reliability of each individual segmentation result, which is fundamentally different from conventional AUC analysis that aims to evaluate the segmentation algorithm itself.
%
% \item It can provide valid $p$-values for the difference in average pixel intensities between different regions, enabling proper control of the risk that the obtained segmentation result is incorrect.
%
% \item It only requires images from the null hypothesis, i.e., images that are known to contain only background information. Furthermore, it does not require any expert side information nor any `ground-truth' images.
%
% \item The background image is assumed to follow an $n$-dimensional normal distribution where $n$ is the number of pixels in the image. If not, the pixel intensities should be appropriately converted to have a normal distribution.
%
% \item It can properly take into account the difference in segmentation algorithms. In other words, even if two segmentation results are same, the $p$-values provided by the PSegI methods can be different because the degree of segmentation bias will be different depending on which segmentation algorithm is used.
%
%\end{itemize}
%
In the framework called \emph{contrario approach} which has been studied in CV community, a similar discussion as this study has been made on the reliability of object detected from a noisy image~\cite{desolneux2000meaningful,desolneux2003grouping,rousseau2008contrario,von2012lsd}.
Contrario approach is based on the \emph{Helmholts principle} that any objects should not be detected if they can be detected with high-probability even from a noise-only image. 
In \cite{rousseau2008contrario}, contrario approach is used in the context of image segmentation problem where the reliability of individual segmentation results is essentially evaluated based on a binomial test.
Unfortunately, however, contrario approach does not properly incorporate the bias arising from the adaptivity of the employed object detection algorithm.
Thus, the reliability measure discussed in \cite{rousseau2008contrario} cannot be used as $p$-value of individual segmentation results. 

%The  Helmholtz principle~\cite{desolneux2007gestalt} states that no detection should be produced on an image of noise. Accordingly, the a contrario approach~\cite{desolneux2000meaningful,desolneux2003grouping,rousseau2008contrario,von2012lsd} proposes to define a noise or a contrario model  where the desired structure is not present.
%
%Simply speaking, a-contrario approach quantifies the significance of the object region by a binomial-test.
%
%However, a-contrario approach is invalid in the sense that it does not address the segmentation bias issue.
%
%The contrario approach does not properly incorporate the bias arising from the adaptivity of the segmentation algorithm.

To the best of our knowledge, no previous studies have tackled the seemingly simple but challenging problem of computing $p$-values for individual segmentation results by properly correcting for segmentation bias.

\paragraph{Notation.}
We use the following notation in the rest of the paper.
For a scalar $s$, ${\rm sgn}(s)$ is the sign of $s$, i.e., ${\rm sgn}(s) = 1$ if $s \ge 0$ and $-1$ otherwise.
For a condition $c$, $\bm 1\{c\}$ is the indicator function, which returns 1 if $c$ is true and $0$ otherwise.
For natural number $j < n$, $\bm e_{j}$ is a vector of length $n$ whose $j^{\rm th}$ element is 1 and whose other elements are 0.
Similarly, for a set $\cS \subseteq \{1, \ldots, n\}$, $\bm e_\cS$ is an $n$-dimensional vector whose elements in $\cS$ are 1 and 0 otherwise.

%%%%%%%%%%%%%%%%%%%%%%%%%%%%%%%%%%
\section{Problem Setup}
\label{sec:sec2}
Consider an image with $n$ pixels.
We denote the preprocessed pixel values after appropriate filtering operations as $x_1, \ldots, x_n \in \RR$, i.e., the $n$-dimensional vector $\bm x := [x_1, \ldots, x_n]^\top \in \RR^n$ represents the preprocessed image. 
For simplicity, we only study segmentation problems in which an image is divided into two regions\footnote{The proposed PSegI framework can be easily extended to cases where an image is divided into more than two regions.}.
We call these two regions the object region and the background region for clarity.
After a segmentation algorithm is applied, $n$ pixels are classified into one of the two regions. 
We denote the set of pixels classified into the object and background regions as $\cO$ and $\cB$, respectively.
With this notation, a segmentation algorithm $\cA$ is considered to be a function that maps an image $\bm x$ into the two sets of pixels $\cO$ and $\cB$, i.e., $\{\cO, \cB\} = \cA(\bm x)$.
%
%Consider an image with $n$ pixels.
%%
%We denote the pixel intensities as $x_1, \ldots, x_n \in \RR$. The $n$-dimensional vector $\bm x := [x_1, \ldots, x_n]^\top \in \RR^n$ represents the image.
%%
%For simplicity, we only study segmentation problems in which an image is divided into two regions\footnote{The proposed PSegI framework can be easily extended to cases where an image is divided into more than two regions.}.
%%
%We call these two regions the object region and the background region for clarity.
%%
%After a segmentation algorithm is applied, $n$ pixels are classified into one of the two regions. 
%%
%We denote the set of pixels classified into the object and background regions as $\cO$ and $\cB$, respectively.
%%
%With this notation, a segmentation algorithm $\cA$ is considered to be a function that maps an image $\bm x$ into the two sets of pixels $\cO$ and $\cB$, i.e., $\{\cO, \cB\} = \cA(\bm x)$.

\subsection{Testing Individual Segmentation Results}
To quantify the reliability of individual segmentation results, consider a score $\Delta$ that represents how much the object and the background regions are different. 
The PSegI framework can be applied to any scores if it is written in the form of $\Delta = \bm \eta^\top \bm x$ where $\bm \eta \in \RR^n$ is any $n$-dimensional vector. 
For example, if we denote the average pixel values of the object and the background regions as 
\begin{align*}
 m_{\rm ob} = \frac{1}{|\cO|} \sum_{p \in \cO} x_p,
 ~
 m_{\rm bg} = \frac{1}{|\cB|} \sum_{p \in \cB} x_p,
\end{align*}
and
define the $\bm \eta$ as
\begin{align*}
 \eta_i = \mycase{
 {\rm sgn}(m_{\rm ob} - m_{\rm bg}) / |\cO|, & \text{ if } i \in \cO, \\
 {\rm sgn}(m_{\rm bg} - m_{\rm ob}) / |\cB|, & \text{ if } i \in \cB, 
 } 
\end{align*}
then the score $\Delta$ represents the absolute average difference in the pixel values between object and background region 
\begin{align}
 \label{eq:pixel-difference}
 \Delta = | m_{\rm ob} - m_{\rm bg} |.
\end{align}
In what follows, for simplicity, we assume that the score $\Delta$ is in the form of \eq{eq:pixel-difference}, but any other scores in the form of $\bm \eta^\top \bm x$ can be employed. Other specific examples are discussed in supplement A.

%To quantify the reliability of individual segmentation results, we simply consider the difference in average pixel intensities between the object and background regions\footnote{Any statistics can be used in our framework if it is represented as a linear combination of pixel intensities. Specific examples are considered in supplement A.}, defined as 
%\begin{align}
% \label{eq:pixel-difference}
% \Delta = | m_{\rm ob} - m_{\rm bg} |,
%\end{align}
%where
%\begin{align*}
% m_{\rm ob} = \frac{1}{|\cO|} \sum_{p \in \cO} x_p,
% ~
% m_{\rm bg} = \frac{1}{|\cB|} \sum_{p \in \cB} x_p. 
%\end{align*}
%%
If the difference $\Delta$ is sufficiently large, it implies that the segmentation result is reliable. 
As discussed in \S1, it is non-trivial to properly evaluate the statistical significance of the difference $\Delta$ since it can be deceptively large due to the effect of segmentation bias.
To quantify the statistical significance of the difference $\Delta$, we consider a statistical hypothesis test with the following null hypothesis $\rm H_0$ and alternative hypothesis $\rm H_1$: 
\begin{align}
 \label{eq:hypothesis_testing}
 {\rm H_0}: \mu_{\rm ob} = \mu_{\rm bg}
 ~~~
 {\rm vs.}
 ~~~
 {\rm H_1}: \mu_{\rm ob} \neq \mu_{\rm bg},
\end{align}
where
$\mu_{\rm ob}$
and
$\mu_{\rm bg}$
are the true means of the pixel intensities in the object and background regions, respectively. 
Under the null hypothesis ${\rm H_0}$,
we assume that an image consists only of background information, and that the statistical variability of the background information can be represented by an $n$-dimensional normal distribution $N(\bm \mu, \Sigma)$, 
where
$\bm \mu \in \RR^n$
is the unknown mean vector 
and 
$\Sigma \in \RR^{n \times n}$
is the covariance matrix,
which is known or estimated from independent data.
In practice, we estimate the covariance matrix $\Sigma$ from a null image in which we know that there exists no object.
For example, in medical image analysis, it is not uncommon to assume the availability of such null images. 

In a standard statistical test, the $p$-value is computed based on the null distribution $\PP_{\rm H_0}(\Delta)$, i.e., the sampling distribution of the test statistic $\Delta$ under the null hypothesis. 
%
%If we consider an \emph{unrealistic} assumption that the segmentation result $\{\cO, \cB\}$ is determined independently of the pixel intensities $\bm x$, the null distribution can be computed analytically. 
%
%In \emph{reality}, however, the segmentation result $\{\cO, \cB\}$ depends on the pixel intensities $\bm x$ in a complicated way through the segmentation algorithm $\cA$, so the computation of the null distribution is intractable.
%
%We call the $p$-value computed under the above unrealistic assumption as \emph{naive $p$-value}. 
%
%When the unrealistic assumption is not met (as it is not in reality), the naive $p$-values are highly underestimated due to the segmentation bias, and hence the probability of finding incorrect segmentation results cannot be properly controlled. 
%
On the one hand, if we naively compute the $p$-values from the pixel intensities in $\cO$ and $\cB$ without considering that $\{\cO, \cB\}$ was obtained with a segmentation algorithm, these naive $p$-values will be highly underestimated due to segmentation bias, and hence the probability of finding incorrect segmentation results cannot be properly controlled. 
On the other hand, it is intractable to compute the sampling distribution $\PP_{\rm H_0}(\Delta)$ by properly incorporating the effect of segmentation bias since the segmentation algorithm is complicated. 

\begin{figure}[t]
 \begin{center}
  \includegraphics[width=0.7\linewidth]{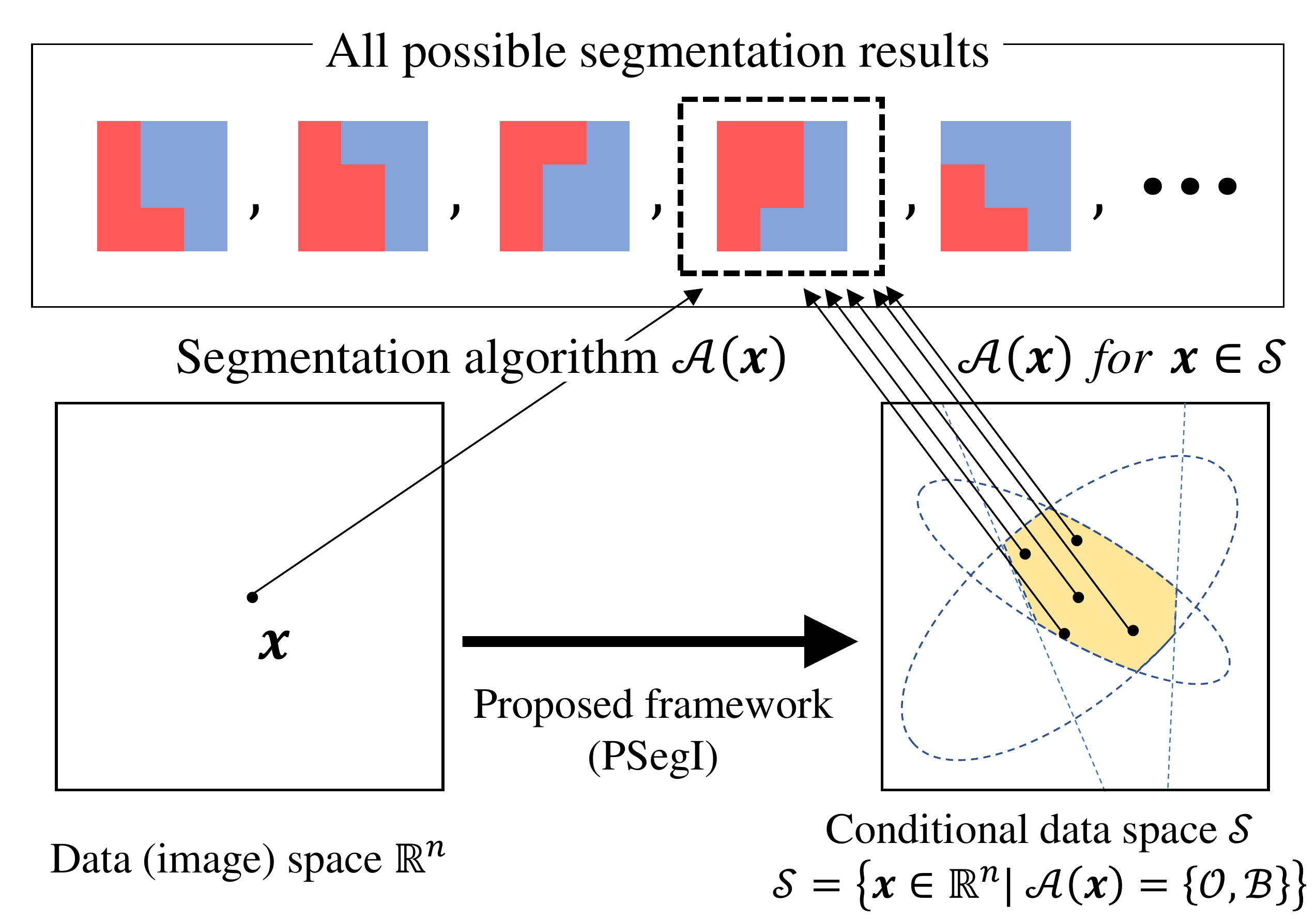}
  \caption{
  Schematic illustration of the basic idea used in the proposed PSegI framework. 
  By applying a segmentation algorithm $\cA$ to an image $\bm x$ in the data space $\RR^n$, a segmentation result $\{\cO, \cB\}$ is obtained. 
  In the PSegI framework, the statistical inference is conducted conditional on the subspace $\cS = \{\bm x \in \RR^n \mid \cA(\bm x) = \{\cO, \cB\}\}$; i.e., the subspace is selected such that an image taken from the subspace has the same segmentation result $\{\cO, \cB\}$. 
  By conditioning on the segmentation result $\{\cO, \cB\}$, valid $p$-values that properly account for segmentation bias can be computed. 
  The proposed PSegI framework is applicable to a given segmentation algorithm if the segmentation process can be characterized by a finite set of quadratic inequalities on $\bm x$. 
  }
  \label{fig: SI}
 \end{center}
\end{figure}

\paragraph{Selective inference.}
SI is a type of conditional inference, in which a statistical test is conducted based on a conditional sampling distribution of the test statistic under the null hypothesis.
In our problem, to account for segmentation bias, we consider the statistical test for the difference $\Delta$ conditional on the segmentation result $\{\cO, \cB\}$, and consider the following conditional null distribution:
\begin{align}
 \label{eq:conditional-null-distribution}
 \PP_{\rm H_0}(\Delta \mid \{\cO, \cB\} = \cA(\bm x)). 
\end{align}
In \eq{eq:conditional-null-distribution}, we do not consider all possible images $\bm x \in \RR^n$, but rather a subset of images $\bm x$ such that $\{\cO, \cB\} = \cA(\bm x)$.
Figure \ref{fig: SI} shows a schematic illustration of the basic idea used in the proposed PSegI framework.
Basically, by considering the sampling distribution only for the cases for which the segmentation result is $\{\cO, \cB\}$, segmentation bias can be properly corrected. 
In this paper, we propose concrete methods for computing $p$-values based on the conditional null distribution \eq{eq:conditional-null-distribution} when a GC-based or TH-based segmentation algorithm is used as $\cA$. 
We theoretically prove that the computed $p$-values are valid in the sense that they control the risk that the observed segmentation result is incorrect at a specified significance level $\alpha$ (e.g., $\alpha = 0.05$).

\subsection{Graph-cut-based Segmentation}
As one of the two examples of segmentation algorithm $\cA$, we consider the GC-based segmentation algorithm in \cite{boykov2001gc,boykov2006gc}. 
In GC-based segmentation algorithms, the target image is considered to be a directed graph 
$\cG = \{\cV, \cE\}$,
where
$\cV$
and 
$\cE$
are the sets of nodes and edges, respectively. 
Let
$\cP$
be the set of all $n$ pixels,
and
$\cN$
be the set of all directed edges from each pixel to its eight adjacent nodes (each pixel is connected to its horizontally, vertically, and diagonally adjacent pixels). 
Furthermore, consider two terminal nodes $S$ and $T$. 
Then,
$\cV$ and $\cE$ of the graph $\cG$ 
are defined as 
$\cV = \cP \cup \{S, T\}$
and
$\cE = \cN \cup \bigcup_{p \in \cP} \{(S, p), (p, T)\}$,
where
for the two nodes $p$ and $q$,
$(p, q)$
indicates the directed edge from $p$ to $q$. 
At each edge of the graph $(p, q) \in \cE$, non-negative weights $w_{(p, q)}$ are defined based on the pixel intensities (see \S3.2). 

In GC-based segmentation algorithms, segmentation into object and background regions is conducted by cutting the graph into two parts.
Let us write the ordered partition of the graph as
$(\cV_s, \cV_t)$,
where 
$\cV_s$
and
$\cV_t$
constitute a partition of $\cV$.
If
$S \in \cV_s$ 
and 
$T \in \cV_t$,
the ordered partition
$(\cV_s, \cV_t)$
is called an $s$-$t$ cut. 
Let $\cE_{\rm cut} \subset \cE$ be the set of edges $(p, q) \in \cE$ such that $p$ belongs to $\cV_s$ and $q$ belongs to $\cV_t$. 
The cost function of an $s$-$t$ cut $(\cV_s, \cV_t)$ is defined as
%\begin{align*}
$L_{\rm cut}(\cV_s, \cV_t) = \sum_{(p, q) \in \cE_{\rm cut}} w_{(p, q)}$. 
%\end{align*}
%
The GC-based segmentation algorithm is formulated as the optimization problem for finding the optimal $s$-$t$ cut:
\begin{align}
 \label{eq:mincut-problem}
(\cV_s^*, \cV_t^*) = \arg \min_{(\cV_s, \cV_t)} L_{\rm cut}(\cV_s, \cV_t).
\end{align}
Then, the segmentation result $\{\cO, \cB\}$ is obtained as 
%\begin{align*}
$\cO \leftarrow \cV_s^* \setminus \{S\}$,
and 
$\cB \leftarrow \cV_t^* \setminus \{T\}$.
%\end{align*}
%
The minimum cut problem
\eq{eq:mincut-problem}
is known to be a dual problem of the maximum flow problem, for which polynomial time algorithms exist~\cite{ford2009maximal,goldberg1988new,dinic1970algorithm}.
Among the several implementations of the maximum flow problem, we employed the one presented in \cite{boykov2004experimental}, in which three stages, called
the grow stage,
the augment stage,
and
the adopt stage,
are iterated. 
Briefly, a path from $S$ to $T$ is obtained in the grow stage. The edge with the minimum weight in the path is selected and all the weights of the path are reduced by the minimum weight to account for the flow in the augment stage. The data structure of the updated graph is reconstructed in the adopt stage (see \cite{boykov2004experimental} for details). 

\subsection{Threshold-based Segmentation}
Next, we briefly describe TH-based segmentation algorithms~\cite{sezgin2004survey}.
In TH-based segmentation algorithms, pixels are simply classified into either the object or background class depending on whether their intensity is greater or smaller than a certain threshold.
According to the application and the features of the target images, various approaches for determining the threshold have been proposed. 
In the following, we first describe a global TH algorithm in which a single threshold is used for the entire image, and then present a local TH algorithm in which different thresholds are used for different pixels.

In the method proposed by Otsu~\cite{otsu1979global}, the global threshold is selected to maximize the dispersion between the object and background pixels.
Here, dispersion is defined so that the between-region variance is maximized and the within-region variance is minimized. 
Since the sum of these two variances is the total variance and does not depend on the threshold, we can only maximize the former.
Denoting the number, mean, and variance of the pixels with values greater (resp. smaller) than the threshold $t$ as $\overline{n}_t$, $\overline{\mu}_t$, and $\overline{\sigma}^2_t$ (resp. $\underline{n}_t$, $\underline{\mu}_t$, and $\underline{\sigma}^2_t$), respectively, the between-region variance with the threshold $t$ is written as
$\sigma_{\rm bet}^2(t) = \frac{\overline{n}(t) \underline{n}(t)(\overline{\mu}(t) - \underline{\mu}(t) )^2}{n^2}$.
The global threshold is then determined as $t^* = \arg \max_t \sigma_{\rm bet}^2(t)$.
Although this algorithm is simple, it is used in many practical applications.

The local thresholding approach allows more flexible segmentation since the threshold is determined per pixel. 
The method
proposed by White and Rohrer~\cite{white1983local}
determines the pixel-wise threshold 
by comparing the pixel intensity
with
the average pixel intensity of its neighbors.
Here, neighbors are defined by a square window around a pixel, and the local threshold for the pixel is determined as
$t_p^* = |\cW_p|^{-1} \sum_{q \in \cW_p} x_q / \theta$, 
where
$\cW_p$
is the set of pixels within the window around the pixel $p$,
and
$\theta$
is a scalar value specified based on the properties of the image.

%%%%%%%%%%%%%%%%%%%%%%%%%%%%%%%
\section{Post-segmentation Inference}
\label{sec:sec3}
In this section,
we consider the problem of how to provide 
a valid $p$-value
for the observed difference in average pixel intensities
between the object and background regions
$\Delta$ 
when the two regions are obtained by applying a segmentation algorithm
$\cA$
to an image $\bm x$. 
In the proposed PSegI framework, 
we solve this problem 
by considering the sampling distribution
of 
$\Delta$
conditional on the event
$\cA(\bm x) = \{\cO, \cB\}$
under the null hypothesis that
$\mu_{\rm ob} = \mu_{\rm bg}$; 
i.e., 
the actual mean pixel intensities of the object and background regions are the same. 
By conditioning on the segmentation result $\{\cO, \cB\}$, the effect of segmentation bias is properly corrected. 

By definition, a valid $p$-value must be interpreted as an upper bound of the probability that the obtained segmentation result $\{\cO, \cB\}$ is incorrect\footnote{Naive $p$-values do not satisfy this property due to segmentation bias.}.
To this end, in our conditional inference, a valid $p$-value $\mathfrak{p}$ must satisfy 
\begin{align*}
 %\label{eq:valid-p-property}
 \PP_{\rm H_0}(\mathfrak{p} \le \alpha \mid \cA(\bm x) = \{\cO, \cB\}) = \alpha, ~ \forall~\alpha \in [0, 1].
\end{align*}
This property is satisfied if and only if $\mathfrak{p}$ is uniformly distributed in $[0, 1]$. 
Therefore, our problem is cast into the problem of computing a function of the test statistic
$\Delta$ 
that follows
${\rm Unif}[0, 1]$
when the test statistic follows the conditional sampling distribution 
\begin{align}
 \label{eq:conditional-sampling-distribtuion}
 \PP_{\rm H_0}(\Delta ~\mid~ \cA(\bm x) = \{\cO, \cB\}). 
\end{align}

In \S3.1, we first present our main result for the proposed PSegI framework.
Here, we show that if the event $\cA(\bm x) = \{\cO, \cB\}$ is characterized by a finite set of quadratic inequalities on $\bm x$, then a valid $p$-value can be exactly computed. 
In \S3.2 and \S3.3, as examples of the proposed PSegI framework, we  develop concrete methods of, respectively, the PSegI framework for a GC-based segmentation algorithm~\cite{boykov2001gc,boykov2006gc} and a TH-based segmentation algorithm~\cite{otsu1979global,white1983local}. 
Our key finding is that the event $\cA(\bm x) = \{\cO, \cB\}$ can be characterized by a finite set of quadratic inequalities on $\bm x$ for these segmentation algorithms and thus valid $p$-values can be computed by using the result in \S3.1.

\subsection{Selective Inference for Segmentation Results}
The following theorem is the core of the proposed PSegI framework. 
\begin{theo}
 \label{theo:main-theorem}
 Suppose that an image $\bm x$ with size $n$ is drawn from an $n$-dimensional normal distribution $N(\bm \mu, \Sigma)$ with unknown $\bm \mu$ and known or independently estimated $\Sigma$. 
 If the event $\cA(\bm x) = \{\cO, \cB\}$ is characterized by a finite set of quadratic inequalities on $\bm x$
 in the form of
 \begin{align}
  \label{eq:quadratic-inequalities}
  \bm x^\top A_j \bm x + \bm b_j^\top \bm x + c_j \le 0, ~ j = 1, 2, \ldots, 
 \end{align}
 with certain 
 $A_j \in \RR^{n \times n}$, 
 $\bm b_j \in \RR^n$,  
 and
 $c_j \in \RR$, 
 $j = 1, 2, \ldots$, 
 then
 \begin{align}
  \mathfrak{p} = 1 - F_{0, \bm \eta^\top \Sigma \bm \eta}^{E(\bm z)}(|m_{\rm ob} - m_{\rm bg}|)
 \end{align}
 is a valid $p$-value in the sense that
 $\mathfrak{p}$
 is uniformly distributed in $[0, 1]$ conditional on the event $\cA(\bm x) = \{\cO, \cB\}$ and the sign of $m_{\rm ob} - m_{\rm bg}$, 
 where
 $F_{m, s^2}^E$
 is a cumulative distribution function of a truncated normal distribution with mean $m$, variance $s^2$, and truncation intervals $E$,
 and
 $\bm \eta = {\rm sgn}(m_{\rm ob} - m_{\rm bg}) (|\cO|^{-1} \bm e_{\cO} - |\cB|^{-1} \bm e_{\cB})$. 
 Here,
% \begin{align*}
 $
 E(\bm z) =
  \bigcap_j
  \{\tau > 0 \mid (\bm z + \tau \bm y)^\top A_j (\bm z + \tau \bm y) + \bm b_j^\top (\bm z + \tau \bm y) + c_j \le 0\}
 $,
% \end{align*}
 where
 $
 \bm y = \Sigma \bm \eta^\top \bm \eta / (\bm \eta^\top \Sigma \bm \eta)
 $
 and 
 $\bm z = \bm x - (m_{\rm ob} - m_{\rm bg}) \bm y$.
\end{theo}
The proof of Theorem~\ref{theo:main-theorem} is presented in supplement B.
This theorem is an adaptation of Theorem 5.2 in \cite{lee2016} and Theorem 3.1 in \cite{loftus2015selective}, in which SI on the selected features of a linear model was studied. 
%
%Here, we consider the conditional sampling distribution of the test statistic not only on the event $A(\bm x) = \{\cO, \cB\}$ but also on the sign of $m_{\rm ob} - m_{\rm bg}$ to handle the absolute operator of $\Delta$.
%
Note that the normality assumption in Theorem~\ref{theo:main-theorem} does NOT mean that $n$ pixel values are normally distributed; it means that the noise (deviation from the unknown true mean value) in each pixel value is normally distributed. 

\subsection{Valid $p$-values for GC-based segmentation}\label{GC-seg}
As briefly described in \S2-2, GC-based segmentation is conducted by solving the maximum flow optimization problem on the directed graph. 
Basically, all the operations in this optimization process can be decomposed into additions, subtractions, and comparisons of the weights $w_{(p, q)}$ of the directed graph. 
This suggests that as long as each weight $w_{(p, q)}$ is written as a quadratic function of the image $\bm x$, the event that the GC-based segmentation algorithm produces the segmentation result $\{\cO, \cB\}$ can be fully characterized by a finite set of quadratic inequalities in the form of \eq{eq:quadratic-inequalities}. 
In the following, we explain how to set the weights $w_{(p, q)}$ for each edge $(p, q) \in \cE$. 
%
%Although we slightly modified from the existing CG-based segmentation algorithm in \cite{} to use PSegI framework, our experience suggests that the segmentation result has not changed. 
%
For properly defining the weights, it is necessary to introduce seed pixels for the object and background regions. 
The pixels known or highly plausible to be in the object or background regions are set as the seed pixels, denoted as 
$\cO^{\rm se}, \cB^{\rm se} \subset \cP$,
respectively. 
The seed pixels may be specified by human experts, or the pixel with the largest or smallest intensity may be specified as the object or background seed pixel, respectively.

When the two pixel nodes $p, q \in \cP$ are adjacent to each other, the weight $w_{(p, q)}$ is determined based on the similarity of their pixel intensities and the distance between them. 
Pixel similarity is usually defined based on the properties of the target image.
To provide flexible choice of the similarity function, we employ a quadratic spline approximation, which allows one to specify the desired similarity function with arbitrary approximation accuracy.
For example, Figure 1 in supplement C  shows an example of the quadratic spline approximation of commonly used weights
$w_{(p,q)} = \exp(- (x_p - x_q)^2/(2 \sigma^2) ){\rm dist}(p, q)^{-1}$, where ${\rm dist}(p, q)$ is the distance between the two nodes. 

The weight
between the terminal node
$S$
and the general pixel node
$p \in \cP \setminus (\cO^{\rm se} \cup \cB^{\rm se})$
is usually determined
based on
the negative log-likelihood of the pixel in the object region. 
Under the normality assumption, it is written as 
%\begin{align*}
$
 w_{S,p}
 =
 -\log \PP(x_p \mid p \in \cO)
 \simeq
 \log(2 \pi \sigma^2 + (x_p - m_{\rm ob}^{\rm se})^2/(2 \sigma^2),
$
 %\end{align*}
where
$m_{\rm ob}^{\rm se} = \sum_{i \in \cO^{\rm se}} x_i / |\cO^{\rm se}|$
is the estimate of the mean pixel intensity in the object region from the object seed pixel intensities. 
The weight between the terminal node $S$ and an object seed pixel node $p \in \cO^{\rm se}$ should be sufficiently large. It is usually determined as 
$w_{S, p} = 1 + \max_{q \in \cP} \sum_{r : (q, r) \in \cN} w_{(q,r)}$. 
The weight between the terminal node $S$ and a background seed node $p \in \cB^{\rm se}$ is set to zero. 
The weights between the terminal node $T$ and pixel nodes are determined in the same way. 

Since
all the weights for the edges are represented
by
quadratic equations
and
quadratic constraints
on
$\bm x$,
all the operations for solving the minimum cut (or maximum flow) optimization problem
\eq{eq:mincut-problem} 
can be fully characterized by a finite set of quadratic inequalities in the form of 
\eq{eq:quadratic-inequalities}. 
Thus, valid $p$-values of the segmentation result obtained with the GC-based segmentation algorithm can be computed using the PSegI framework.
In supplement C,
we present all the matrices 
$A_j$,
vectors
$\bm b_j$,
and
scalars
$c_j$
needed for characterizing a GC-based segmentation event. 

\subsection{Valid $p$-values for TH-based segmentation}
Both the global and local TH algorithms are fit into the PSegI framework.
%
%Due to the space limitation, PSegI method for global threshold algorithm is presented in supplement.
%
First, consider the global TH algorithm. 
For simplicity, consider selecting a global threshold $t^*$ from 256 values $t \in \{0, 1, \ldots, 255\}$. 
An event that the global threshold $t^*$ is selected can be simply written as 
\begin{align}
 \label{eq:global-threshold1}
 \sigma_{\rm bet}^2(t^*) \ge \sigma_{\rm bet}^2(t), t \in \{0, \ldots, 255\}. 
\end{align}
Let
$\overline{\bm u}(t)$
and
$\underline{\bm u}(t)$ 
be $n$-dimensional vectors whose elements are defined as 
\begin{align*}
 \overline{u}(t)_p = \mycase{
 1 & \text{ if } x_p \ge t, \\
 0 & \text{ otherwise;}
 }
 \underline{u}(t)_p = \mycase{
 0 & \text{ if } x_p \ge t, \\
 1 & \text{ otherwise.}
 }
\end{align*}
Then,
since
the between-region variance
$\sigma_{\rm bet}^2(t)$ 
is written as the quadratic function 
\begin{align*}
% \sigma_{\rm bet}^2(t)
% = 
 \bm x^\top
 \left(
 \frac{\underline{n}(t)}{\overline{n}(t)}
 \overline{\bm u}(t) \overline{\bm u}(t)^\top
 + 
 \frac{\overline{n}(t)}{\underline{n}(t)}
 \underline{\bm u}(t) \underline{\bm u}(t)^\top
 -
 2 \overline{\bm u}(t) \underline{\bm u}(t)^\top
 \right)
 \bm x,
\end{align*}
the event in 
\eq{eq:global-threshold1}
is represented by 255 quadratic inequalities on $\bm x$. 
Furthermore, it is necessary to specify whether pixels are in the object or background region at each threshold $t \in \{0, \ldots, 255\}$. 
To this end, consider conditioning on the order of pixel intensities, which is represented by a set of $n-1$ linear inequalities: 
\begin{align}
 \label{eq:global-threshold-order}
 \bm e_{(i)}^\top \bm x \le \bm e_{(i+1)}^\top \bm x, i = 1, \ldots, n-1,
\end{align}
where $(1), (2), \ldots, (n)$ is the sequence of pixel IDs such that $x_{(1)} \le x_{(2)} \le \ldots \le x_{(n)}$. 
Since the conditions 
\eq{eq:global-threshold1}
and 
\eq{eq:global-threshold-order}
are represented by sets of quadratic and linear inequalities on $\bm x$,
valid $p$-values of the segmentation result obtained with the global TH algorithm can be computed 
using the PSegI framework.

Next, consider the local threshold approach.
The conditions under which the $p^{\rm th}$ pixel is classified into the object or background region are simply written as a set of linear inequalities on $\bm x$ as 
\begin{align*}
 & x_p \ge (|\cW_p|^{-1} \sum_{q \in \cW_p} x_q)/\theta 
 ~\Leftrightarrow~
 \bm e_p^\top \bm x \ge |\cW_p|^{-1} \bm e_{\cW_p}^\top \bm x, \\
 & x_p \le (|\cW_p|^{-1} \sum_{q \in \cW_p} x_q)/\theta
 ~\Leftrightarrow~
 \bm e_p^\top \bm x \le |\cW_p|^{-1} \bm e_{\cW_p}^\top \bm x, 
\end{align*}
respectively.
Thus,
valid $p$-values of the segmentation result obtained with the local TH-based algorithm can be computed 
using the PSegI framework.

%%%%%%%%%%%%%%%%%%%%%%%%%%%%%%%
\section{Experiment}
We confirm the validity of the proposed method by numerical experiments.
First, we evaluated the false positive rate (FPR) and the true positive rate (TPR) of the proposed method with artificial data. 
Then, we applied the proposed method to medical images as a practical application.
We compared the proposed method with the naive method, which assumes that $\Delta \sim N(0, \tilde{\sigma}^2)$, where $\tilde{\sigma}^2$ is computed based on the segmentation result without considering segmentation bias. 
We denote the $p$-values obtained using the proposed method and the naive method as selective-$p$ and naive-$p$, respectively. 

\subsection{Experiments using artificial data}
In the artificial data experiments, Monte Carlo simulation was conducted $10^{5}$ times.
The significance level was set to $\alpha = 0.05$ and the FPRs and TPRs were estimated as $10^{-5} \sum_{i = 1}^{10^{5}} \bm{1} \left\{pvalue_i < \alpha \right\}$, where $pvalue_i$ is the $p$-value at the $i^{\rm th}$ Monte Carlo trial. 
Data were generated with the range of pixel values $x\in[0,1]$. The maximum and minimum values were used as the seeds for the object and background regions, respectively.
Note that these seed selections were incorporated as selection events.

In the experiment for FPR, the data were randomly generated as $\bm x \sim {\rm N}(\bm{0.5}_{n},0.5 I_{n\times n})$ for $n=9, 25, 100, 225, 400, 625, 900$. 
\begin{figure}[t]
	\begin{minipage}[b]{0.49\hsize}
		\centering
		\includegraphics[width = 0.8\linewidth]{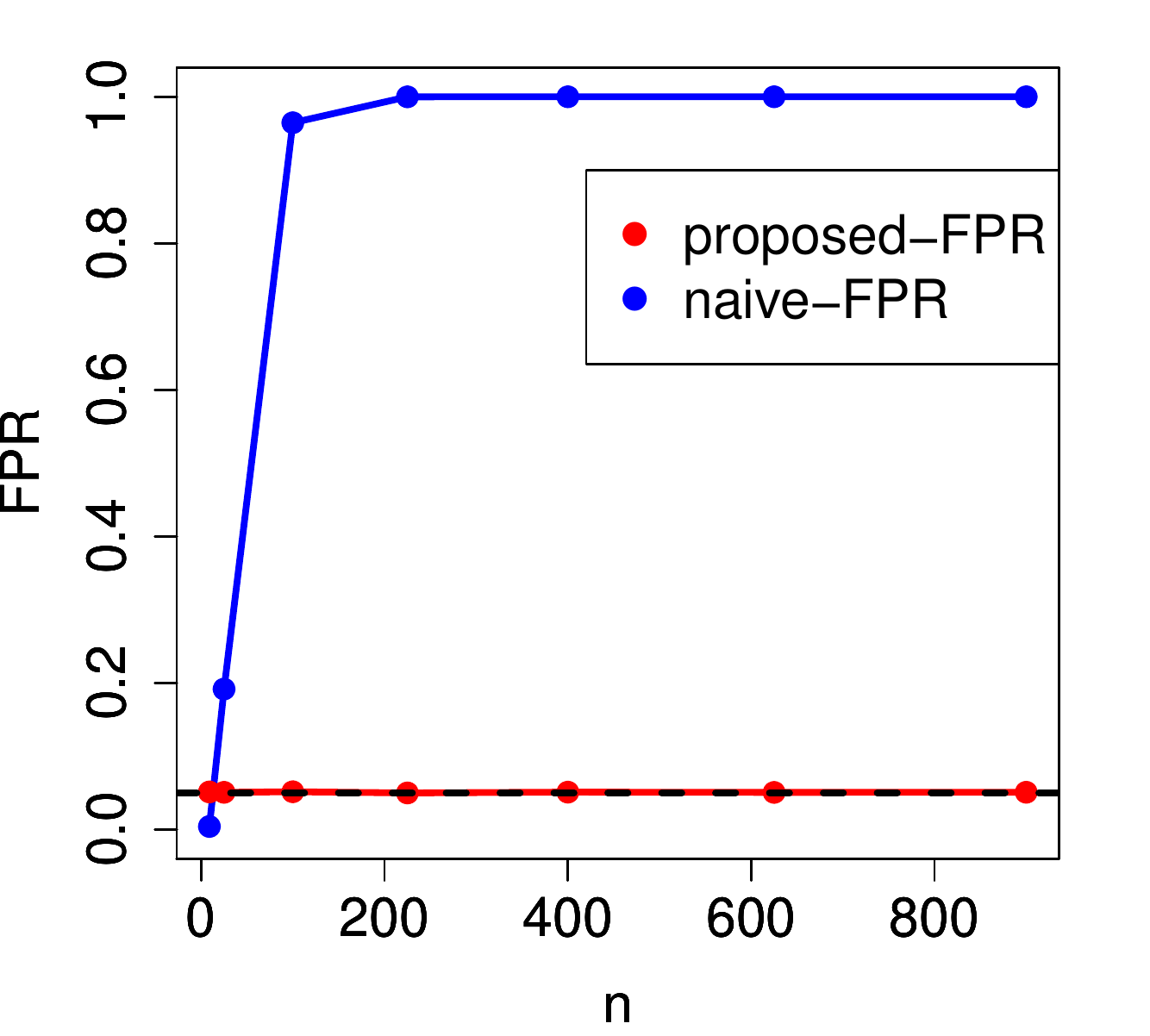}
		\subcaption{FPR for GC} \label{fig: FPR}
	\end{minipage}
	\begin{minipage}[b]{0.49\hsize}
		\centering
		\includegraphics[width = 0.8\linewidth]{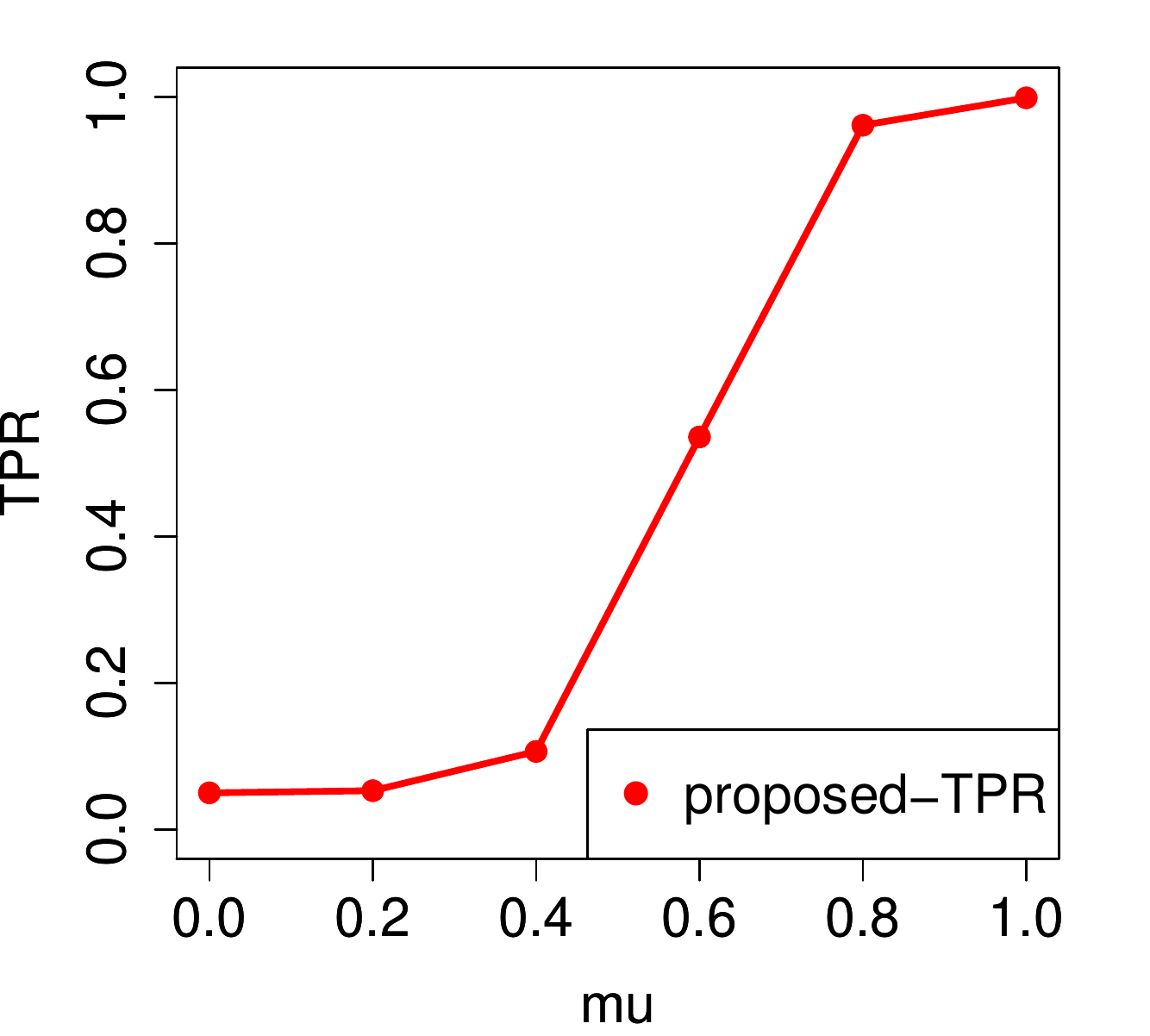}
		\subcaption{TPR for GC} \label{fig: TPR}
	\end{minipage}
	\\
	\begin{minipage}[b]{0.49\hsize}
		\centering
		\includegraphics[width=0.8\linewidth]{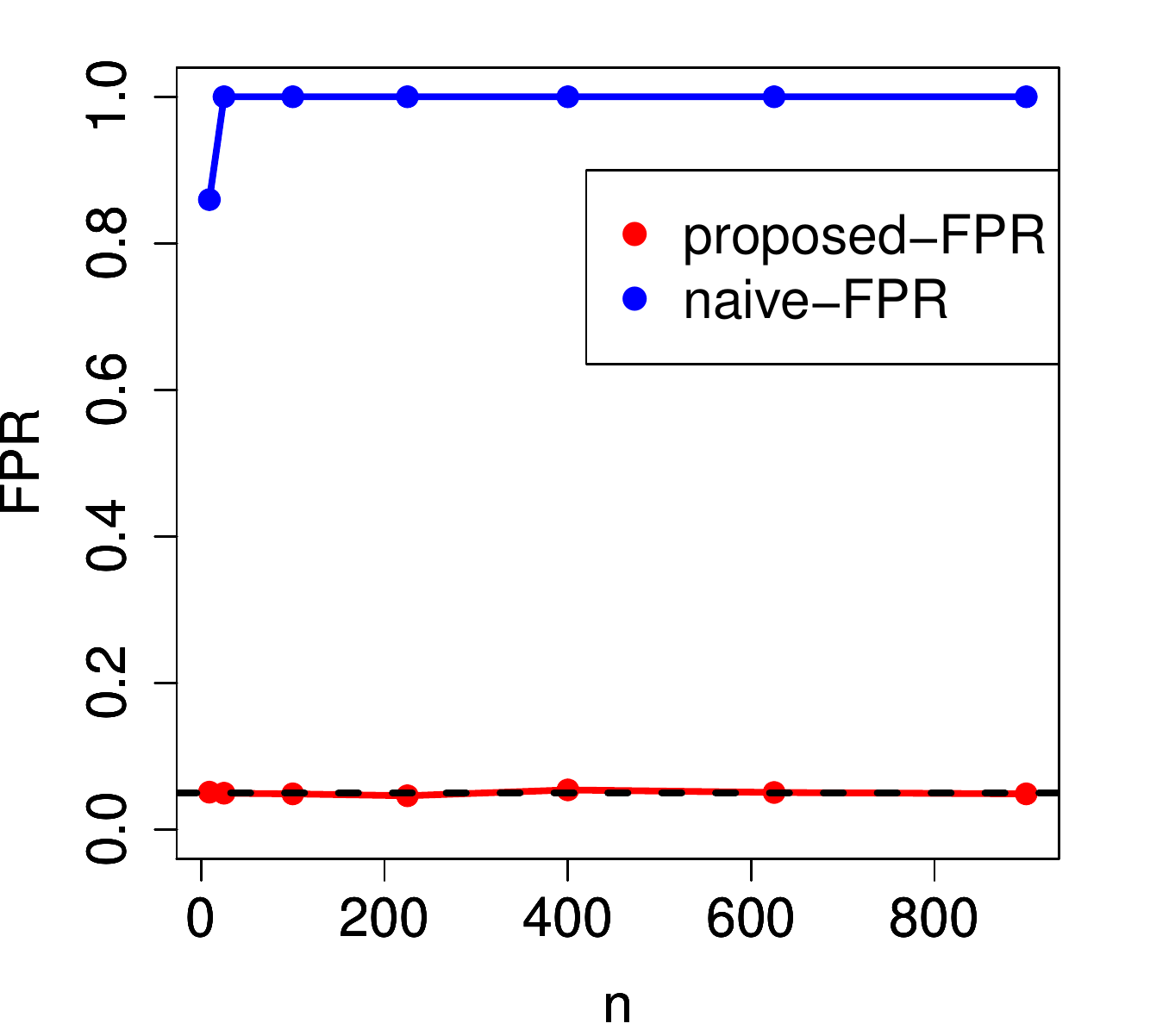}
		\subcaption{FPR for TH} \label{fig: thFPR}
	\end{minipage}
	\begin{minipage}[b]{0.49\hsize}
		\centering
		\includegraphics[width=0.8\linewidth]{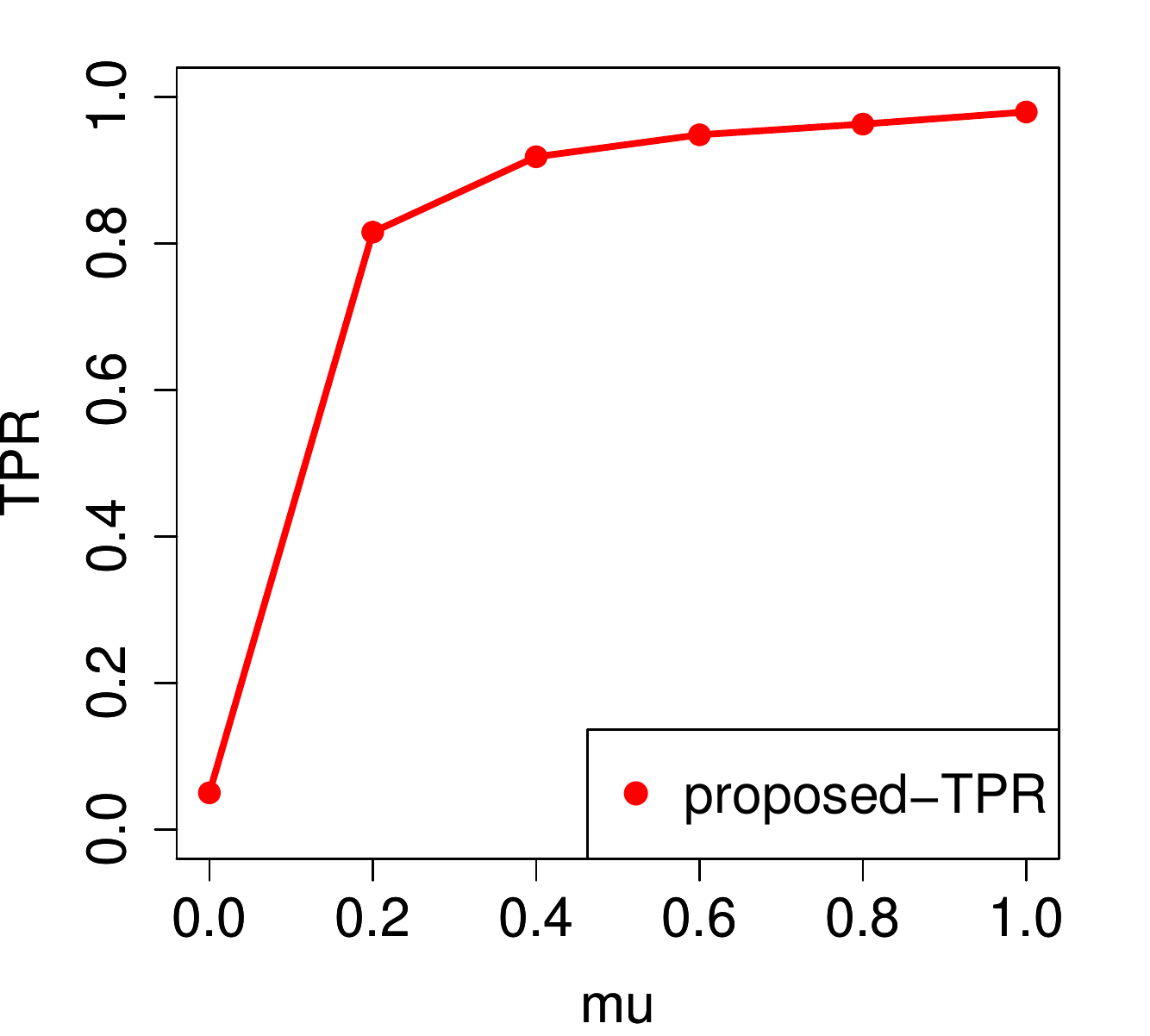}
		\subcaption{TPR for TH} \label{fig: thTPR}
	\end{minipage}
	\caption{
 Results of artificial data experiments using GC- and TH-based segmentation algorithms.
 %
 %The plots a and b show the FPRs and TPRs for GC-based segmentation algorithm, whereas the plots c and d show the FPRs and TPRs for TH-based segmentation algorithm. 
 %
 (a) and (c) show that the FPRs of the proposed method are properly controlled at the desired significance level $\alpha=0.05$ for all image sizes $n$. 
 In contrast, the naive method completely failed to control the FPRs; the degree of failure increased with increasing number of pixels $n$. 
 (b) and (d) show that the proposed method successfully identified the correct segmentation results when the difference between the two regions $\mu$ was large. 
 }
\label{fig: exp}
\end{figure}
Next, in the experiment for TPR, data were randomly generated as $\bm x \sim \mathrm{N}(\bm \mu,0.1^{2} I_{n\times n})$. 
Here, $\bm \mu$ is an $n$-dimensional vector that contains $100 \times 100$ elements whose upper left submatrix with size $50 \times 50$ has a mean value $\mu_S$ and whose remaining values have mean value $\mu_T$. Cases with $\mu = \mu_{S} - \mu_{T} = 0.0,0.2,\ldots,1.0$ were investigated.
%
%the following matrix 
%\begin{align*}
%M=\begin{pmatrix}
%M_{1}&M_{2}\\
%M_{3}&M_{4}
%\end{pmatrix}\in \mathbb{R}^{100\times100}
%\end{align*}
%where $M_{1}=\mu_{S}\cdot I_{50\times50}$, $M_{2}=M_{3}=M_{4}=\mu_{T}\cdot I_{50\times50}$ with 

The results are shown in Figure~\ref{fig: exp}.
Figures~\ref{fig: exp}a-b and c-d show the results for the GC- and TH-based segmentation algorithms, respectively.
As shown in Figures~\ref{fig: exp}a and c, the proposed method controlled the FPRs at the desired significance level, whereas the naive method could not. 
The FPR of the naive method increases with image size $n$ since the deceptive difference in the mean value between the two regions increases. 
Figures~\ref{fig: exp}b and d show that the TPR of the proposed method increases as the difference between the two regions $\mu$ increases. 

\subsection{Experiments using medical images}
In this section, we applied the proposed method and the naive method to pathological images and computed tomography (CT) images.
In the experiments with pathological images, the GC-based segmentation algorithm was employed to extract fibrous tissue regions in pathological tissue specimens.
The quantitative analysis of pathological images is useful for computer-aided diagnosis, and the extraction of specified areas is practically important~\cite{comaniciu2001pathol,ta2009pathol,xu2014weakly}.
The pathological images were obtained by scanning tissue specimens of the spleen and cervical lymph node stained with hematoxylin and eosin. % at Nagoya University Hospital.
%
%As a scanning equipment Aperio ScanScope XT (Leica Biosystems, Germany) was utilized and the glass slides were scanned at 20x maginification.
%
From the scanned whole-slide images, several region-of-interest (ROI) images were manually extracted with and without fibrous regions at 5x magnification.
The GC-based segmentation algorithm was applied to the above images and a significance test was performed for the segmented regions.
Variance was set to $\Sigma= \hat{\sigma}^{2}I_{n\times n}$, where $\hat{\sigma}^{2}$ was estimated from independent data with the maximum likelihood method. 
%
%In this experiment, the seed regions were manually selected, but the effect of the manual selection was not considered when computing $p$-values.
%
Figures~\ref{fig: real} and \ref{fig: real2} show the segmentation results. 
It can be observed that the $p$-values obtained with the proposed method (selective-$p$) are smaller than $\alpha = 0.05$ only when there are actually fibrous regions in the images.
In contrast, the naive method always gives zero $p$-values, even for images that do not contain fibrous regions. 

In experiments with CT images, we aimed to extract the tumor region in the liver.
In CT image analysis, the segmentation of the organ and tumor is practically important~\cite{massoptier2008ct,Ye2010ct,bae1993automatic,boykov2000interactive,gu2013automated,shimizu2007segmentation}.
In the experiments, we used CT images from the 2017 MICCAI Liver Tumor Segmentation Challenge.
Each image is a 3D volume composed of hundreds of 512$\times$512-pixel CT slices.
From such CT volumes, we extracted and cropped 2D slice images that included the liver with and without tumor regions, in which CT values of $-$150 to 250 HU were assigned to the 8-bit grayscale image.
Here, the local TH-based segmentation algorithm was employed for identifying liver tumor regions since CT values in tumor regions are lower than those in surrounding organ regions.
Before applying the local TH algorithm, original images were blurred with Gaussian filtering with a filter size of 11$\times$11.
The parameters for local thresholding were a window size of $50$ and $\theta=1.1$.
The results of local thresholding for CT images are shown in Figures \ref{fig: real3} and \ref{fig: real4}.
It can be observed that the $p$-values obtained with the proposed method (selective-$p$) are smaller than the significance level $\alpha = 0.05$ only when there are actually tumor regions in the images. 
In contrast, the naive method always gives zero $p$-values, even for images that do not contain tumor regions. 
More results are found in the supplement D.

%\begin{table}[t]
%\begin{center}
%\caption{$p$-values calculated by naive and the proposed method for pathological images}\label{table: pvalue}
%\begin{tabular}{|c|c|c|c|c|} \hline
%Image&Fig. \ref{fig: real} \subref{fig: 1}&Fig. \ref{fig: real} \subref{fig: 2}&Fig. \ref{fig: real2} \subref{fig: 3}&Fig. \ref{fig: real2} \subref{fig: 4} \\ \hline
%naive $p$-value &0.00&0.00&0.00&0.00 \\ \hline
%selective $p$-value &0.00&0.00&0.35&0.80 \\ \hline
%\end{tabular}
%\end{center}
%\end{table}

\begin{figure}[t]
\begin{center}
 \begin{minipage}[b]{0.18\hsize}
  \centering
  \includegraphics[width = \hsize]{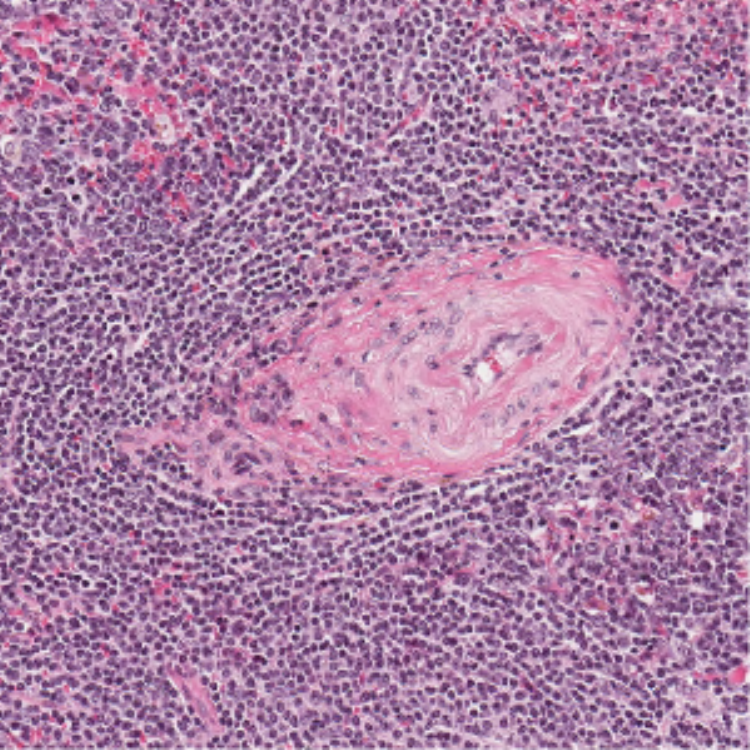}
  \subcaption{Original} \label{fig: 1}
 \end{minipage}
 \begin{minipage}[b]{0.18\hsize}
  \centering
  \includegraphics[width = \hsize]{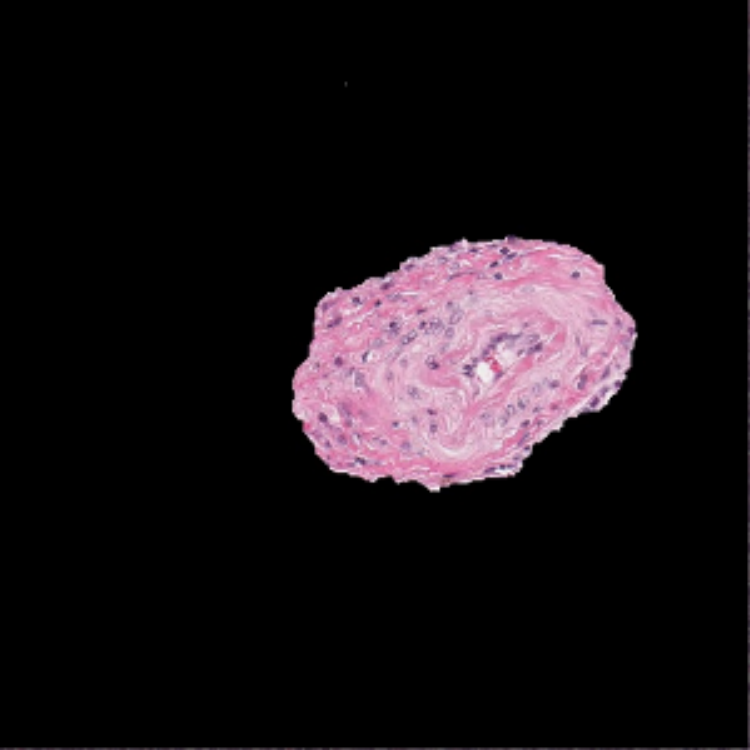}
  \subcaption{Object} \label{fig: 1_obj}
 \end{minipage}
 \begin{minipage}[b]{0.18\hsize}
  \centering
  \includegraphics[width = \hsize]{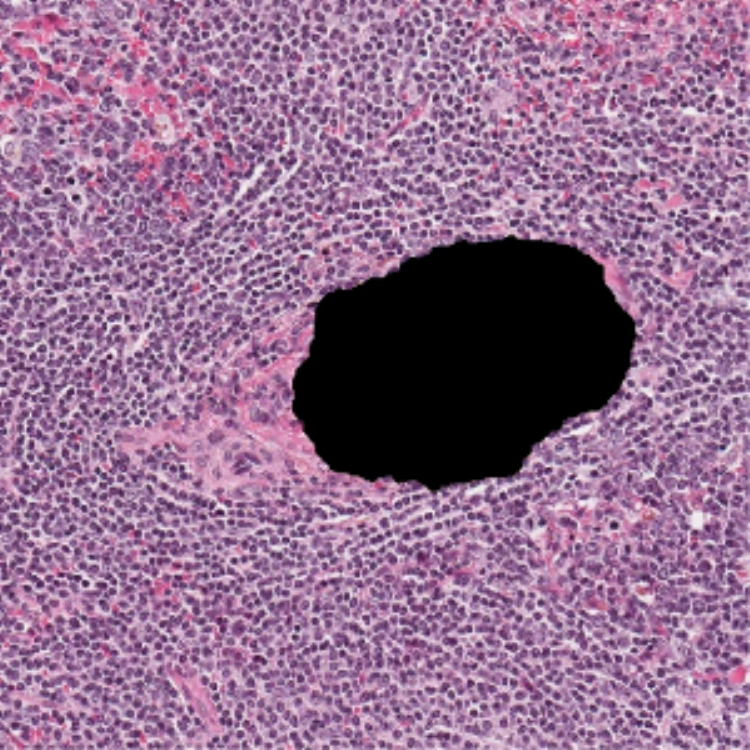}
  \subcaption{Background} \label{fig1_bkg}
 \end{minipage}
 \begin{center}
 \vspace{-5mm}
  (naive-$p$ = {\bf 0.00} and selective-$p$ = {\bf 0.00})
 \end{center}
 
 \begin{minipage}[b]{0.18\hsize}
  \centering
  \includegraphics[width = \hsize]{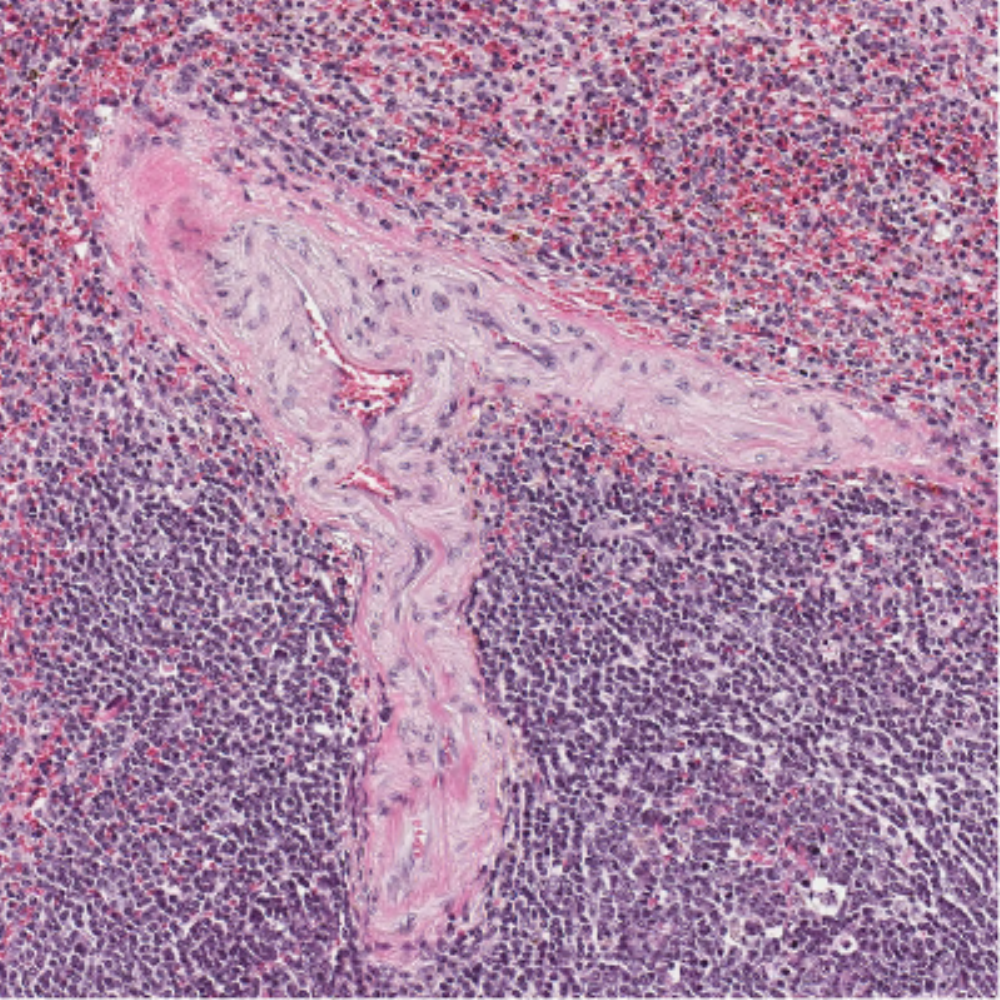}
  \subcaption{Original} \label{fig: 2}
 \end{minipage}
 \begin{minipage}[b]{0.18\hsize}
  \centering
  \includegraphics[width = \hsize]{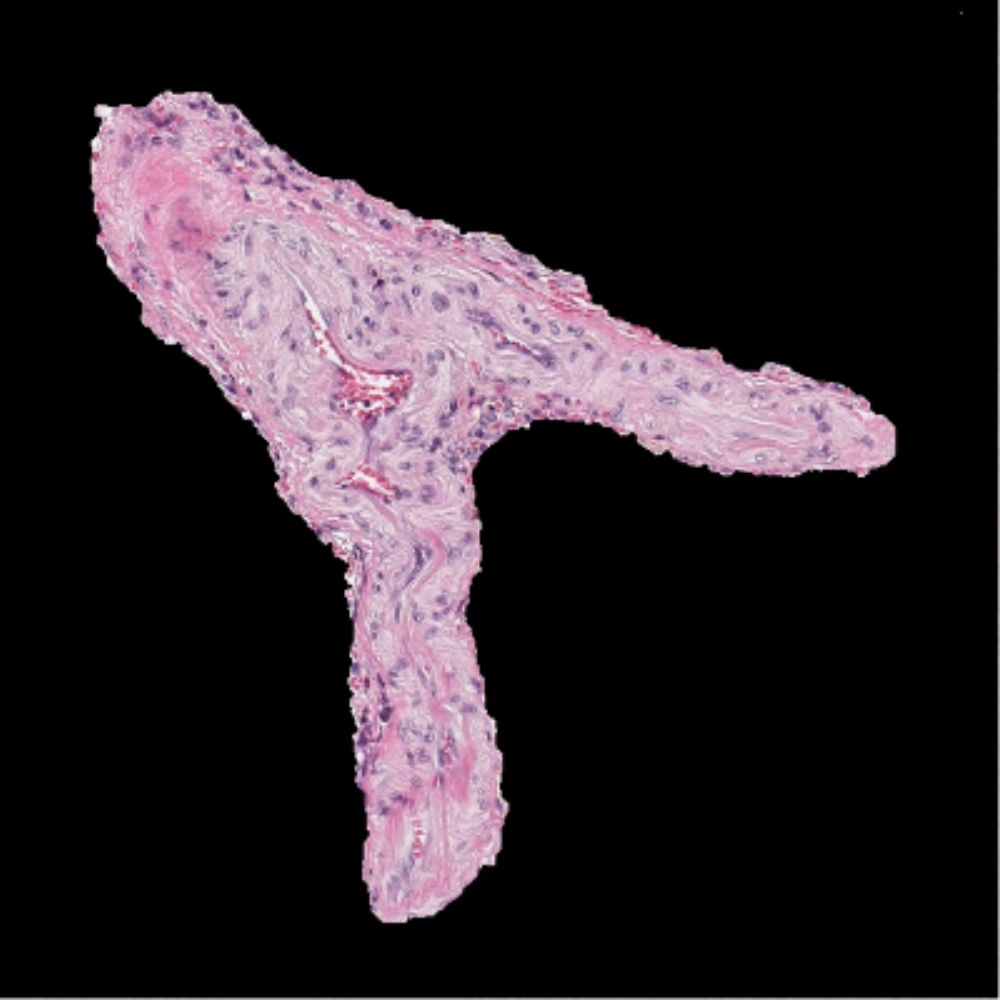}
  \subcaption{Object} \label{fig: 2_obj}
 \end{minipage}
 \begin{minipage}[b]{0.18\hsize}
  \centering
  \includegraphics[width = \hsize]{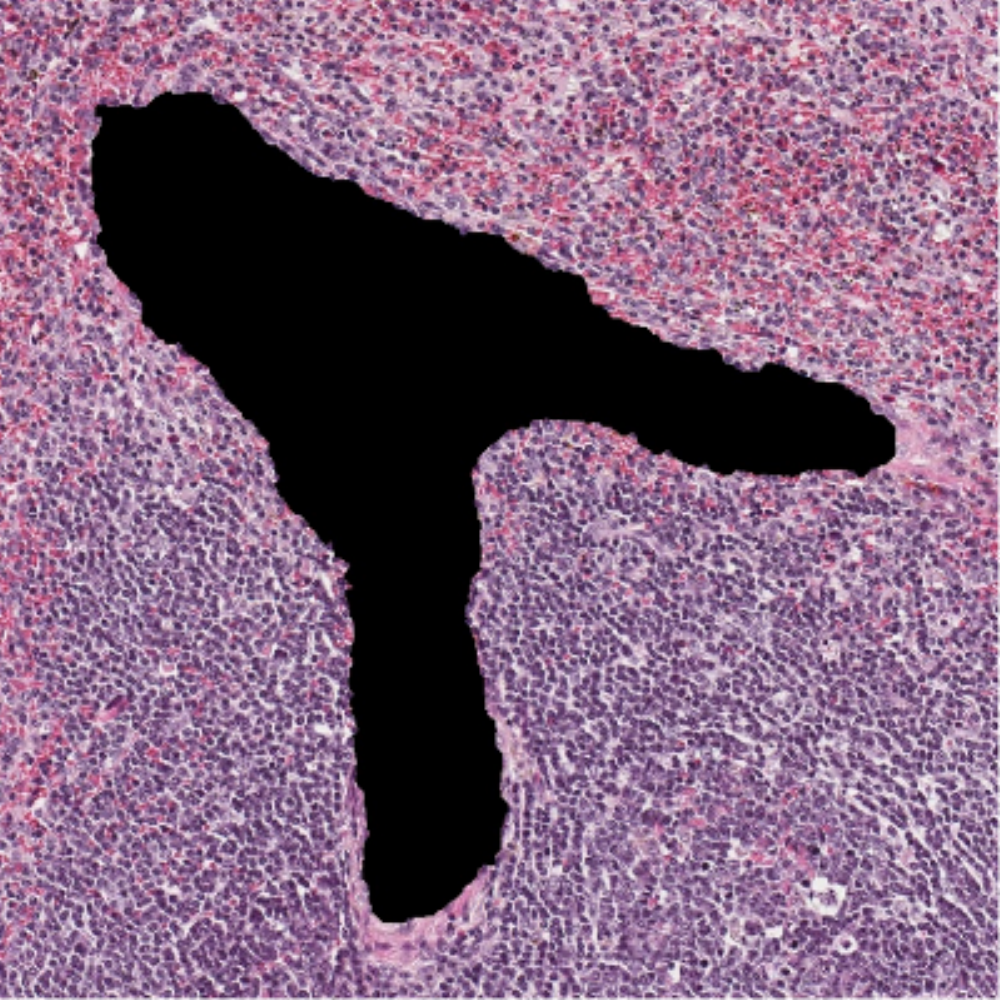}
  \subcaption{Background} \label{fig: 2_bkg}
 \end{minipage}
 \begin{center}
  \vspace{-5mm}
  (naive-$p$ = {\bf 0.00} and selective-$p$ = {\bf 0.00})
 \end{center}
 \caption{
 Segmentation results for pathological images with fibrous regions.
 The $p$-values obtained with the proposed method (selective-$p$) are smaller than $\alpha=0.05$, indicating that these segmentation results correctly identified the fibrous regions. 
 }
 \label{fig: real}

  \hspace*{1mm}

 \begin{minipage}[b]{0.18\hsize}
  \centering
  \includegraphics[width = \hsize]{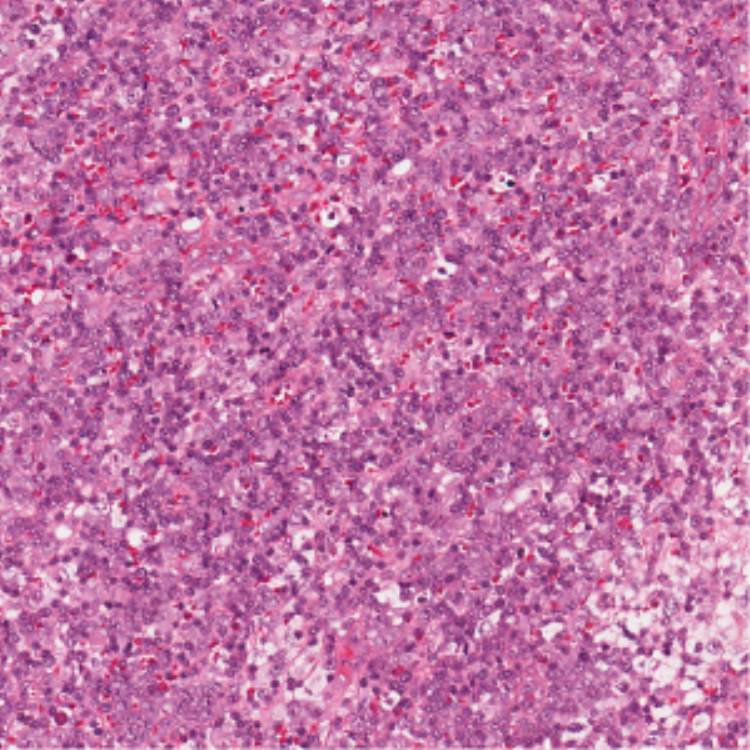}
  \subcaption{Original} \label{fig: 3}
 \end{minipage}
 \begin{minipage}[b]{0.18\hsize}
  \centering
  \includegraphics[width = \hsize]{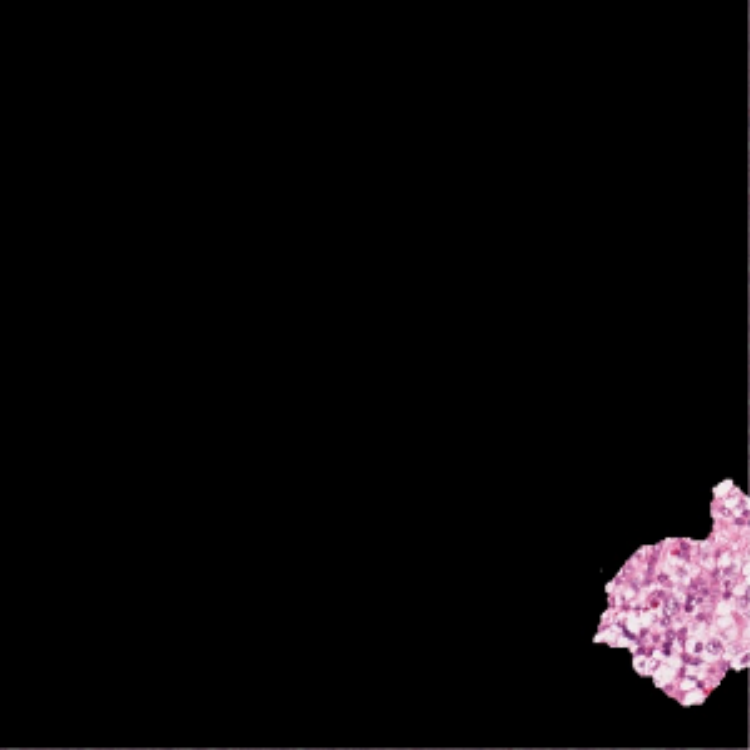}
  \subcaption{Object} \label{fig: 3_obj}
 \end{minipage}
 \begin{minipage}[b]{0.18\hsize}
  \centering
  \includegraphics[width = \hsize]{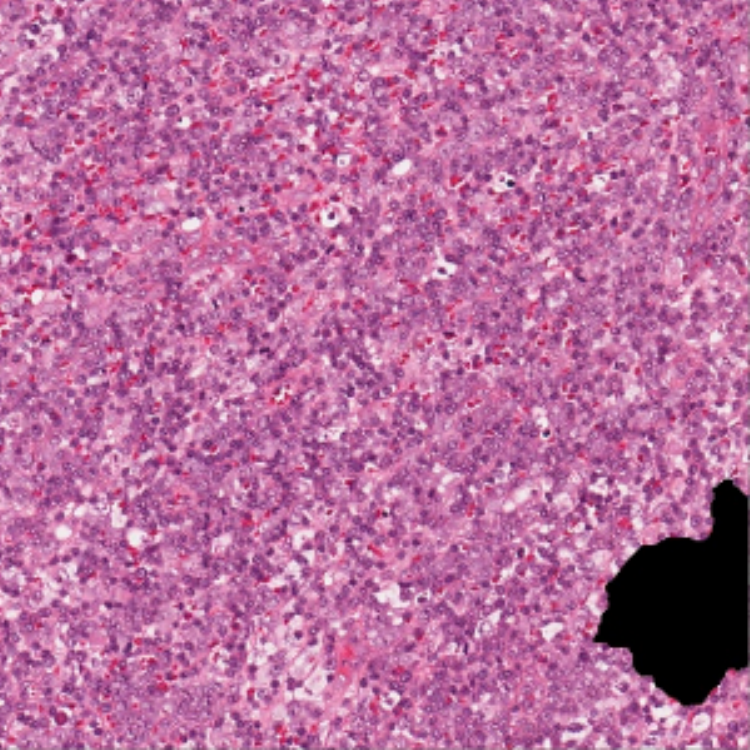}
  \subcaption{Background} \label{fig: 3_bkg}
 \end{minipage}
 \begin{center}
  \vspace{-5mm}
  (naive-$p$ = {\bf 0.00} and selective-$p$ = 0.35)
 \end{center}

 \begin{minipage}[b]{0.18\hsize}
  \centering
  \includegraphics[width = \hsize]{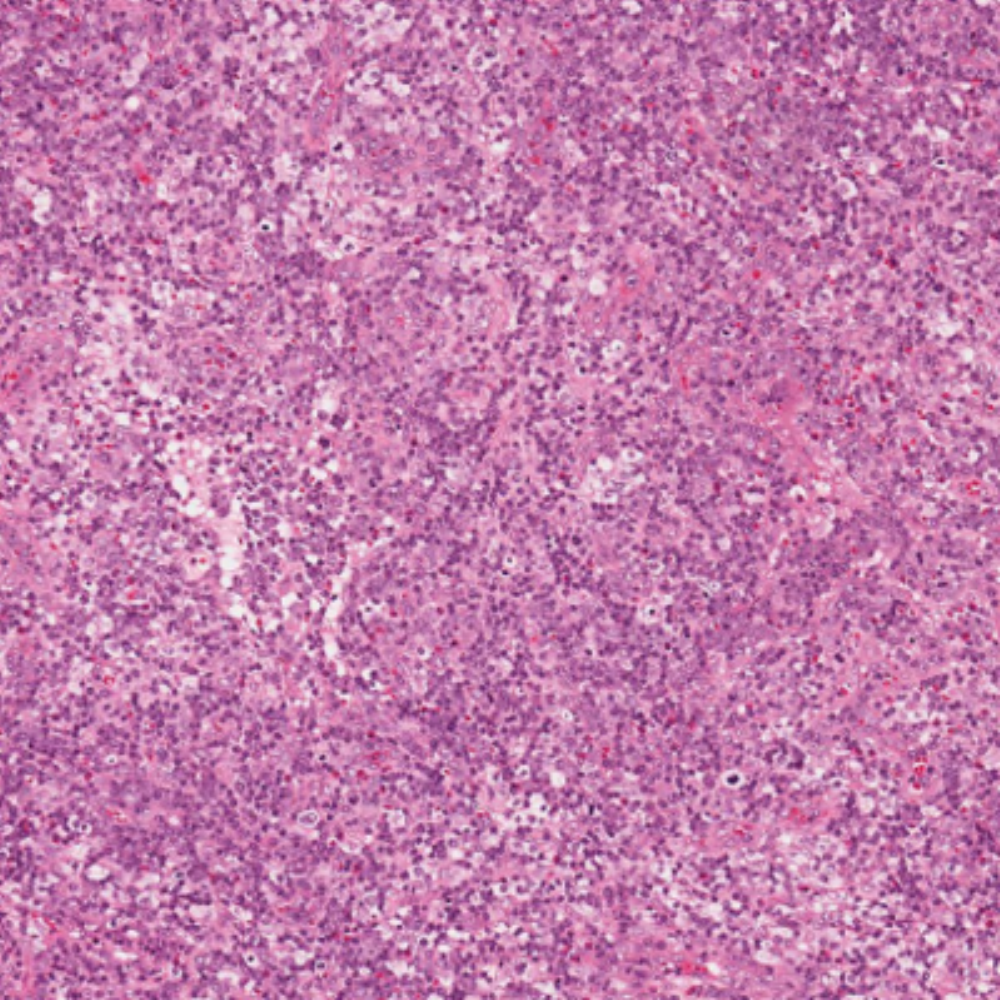}
  \subcaption{Original} \label{fig: 4}
 \end{minipage}
 \begin{minipage}[b]{0.18\hsize}
  \centering
  \includegraphics[width = \hsize]{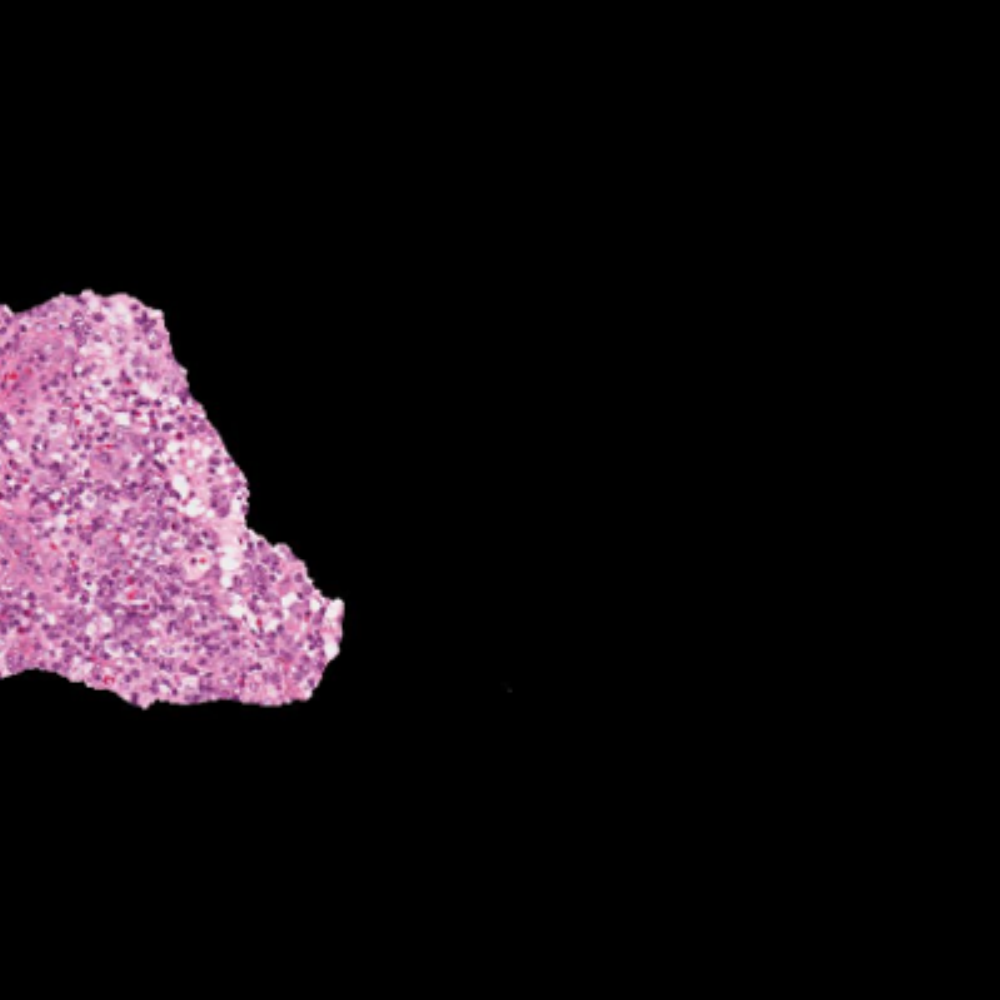}
  \subcaption{Object} \label{fig: 4_obj}
 \end{minipage}
 \begin{minipage}[b]{0.18\hsize}
  \centering
  \includegraphics[width = \hsize]{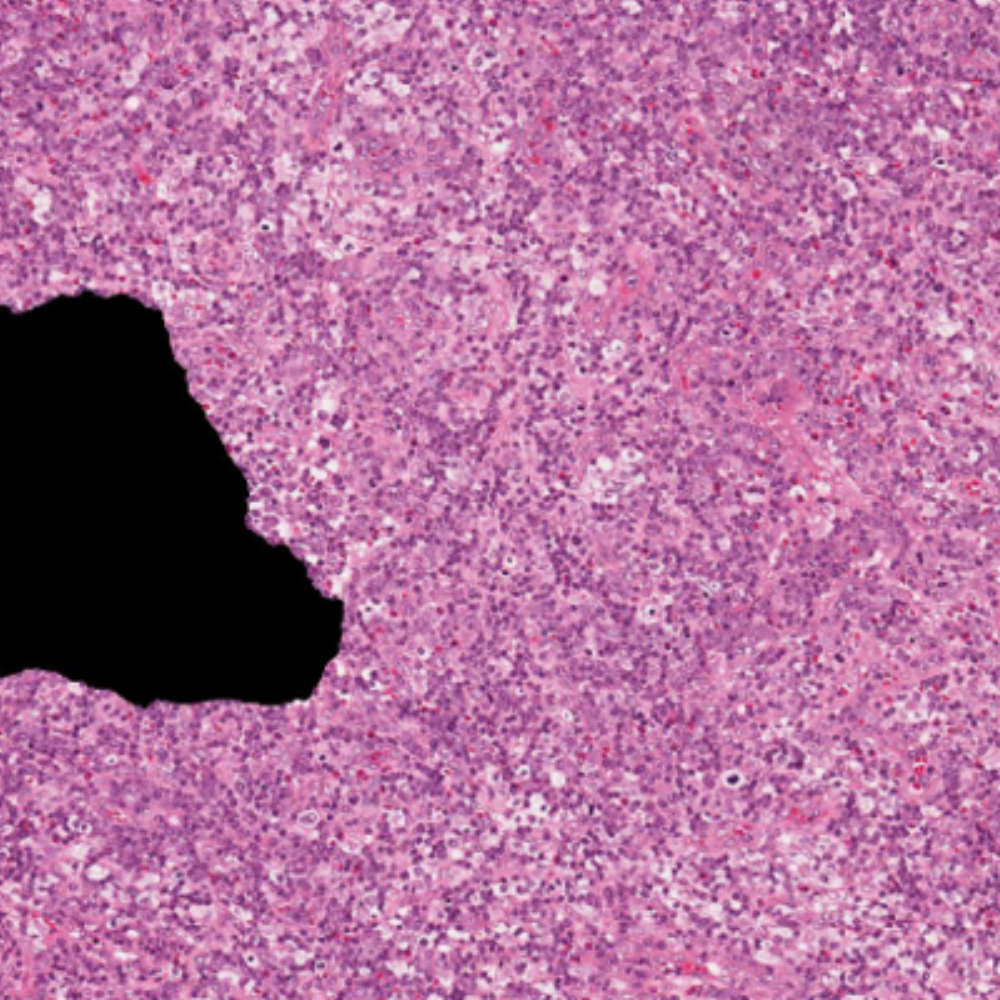}
  \subcaption{Background} \label{fig: 4_bkg}
 \end{minipage}
 \begin{center}
  \vspace{-5mm}
  (naive-$p$ = {\bf 0.00} and selective-$p$ = 0.73)  
 \end{center}
 \caption{
 Segmentation results for pathological images without fibrous regions.
 The $p$-values obtained with the proposed method (selective-$p$) are greater than $\alpha=0.05$, indicating that the differences between the two regions in these images are deceptively large due to segmentation bias. 
 It is obvious that these images do not contain specific objects. 
 }
 \label{fig: real2}
\end{center}
\end{figure}
 
%\begin{table}[t]
%\begin{center}
% \caption{$p$-values calculated by naive and the proposed method for CT images}\label{table: pvalue2}
%\begin{tabular}{|c|c|c|c|c|} \hline
%Image&Fig. \ref{fig: real3} \subref{fig: 5}&Fig. \ref{fig: real3} \subref{fig: 6}&Fig. \ref{fig: real4} \subref{fig: 7}&Fig. \ref{fig: real4} \subref{fig: 8} \\ \hline
%naive $p$-value & 0.00 & 0.00 & 0.00 & 0.00 \\ \hline
%selective $p$-value& 0.00 & 0.00 &0.21 & 0.77 \\ \hline
%\end{tabular}
%\end{center}
%\end{table}

\begin{figure}[t]
\begin{center}
 \begin{minipage}[b]{0.18\hsize}
  \centering
  \includegraphics[width = \linewidth]{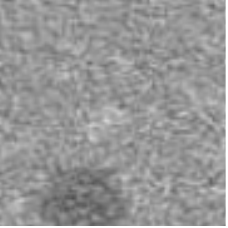}
  \subcaption{Original} \label{fig: 5}
 \end{minipage}
 \begin{minipage}[b]{0.18\hsize}
  \centering
  \includegraphics[width = \linewidth]{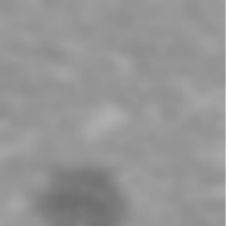}
  \subcaption{Blurred} \label{fig: 5_g}
 \end{minipage}
 \begin{minipage}[b]{0.18\hsize}
  \centering
  \includegraphics[width =\linewidth]{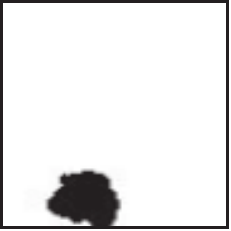}
  \subcaption{Binarized} \label{fig: 5_b}
 \end{minipage}
 \begin{center}
  \vspace{-5mm}
  (naive-$p$ = {\bf 0.00} and selective-$p$ = {\bf 0.00})
 \end{center}
 
 \begin{minipage}[b]{0.18\hsize}
  \centering
  \includegraphics[width = \linewidth]{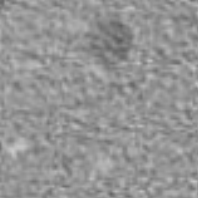}
  \subcaption{Original} \label{fig: 6}
 \end{minipage}
 \begin{minipage}[b]{0.18\hsize}
  \centering
  \includegraphics[width = \linewidth]{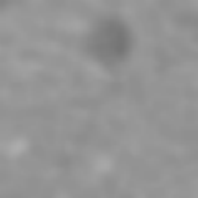}
  \subcaption{Blurred} \label{fig: 6_g}
 \end{minipage}
 \begin{minipage}[b]{0.18\hsize}
  \centering
  \includegraphics[width = \linewidth]{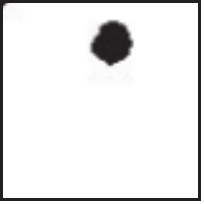}
  \subcaption{Binarized} \label{fig: 6_b}
 \end{minipage}
 \begin{center}
  \vspace{-5mm}
  (naive-$p$ = {\bf 0.00} and selective-$p$ = {\bf 0.00})
 \end{center}
 \caption{
 Segmentation results for CT images with tumor regions.
 The $p$-values obtained with the proposed method (selective-$p$) are smaller than $\alpha=0.05$. These images contain ground-truth tumor regions, which were successfully identified by the segmentation algorithm. 
 }
 \label{fig: real3}

 \hspace*{1mm}
 
 \begin{minipage}[b]{0.18\hsize}
  \centering
  \includegraphics[width = \linewidth]{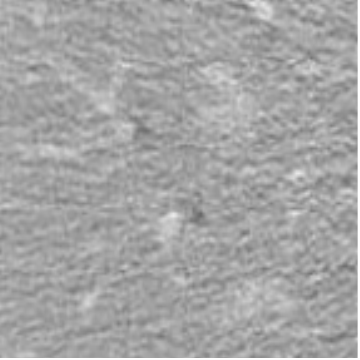}
  \subcaption{Original} \label{fig: 7}
 \end{minipage}
 \begin{minipage}[b]{0.18\hsize}
  \centering
  \includegraphics[width =\linewidth]{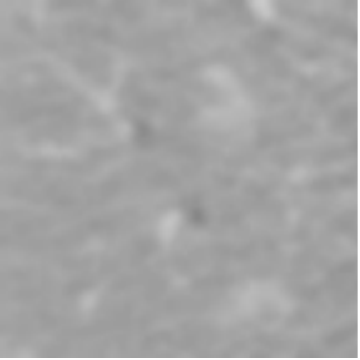}
  \subcaption{Blurred} \label{fig: 7_g}
 \end{minipage}
 \begin{minipage}[b]{0.18\hsize}
  \centering
  \includegraphics[width = \linewidth]{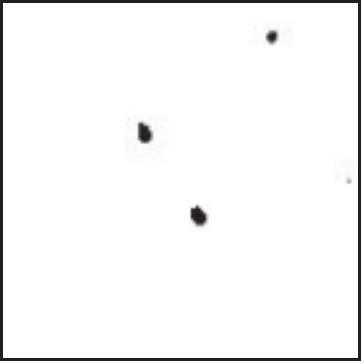}
  \subcaption{Binarized} \label{fig: 7_b}
 \end{minipage}
 \begin{center}
  \vspace{-5mm}
  (naive-$p$ = {\bf 0.00} and selective-$p$ = 0.21)
 \end{center}

 \begin{minipage}[b]{0.18\hsize}
  \centering
  \includegraphics[width = \linewidth]{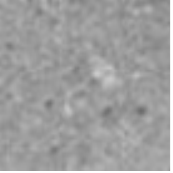}
  \subcaption{Original} \label{fig: 8}
 \end{minipage}
 \begin{minipage}[b]{0.18\hsize}
  \centering
  \includegraphics[width = \linewidth]{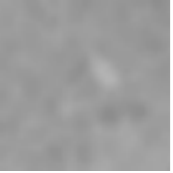}
  \subcaption{Blurred} \label{fig: 8_g}
 \end{minipage}
 \begin{minipage}[b]{0.18\hsize}
  \centering
  \includegraphics[width = \linewidth]{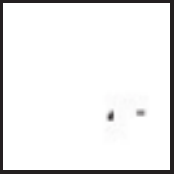}
  \subcaption{Binarized} \label{fig: 8_b}
 \end{minipage}
 \begin{center}
  \vspace{-5mm}
  (naive-$p$ = {\bf 0.00} and selective-$p$ = 0.77)
 \end{center}
 \caption{
 Segmentation results for CT images without tumor regions.
 The $p$-values obtained with the proposed method (selective-$p$) are greater than $\alpha=0.05$.
 These images do not contain any ground-truth tumor regions. 
 The differences between the two regions in these images are deceptively large due to segmentation bias. 
 }
 \label{fig: real4}
\end{center}
\end{figure}

\clearpage
%%%%%%%%%%%%%%%%%%%%%%%%%%%%%%%
\section{Conclusions}
\label{sec:sec5}
In this paper, we proposed a novel framework called PSegI for providing a reliability metric for individual segmentation results by quantifying the statistical significance of the difference between the object and background regions in the form of $p$-values.
Although this problem is challenging due to segmentation bias, we overcome this difficulty by introducing an SI approach. 
%
%We developed methods for computing valid $p$-values for a graph cut-based and threshold-based segmentation algorithms. 
%
%We plan to develop PSegI methods for other segmentation algorithms. 
%%%%%%%%%%%%%%%%%%%%%%%%%%%%%%%

{\small
\bibliographystyle{ieee}
\bibliography{ref}

\begin{thebibliography}{10}\itemsep=-1pt

\bibitem{bae1993automatic}
K.~T. Bae, M.~L. Giger, C.-T. Chen, and C.~E. Kahn.
\newblock Automatic segmentation of liver structure in ct images.
\newblock {\em Medical physics}, 20(1):71--78, 1993.

\bibitem{boykov2006gc}
Y.~Boykov and G.~Funka-Lea.
\newblock Graph cuts and efficient nd image segmentation.
\newblock {\em International journal of computer vision}, 70(2):109--131, 2006.

\bibitem{boykov2000interactive}
Y.~Boykov and M.-P. Jolly.
\newblock Interactive organ segmentation using graph cuts.
\newblock In {\em International conference on medical image computing and
  computer-assisted intervention}, pages 276--286. Springer, 2000.

\bibitem{boykov2004experimental}
Y.~Boykov and V.~Kolmogorov.
\newblock An experimental comparison of min-cut/max-flow algorithms for energy
  minimization in vision.
\newblock {\em IEEE Transactions on Pattern Analysis \& Machine Intelligence},
  (9):1124--1137, 2004.

\bibitem{boykov2001gc}
Y.~Y. Boykov and M.-P. Jolly.
\newblock Interactive graph cuts for optimal boundary \& region segmentation of
  objects in nd images.
\newblock In {\em Proceedings eighth IEEE international conference on computer
  vision. ICCV 2001}, volume~1, pages 105--112. IEEE, 2001.

\bibitem{comaniciu2001pathol}
D.~Comaniciu and P.~Meer.
\newblock Cell image segmentation for diagnostic pathology.
\newblock pages 541--558, 2002.

\bibitem{desolneux2003grouping}
A.~Desolneux, L.~Moisan, and J.-M. More.
\newblock A grouping principle and four applications.
\newblock {\em IEEE Transactions on Pattern Analysis and Machine Intelligence},
  25(4):508--513, 2003.

\bibitem{desolneux2000meaningful}
A.~Desolneux, L.~Moisan, and J.-M. Morel.
\newblock Meaningful alignments.
\newblock {\em International Journal of Computer Vision}, 40(1):7--23, 2000.

\bibitem{desolneux2007gestalt}
A.~Desolneux, L.~Moisan, and J.-M. Morel.
\newblock {\em From gestalt theory to image analysis: a probabilistic
  approach}, volume~34.
\newblock Springer Science \& Business Media, 2007.

\bibitem{dinic1970algorithm}
E.~A. Dinic.
\newblock Algorithm for solution of a problem of maximum flow in networks with
  power estimation.
\newblock In {\em Soviet Math. Doklady}, volume~11, pages 1277--1280, 1970.

\bibitem{elnakib2011survey}
A.~Elnakib, G.~Gimel^^ef^^bf^^bdffarb, J.~S. Suri, and A.~El-Baz.
\newblock Medical image segmentation: a brief survey.
\newblock pages 1--39, 2011.

\bibitem{fithian2014optimal}
W.~Fithian, D.~Sun, and J.~Taylor.
\newblock Optimal inference after model selection.
\newblock {\em arXiv preprint arXiv:1410.2597}, 2014.

\bibitem{ford2009maximal}
L.~R. Ford and D.~R. Fulkerson.
\newblock Maximal flow through a network.
\newblock In {\em Classic papers in combinatorics}, pages 243--248. Springer,
  2009.

\bibitem{goldberg1988new}
A.~V. Goldberg and R.~E. Tarjan.
\newblock A new approach to the maximum-flow problem.
\newblock {\em Journal of the ACM (JACM)}, 35(4):921--940, 1988.

\bibitem{grady2006random}
L.~Grady.
\newblock Random walks for image segmentation.
\newblock {\em IEEE Transactions on Pattern Analysis \& Machine Intelligence},
  (11):1768--1783, 2006.

\bibitem{gu2013automated}
Y.~Gu, V.~Kumar, L.~O. Hall, D.~B. Goldgof, C.-Y. Li, R.~Korn, C.~Bendtsen,
  E.~R. Velazquez, A.~Dekker, H.~Aerts, et~al.
\newblock Automated delineation of lung tumors from ct images using a single
  click ensemble segmentation approach.
\newblock {\em Pattern recognition}, 46(3):692--702, 2013.

\bibitem{hershkovitch2018model}
T.~Hershkovitch and T.~Riklin-Raviv.
\newblock Model-dependent uncertainty estimation of medical image segmentation.
\newblock In {\em 2018 IEEE 15th International Symposium on Biomedical Imaging
  (ISBI 2018)}, pages 1373--1376. IEEE, 2018.

\bibitem{kendall2017uncertainties}
A.~Kendall and Y.~Gal.
\newblock What uncertainties do we need in bayesian deep learning for computer
  vision?
\newblock In {\em Advances in neural information processing systems}, pages
  5574--5584, 2017.

\bibitem{kriegeskorte2009circular}
N.~Kriegeskorte, W.~K. Simmons, P.~S. Bellgowan, and C.~I. Baker.
\newblock Circular analysis in systems neuroscience: the dangers of double
  dipping.
\newblock {\em Nature neuroscience}, 12(5):535, 2009.

\bibitem{lee2016}
J.~D. Lee, D.~L. Sun, Y.~Sun, and J.~E. Taylor.
\newblock Exact post-selection inference, with application to the lasso.
\newblock {\em Annals of Statistics}, 44(3):907--927, 2016.

\bibitem{li2012segCT}
B.~N. Li, C.~K. Chui, S.~Chang, and S.~H. Ong.
\newblock A new unified level set method for semi-automatic liver tumor
  segmentation on contrast-enhanced ct images.
\newblock {\em Expert Systems with Applications}, 39(10):9661--9668, 2012.

\bibitem{li2015segCT}
W.~Li, F.~Jia, and Q.~Hu.
\newblock Automatic segmentation of liver tumor in ct images with deep
  convolutional neural networks.
\newblock {\em Journal of Computer and Communications}, 3(11):146--151, 2015.

\bibitem{liu2007survey}
Y.~Liu, D.~Zhang, G.~Lu, and W.-Y. Ma.
\newblock A survey of content-based image retrieval with high-level semantics.
\newblock {\em Pattern recognition}, 40(1):262--282, 2007.

\bibitem{loftus2015selective}
J.~R. Loftus and J.~E. Taylor.
\newblock Selective inference in regression models with groups of variables.
\newblock {\em arXiv preprint arXiv:1511.01478}, 2015.

\bibitem{massoptier2008ct}
L.~Massoptier and S.~Casciaro.
\newblock A new fully automatic and robust algorithm for fast segmentation of
  liver tissue and tumors from ct scans.
\newblock {\em European radiology}, 18(8):1658, 2008.

\bibitem{otsu1979global}
N.~Otsu.
\newblock A threshold selection method from gray-level histograms.
\newblock {\em IEEE transactions on systems, man, and cybernetics},
  9(1):62--66, 1979.

\bibitem{papavassiliou2010segtxt}
V.~Papavassiliou, T.~Stafylakis, V.~Katsouros, and G.~Carayannis.
\newblock Handwritten document image segmentation into text lines and words.
\newblock {\em Pattern recognition}, 43(1):369--377, 2010.

\bibitem{qin2011image}
K.~Qin, K.~Xu, F.~Liu, and D.~Li.
\newblock Image segmentation based on histogram analysis utilizing the cloud
  model.
\newblock {\em Computers \& Mathematics with Applications}, 62(7):2824--2833,
  2011.

\bibitem{rousseau2008contrario}
F.~Rousseau, F.~Blanc, J.~de~Seze, L.~Rumbach, and J.-P. Armspach.
\newblock An a contrario approach for outliers segmentation: Application to
  multiple sclerosis in mri.
\newblock In {\em 2008 5th IEEE International Symposium on Biomedical Imaging:
  From Nano to Macro}, pages 9--12. IEEE, 2008.

\bibitem{sauvola2000segtxt}
J.~Sauvola and M.~Pietik{\"a}inen.
\newblock Adaptive document image binarization.
\newblock {\em Pattern recognition}, 33(2):225--236, 2000.

\bibitem{sezgin2004survey}
M.~Sezgin and B.~Sankur.
\newblock Survey over image thresholding techniques and quantitative
  performance evaluation.
\newblock {\em Journal of Electronic imaging}, 13(1):146--166, 2004.

\bibitem{shimizu2007segmentation}
A.~Shimizu, R.~Ohno, T.~Ikegami, H.~Kobatake, S.~Nawano, and D.~Smutek.
\newblock Segmentation of multiple organs in non-contrast 3d abdominal ct
  images.
\newblock {\em International journal of computer assisted radiology and
  surgery}, 2(3-4):135--142, 2007.

\bibitem{suzumura2017selective}
S.~Suzumura, K.~Nakagawa, Y.~Umezu, K.~Tsuda, and I.~Takeuchi.
\newblock Selective inference for sparse high-order interaction models.
\newblock In {\em Proceedings of the 34th International Conference on Machine
  Learning-Volume 70}, pages 3338--3347. JMLR. org, 2017.

\bibitem{ta2009pathol}
V.-T. Ta, O.~L{\'e}zoray, A.~Elmoataz, and S.~Sch{\"u}pp.
\newblock Graph-based tools for microscopic cellular image segmentation.
\newblock {\em Pattern Recognition}, 42(6):1113--1125, 2009.

\bibitem{taylor2014exact}
J.~Taylor, R.~Lockhart, R.~J. Tibshirani, and R.~Tibshirani.
\newblock Exact post-selection inference for forward stepwise and least angle
  regression.
\newblock {\em arXiv preprint arXiv:1401.3889}, 7:10--1, 2014.

\bibitem{taylor2018post}
J.~Taylor and R.~Tibshirani.
\newblock Post-selection inference for-penalized likelihood models.
\newblock {\em Canadian Journal of Statistics}, 46(1):41--61, 2018.

\bibitem{taylor2015statistical}
J.~Taylor and R.~J. Tibshirani.
\newblock Statistical learning and selective inference.
\newblock {\em Proceedings of the National Academy of Sciences},
  112(25):7629--7634, 2015.

\bibitem{tian2018selective}
X.~Tian, J.~Taylor, et~al.
\newblock Selective inference with a randomized response.
\newblock {\em The Annals of Statistics}, 46(2):679--710, 2018.

\bibitem{tibshirani1996regression}
R.~Tibshirani.
\newblock Regression shrinkage and selection via the lasso.
\newblock {\em Journal of the Royal Statistical Society: Series B
  (Methodological)}, 58(1):267--288, 1996.

\bibitem{tibshirani2016exact}
R.~J. Tibshirani, J.~Taylor, R.~Lockhart, and R.~Tibshirani.
\newblock Exact post-selection inference for sequential regression procedures.
\newblock {\em Journal of the American Statistical Association},
  111(514):600--620, 2016.

\bibitem{von2012lsd}
R.~G. Von~Gioi, J.~Jakubowicz, J.-M. Morel, and G.~Randall.
\newblock Lsd: a line segment detector.
\newblock {\em Image Processing On Line}, 2:35--55, 2012.

\bibitem{white1983local}
J.~M. White and G.~D. Rohrer.
\newblock Image thresholding for optical character recognition and other
  applications requiring character image extraction.
\newblock {\em IBM Journal of research and development}, 27(4):400--411, 1983.

\bibitem{xu2014weakly}
Y.~Xu, J.-Y. Zhu, I.~Eric, C.~Chang, M.~Lai, and Z.~Tu.
\newblock Weakly supervised histopathology cancer image segmentation and
  classification.
\newblock {\em Medical image analysis}, 18(3):591--604, 2014.

\bibitem{Ye2010ct}
X.~Ye, G.~Beddoe, and G.~Slabaugh.
\newblock Automatic graph cut segmentation of lesions in ct using mean shift
  superpixels.
\newblock {\em Journal of Biomedical Imaging}, 2010:19, 2010.

\bibitem{zhao2017survey}
B.~Zhao, J.~Feng, X.~Wu, and S.~Yan.
\newblock A survey on deep learning-based fine-grained object classification
  and semantic segmentation.
\newblock {\em International Journal of Automation and Computing},
  14(2):119--135, 2017.

\end{thebibliography}
}

%%%%%%%%%%%%%%%%%%%%%%%%%%%%%%%
\appendix
\renewcommand{\thesection}{Appendix~\Alph{section}:}
\renewcommand{\thefigure}{\Alph{section}\arabic{figure}}
\setcounter{figure}{0}
%%%%%%%%%%%%%%%%%%%%%%%%%%%%%%%
\section{Specific examples that any test statistic can be used in our framework under certain conditions} \label{App:A}

The basic assumption of ''large object-background mean intensity difference" equals ''reliable segmentation mask" may be not valid for some segmentation tasks.
For some segmentation tasks, the object and background may have similar or even almost identical intensities.
However, any test statistic can be used in our framework if it is represented as a linear combination of pixel intensities, i.e., test statistics after certain filterbank-based representations such as Fourier, Wavelet, Gabor transforms can be also fit into our framework.

\begin{figure}[t]
 \begin{center}
  \begin{minipage}[t]{0.20\hsize}
   \includegraphics[width = \linewidth]{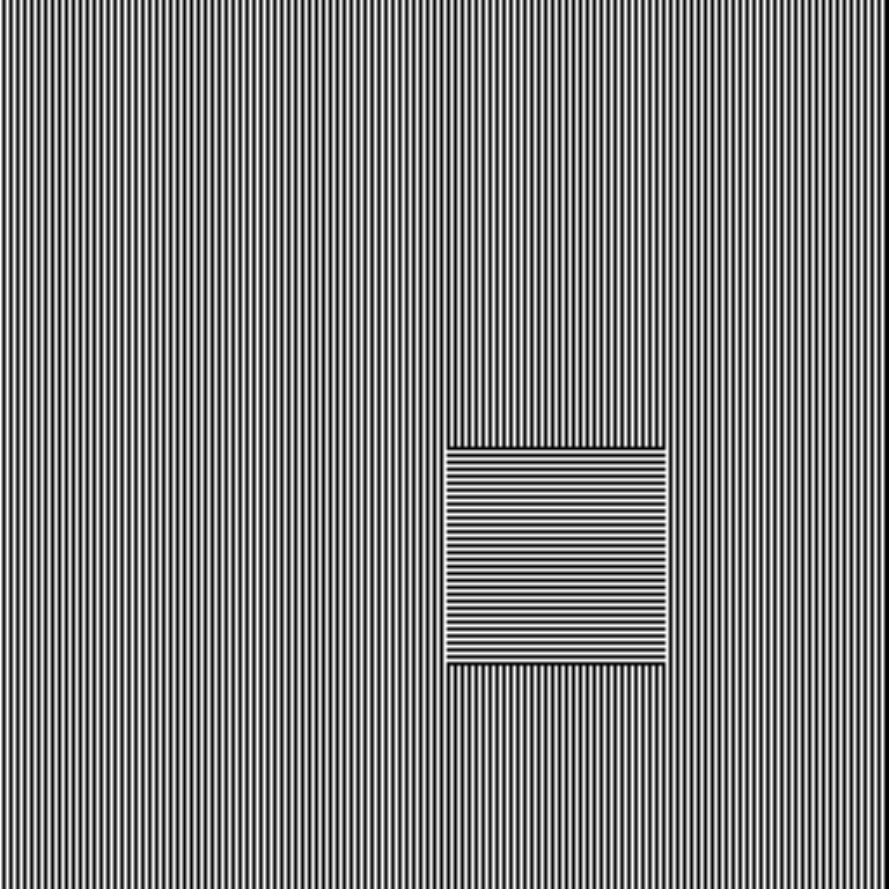}
   \subcaption{Original image} \label{fig: image_filter}
  \end{minipage}
  \begin{minipage}[t]{0.10\hsize}
  \end{minipage}
  \begin{minipage}[t]{0.20\hsize}
   \includegraphics[width =\linewidth]{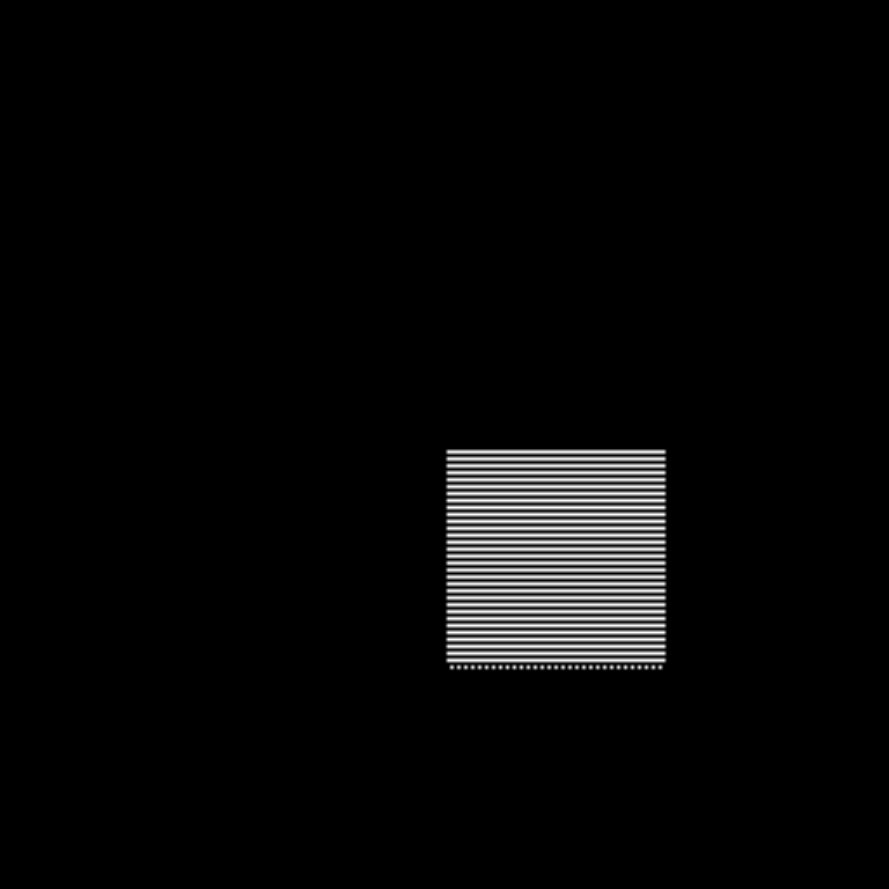}
   \subcaption{Edge extraction}\label{fig: edge}
  \end{minipage}
  \begin{minipage}[t]{0.10\hsize}
  \end{minipage}
  \begin{minipage}[t]{0.20\hsize}
   \includegraphics[width = \linewidth]{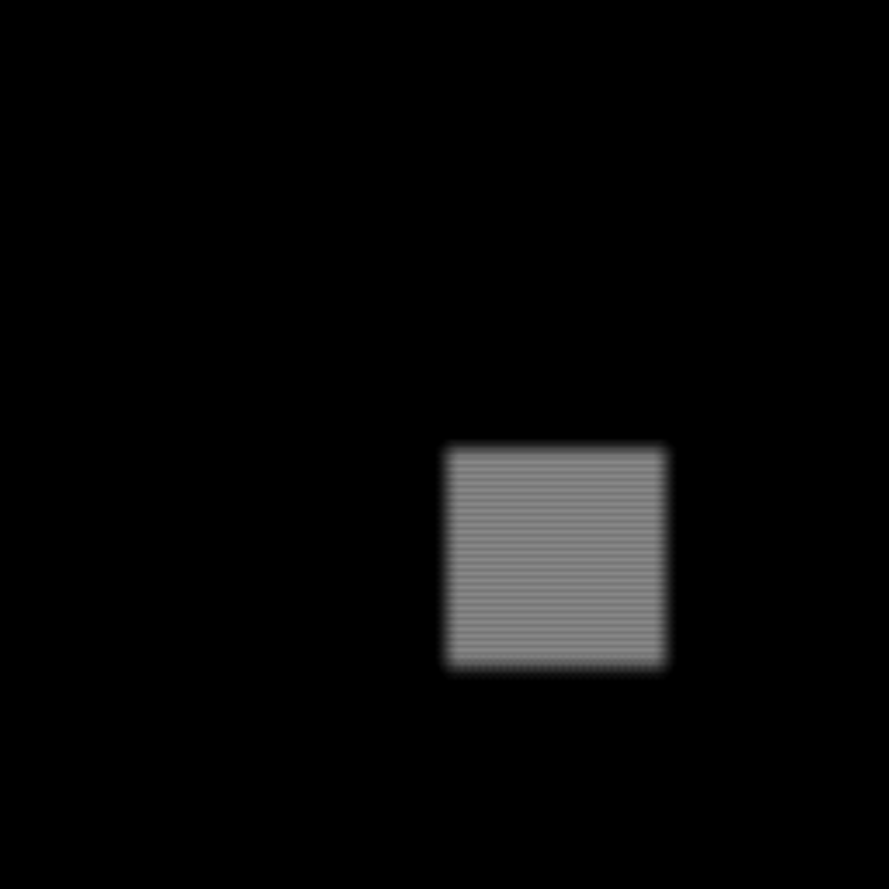}
   \subcaption{Blurred}\label{fig: edge_blur}
  \end{minipage}
  \begin{minipage}[t]{0.10\hsize}
  \end{minipage}
  \begin{minipage}[t]{0.20\hsize}
   \includegraphics[width = \linewidth]{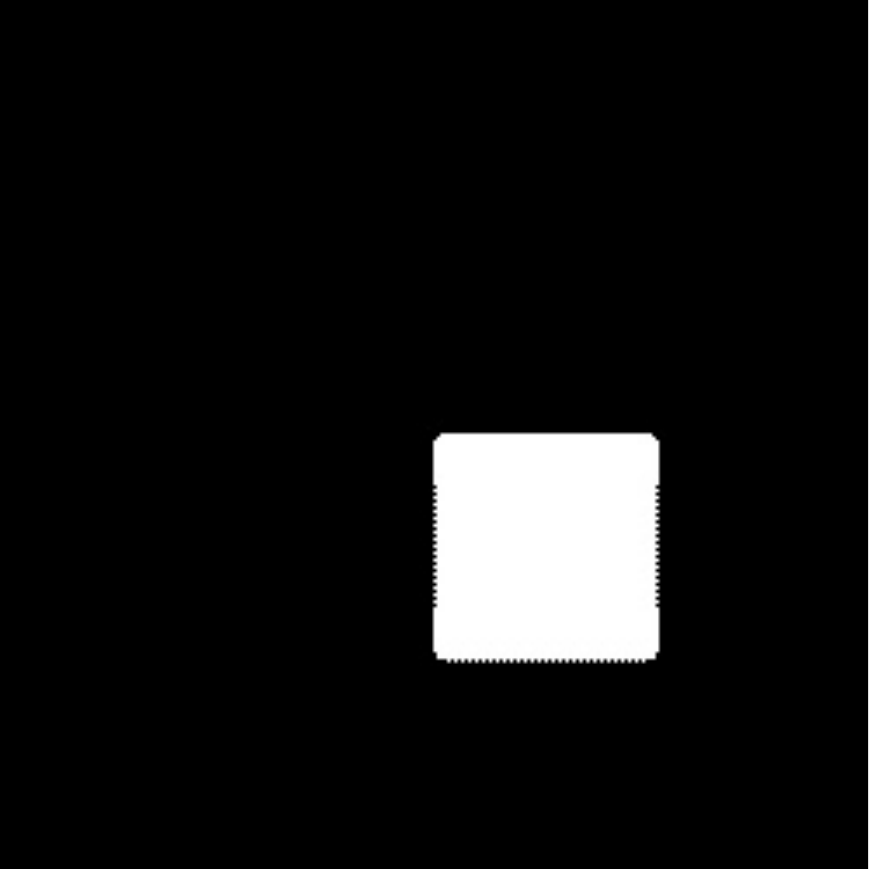}
   \subcaption{Segmentation result}\label{fig: image_seg}
  \end{minipage}
  \caption{
  An example of performing a statistical test using feature values after filter operation.
  (a) There are two regions with different textures, but there is no difference in the average intensity of the two areas.
  (b) The result of applying a first-order differential filter of the y-axis direction to a. 
  (c) The result of applying $5 \times 5$ mean filter to b. 
  (d) The result of applying TH-based segmentation. 
  As a result, selective-$p = {\bf 0.00}$ was obtained.
  }\label{fig: filter}
 \end{center}
\end{figure}

For example, when convolving a certain $3 \times 3$ linear filter that components are represented by $\bm vec(F) := [\phi_1, \ldots, \phi_9]^\top$, (e.g., a first-order differential filter of the y-axis direction $F_1$ is represented by $\bm vec(F_1) := [0,0,0,-1,1,0,0,0,0]^\top$), the image feature after conversion $\bm x'$ is represented by the following.
\begin{align*}
x'_i =& \phi_{1}x_{i-w-1}+\phi_{2}x_{i-w}+\phi_{3}x_{i-w+1} + \phi_{4}x_{i-1}+\phi_{5}x_{i}\\
&~+\phi_{6}x_{i+1} +\phi_{7}x_{i+w-1}+\phi_{8}x_{i+w}+\phi_{9}x_{i+w+1}
\end{align*}
where, $w$ is the width of the image.
We can use this $\bm x'$ instead of $\bm x$. However, note that selection event corresponding to $\bm x$ also changed to $\bm x^\prime$.
Figure~\ref{fig: filter} shows an example that statistical tests are performed using the test statistic after filter transformation in our framework.
By performing the filtering process, two regions can be detected even if there are no differences in average intensities in the original image.
Furthermore, it can be observed that the $p$-values obtained with the proposed method are smaller than $\alpha = 0.05$.
%
%However, We do acknowledge that without ground truth reference, it's hard to evaluate how "reliable" a segmentation mask is. Mean intensity gap between object and background may be one of the most plausible options.

%Although the proposed method is not applicable for any segmentation methods involving nonlinear transformations of pixel data sounds restrictive, many operations can be sufficiently closely approximated by a set of quadratic inequalities, e.g., by quadratic splines (see \S3.2). Furthermore, convolution, max-pooling, and ReLU-based nonlinear transformation etc. in DNN can be fully characterized by a set of quadratic inequalities.

%%%%%%%%%%%%%%%%%%%%%%%%%%%%%%%
\section{Proof of Theorem~1} \label{App:B}
To formally define a valid $p$-value, the difference between random variables and corresponding observations must be clarified. In the rest of this section, for a random variable $a$, $\hat{a}$ is the corresponding observation. 
For notational simplicity, let us write the test statistic as $\Delta = |\delta|$ with $\delta = m_{\rm ob} - m_{\rm bj}$.
Then, the conditional $p$-value for the observed difference $\hat{\Delta}$ in Theorem~1 is formally written as
\begin{align}
 \label{eq:def-pval}
 \mathfrak{p}(\hat{\Delta}) = \PP_{\rm H_0}(\Delta \ge \hat{\Delta} \mid \cA(\bm x) = \{\hat{\cO}, \hat{\cB}\}, {\rm sgn}(\delta) = {\rm sgn}(\hat{\delta})).
\end{align}
By definition, the $p$-value function $\mathfrak{p}(\cdot)$ in \eq{eq:def-pval} should satisfy 
\begin{align}
 \label{eq:pval-property}
 \PP_{\rm H_0}(\mathfrak{p}(\Delta) \le \alpha \mid \cA(\bm x) = \{\hat{\cO}, \hat{\cB}\}, {\rm sgn}(\delta) = {\rm sgn}(\hat{\delta})) = \alpha, ~\forall~ \alpha \in [0,1].
\end{align}
Since the property \eq{eq:pval-property} is satisfied if and only if
\begin{align*}
 \left[ \mathfrak{p}(\Delta) \mid \cA(\bm x) = \{\hat{\cO}, \hat{\cB}\}, {\rm sgn}(\delta) = {\rm sgn}(\hat{\delta}) \right] \sim {\rm Unif}[0, 1],
\end{align*}
we prove the validity of the proposed $p$-value computation method
\begin{align}
 \label{eq:psegI-app}
 {\rm PSegI}\!-\!\mathfrak{p}(\hat{\Delta}) = 1 - F_{0, \bm \eta^\top \Sigma \bm \eta}^{E(\hat{\bm z})}(\hat{\Delta}) 
\end{align}
by showing that 
\begin{align}
 \label{eq:app-goal}
 \left[
 1 - F_{0, \bm \eta^\top \Sigma \bm \eta}^{E(\bm z)}(\Delta)
 \mid
 \cA(\bm x) = \{\hat{\cO}, \hat{\cB}\}, {\rm sgn}(\delta) = {\rm sgn}(\hat{\delta})
 \right]
 \sim {\rm Unif}[0, 1]
\end{align}
in the following proof.

{\it Proof.}
The difference in the average pixel intensities between the object and background regions is written as 
\begin{align*}
\Delta
=
|\delta|
=
|m_{\rm ob} - m_{\rm bg}|
=
\Big|
\frac{1}{|\cO|} \sum_{p \in \cO} x_p
- 
\frac{1}{|\cB|} \sum_{p \in \cB} x_p
\Big|
=
\bm \eta^\top \bm x
\end{align*}
where
\begin{align*}
 \bm \eta
 =
 {\rm sgn}(\delta)
 \left(
 \frac{1}{|\cO|} \bm e_{\cO}
 - 
 \frac{1}{|\cB|} \bm e_{\cB}
\right) 
\end{align*}
Consider a decomposition\footnote{In the case of $\Sigma = I_n$, this decomposition indicates the projection of $\bm x$ to $\bm \eta$ and its orthogonal complement.} of $\bm x$ into two independent components $\bm z$ and $\bm w$ such that 
\begin{align*}
 \bm x = \bm z + \bm w,
 \text{ where }
 \bm z = (I_n - \frac{\Sigma \bm \eta \bm \eta^\top}{\bm \eta^\top \Sigma \bm \eta}) \bm x,
 \text{ and }
 \bm w = \frac{\Sigma \bm \eta \bm \eta^\top}{\bm \eta^\top \Sigma \bm \eta} \bm x.
\end{align*}
Since
$\bm w$
is written as
%\begin{align*}
$
\bm w = \Delta \bm y \text{ with } \bm y = \frac{\Sigma \bm \eta^\top}{\bm \eta^\top \Sigma \bm \eta},
$
%\end{align*}
$\bm x$
is represented as 
\begin{align*}
 \bm x = \bm z + \Delta \bm y.
\end{align*}
Thus, if we fix $\bm z$ to be a certain $\bm z_0$, the quadratic inequality conditions 
\begin{align*}
 \bm x^\top A_j \bm x + \bm b_j^\top \bm x + c_j \le 0, j = 1, 2, \ldots
\end{align*}
specify the range of the test statistic as $\Delta \in E(\bm z_0)$ with
\begin{align*}
 E(\bm z_0) = \cap_j \{\Delta \ge 0 \mid (\bm z_0 + \Delta \bm y)^\top A_j (\bm z_0 + \Delta \bm y) + \bm b_j^\top (\bm z_0 + \Delta \bm y) + c_j \le 0\}.
\end{align*}

This means that the sampling distribution of $\Delta$ conditional on the event
$\cA(\bm x) = \{\hat{\cO}, \hat{\cB}\}$,
${\rm sgn}(\delta) = {\rm sgn}(\hat{\delta})$,
and
$\bm z = \bm z_0$
can be written as
\begin{align}
 \nonumber
 &
 \left[
 \Delta 
 \mid 
 \cA(\bm x) = \{ \hat{\cO}, \hat{\cB} \},  {\rm sgn}(\delta) = {\rm sgn}(\hat{\delta}), \bm z = \bm z_0
 \right]
 \\
 \nonumber
 \stackrel{d}{=}
 &
 \left[
 \Delta 
 \mid 
 \Delta \in E(\bm z_0), \bm z = \bm z_0
 \right]
 \\
 \label{eq:equiv-dist}
 \stackrel{d}{=}
 &
 \left[
 \Delta 
 \mid 
 \Delta \in E(\bm z_0)
 \right],
\end{align}
where
$\stackrel{d}{=}$
denotes equivalence in distribution.
The equivalence on the $2^{\rm nd}$ and $3^{\rm rd}$ lines in \eq{eq:equiv-dist} is from the independence of $\bm z$ and $\Delta$. 

Under the conditions that 
$\cA(\bm x) = \{\hat{\cO}, \hat{\cB}\}$
and
${\rm sgn}(\delta) = {\rm sgn}(\hat{\delta})$,
$\bm \eta$
is considered as a non-random fixed vector, 
and since $\bm x$
is normally distributed, 
$\Delta = \bm \eta^\top \bm x \in E(\bm z_0)$
follows the truncated normal distribution with truncation intervals $E(\bm z_0)$, i.e.,
\begin{align}
 \label{eq:truncated-normal}
 \left[
 \Delta 
 \mid 
 \Delta \in E(\bm z_0)
 \right]
 \sim 
 {\rm TN}(0, \bm \eta^\top \Sigma \bm \eta, E(\bm z_0)),
\end{align}
where
${\rm TN}(\mu, \sigma^2, E)$
indicates the truncated normal distribution with mean $\mu$, variance $\sigma^2$, and truncation intervals $E$. 
From
\eq{eq:equiv-dist}
and 
\eq{eq:truncated-normal},
\begin{align*}
 \left[
 \Delta
 \mid
 \cA(\bm x) = \{\hat{\cO}, \hat{\cB}\},
 {\rm sgn}(\delta) =  {\rm sgn}(\hat{\delta}), 
 \bm z = \bm z_0
 \right]
 \sim 
 {\rm TN}(0, \bm \eta^\top \Sigma \bm \eta, E(\bm z_0)).
\end{align*}
This means that
\begin{align}
 \label{eq:app-unif}
 \left[
 F_{0, \bm \eta^\top \Sigma \bm \eta}^{E(\bm z_0)}(\Delta)
 \mid
 \cA(\bm x) = \{\hat{\cO}, \hat{\cB}\},
 {\rm sgn}(\delta) =  {\rm sgn}(\hat{\delta}), 
 \bm z = \bm z_0
 \right]
 \sim
 {\rm Unif}[0,1],
\end{align}
where
$F_{\mu, \sigma^2}^E$ is the cumulative distribution function of the truncated normal distribution ${\rm TN}(\mu, \sigma^2, E)$. 
By marginalizing over $\bm z$ in \eq{eq:app-unif}, we conclude that 
\begin{align*}
 \left[
 F_{0, \bm \eta^\top \Sigma \bm \eta}^{E(\bm z_0)}(\Delta)
 \mid
 \cA(\bm x) = \{\hat{\cO}, \hat{\cB}\},
 {\rm sgn}(\delta) =  {\rm sgn}(\hat{\delta})
 \right]
 \sim
 {\rm Unif}[0,1].
\end{align*}
This indicates property \eq{eq:app-goal} and hence the validity of the proposed $p$-value computation method in \eq{eq:psegI-app}. 
%%%%%%%%%%%%%%%%%%%%%%%%%
\section{GC-based segmentation event} \label{App:C}
\setcounter{figure}{0}
As stated in \S3.2, the entire process of a maximum flow optimization problem can be decomposed into additions, subtractions, and comparisons of the weights assigned to the edges of the graph.
Thus, if each weight is characterized by quadratic equations and inequalities on the image $\bm x$, the entire segmentation process can be represented by a set of quadratic inequalities in the form of (6) in the main text.

Recall that the graph contains $n + 2$ nodes corresponding to $n$ pixels and two terminal nodes $S$ and $T$.
The weight between two adjacent pixels is determined based on the similarity of their pixel intensities and the distance between them.
Pixel similarity is usually defined based on the properties of the target image.
To provide flexible choice of the similarity function, we employ a quadratic spline approximation, which allows one to specify the desired similarity function with arbitrary approximation accuracy. 
In the experiments in \S4, we used a quadratic spline approximation of the commonly used weight function
\begin{align}
 \label{eq:similarity_function}
 w_{(p,q)} = \exp \left( -\frac{(x_p - x_q)^2}{2 \sigma^2} \right) \frac{1}{{\rm dist}(p, q)}, (p,q) \in \cN,
\end{align}
as shown in Figure C1. 

\begin{figure}[t]
 \begin{center}
  \includegraphics[width = 0.4\linewidth]{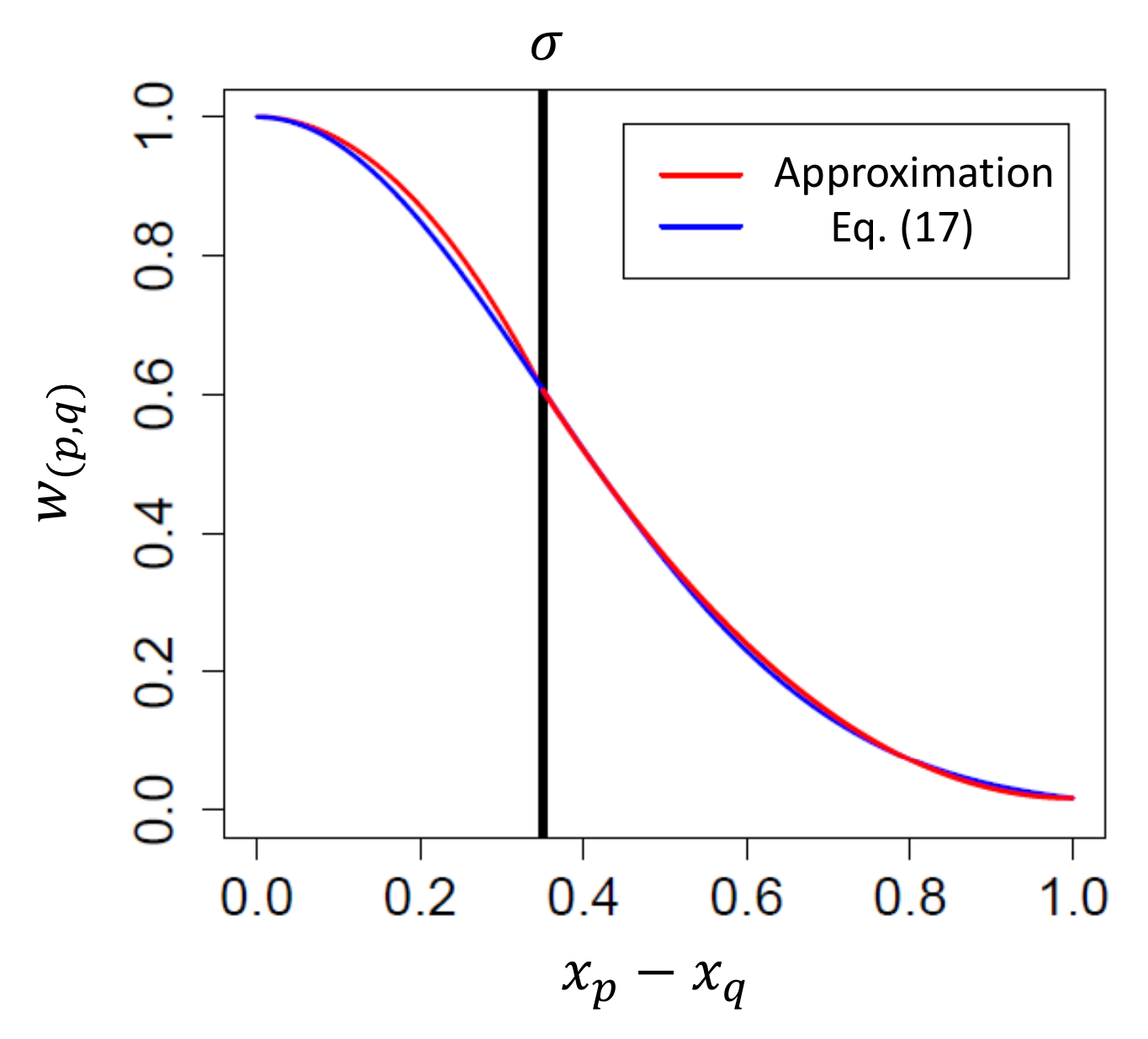}
  \caption{
  Example of a quadratic spline approximation of the commonly used weight function in \eq{eq:similarity_function}.
  }
  \label{fig: spline}
 \end{center}
\end{figure}

In the rest of this section, we demonstrate that for a case with a quadratic spline approximation of \eq{eq:similarity_function}, all the weights in the graph can be characterized by quadratic functions and inequalities on $\bm x$. 
When other similarity functions are used, if an appropriate quadratic spline approximation of the similarity function is employed, similar results can be obtained. 
\begin{itemize}
 \item $(p, q) \in \cN$
       \begin{align}
	\label{eq:gc1}
	w_{(p,q)} = \mycase{
	g_1(x_p - x_q)^2 + h_1, & \text{ if } (x_p - x_q)^2 \le \sigma^2, \\
	g_2(x_q - x_p - 1)^2 + h_2, & \text{ if } (x_p - x_q)^2 > \sigma^2, x_p \le x_q,\\
	g_2(x_p - x_q - 1)^2 + h_2, & \text{ if } (x_p - x_q)^2 > \sigma^2, x_p > x_q,
	}
       \end{align}
       where
       \begin{align*}
	g_1 &= \frac{
	\exp(-\frac{1}{2}) - 1 
	}{
	\sigma^2 {{\rm dist}(p, q)}
	},
	~~~~~ ~~~~~ ~~
	h_1
	=
	\frac{1}{{\rm dist}(p, q)}, 
	\\
	g_2
	&=
	\frac{
	\exp(-\frac{1}{2}) - \exp(-\frac{1}{2\sigma^2})
	}{
	(\sigma-1)^2 {{\rm dist}(p,q)}
	},
	h_2
	=
	\exp\left(
	-\frac{1}{2\sigma^2}
	\right)
	\frac{1}{{\rm dist}(p, q)}.
       \end{align*}
       The first inequality is written as 
       %\begin{align*}
       $
       g_1(x_p - x_q)^2 + h_1 = \bm x^\top A_j \bm x + c_j
       $
       %\end{align*} 
       with 
       %\begin{align*}
       $
	A_j = g_1(\bm e_p - \bm e_q)(\bm e_p - \bm e_q),
	c_j = h_1.
       $
       %\end{align*}
       %
       The second quadratic function is written as 
       %\begin{align*}
       $
	g_2(x_q - x_p - 1)^2 + h_2 = \bm x^\top A_j \bm x + \bm b_j^\top \bm x + c_j
       $
       %\end{align*}
       with 
       %\begin{align*}
       $
	A_j = g_2(\bm e_p - \bm e_q)(\bm e_p - \bm e_q),
	\bm b_j = - 2 g_2 (\bm e_q - \bm e_p),
	c_j = g_2 + h_2.
       $
       %\end{align*}
       %
       The third quadratic function is written as 
       %\begin{align*}
       $
	g_2(x_p - x_q - 1)^2 + h_2 = \bm x^\top A_j \bm x + \bm b_j^\top \bm x + c_j
       $
       %\end{align*}
       with 
       %\begin{align*}
       $
	A_j = g_2(\bm e_p - \bm e_q)(\bm e_p - \bm e_q),
	\bm b_j = - 2 g_2 (\bm e_p - \bm e_q),
	c_j = g_2 + h_2.
       $
       %\end{align*}
       %
       The quadratic inequalities in the condition part are written as
       $
       (x_p - x_q)^2 \le \sigma^2
       \Leftrightarrow
       \bm x^\top A_j \bm x \le 0 
       $
       and 
       $
       (x_p - x_q)^2 > \sigma^2
       \Leftrightarrow
       \bm x^\top A_j \bm x > 0 
       $
       with
       $
       A_j = (\bm e_p - \bm e_q)(\bm e_p - \bm e_q)^\top - \frac{1}{n-1}(I_n - n^{-1} \bm e_\cP \bm e_\cP^\top).
       $
       The linear inequalities in the condition part are written as
       $
       x_p \le x_q
       \Leftrightarrow
       \bm b_j^\top \bm x \le 0
       $
       and 
       $
       x_p > x_q
       \Leftrightarrow
       \bm b_j^\top \bm x > 0
       $
       with
       $
       \bm b_j = \bm e_p - \bm e_q.
       $
       
 \item $p=S, q \in \cP \setminus (\cO^{\rm se} \cup \cB^{\rm se})$
       \begin{align}
	\label{eq:gc2}
	w_{(S,q)} = \lambda \log(2 \pi \sigma_{\rm ob}^2) + \frac{(x_q - m_{\rm ob}^{\rm se})^2}{2 \sigma_{\rm ob}^2}
       \end{align}
       Noting that 
       $m_{\rm ob}^{\rm se}$ is a linear function of $\bm x$
       and
       assuming that 
       $\sigma_{\rm ob}^2$ is known or independently estimated as before, 
       the weight in \eq{eq:gc2} is written as
       $
       \bm x_j^\top A_j \bm x + c_j
       $
       with
       $
       A_j = \frac{\lambda}{2 \sigma_{\rm ob}^2}(\bm e_p - \bm e_{\cO^{\rm se}}/|\cO^{\rm se}|)(\bm e_p - \bm e_{\cO^{\rm se}}/|\cO^{\rm se}|)^\top,
       c_j = \log(2 \pi \sigma_{\rm ob}^2)
       $.
       
 \item $p=S, q \in \cO^{\rm se}$
       \begin{align}
	\label{eq:gc3}
	w_{(S,q)} = 1 + \max_{p \in \cP} \sum_{r:(p,r) \in \cN} w_{(p,r)}.
       \end{align}
       Let
       $k_p = \sum_{r:(p,r) \in \cN} w_{(p,r)}$
       for $p \in \cP$.
       Since
       $k_p$ is the 
       sum of the weights characterized by quadratic functions and inequalities, as in \eq{eq:gc1},
       $k_p$
       is 
       also characterized by quadratic functions and inequalities.
       Noting that
       $k_{\max} = \max_{p \in \cP} k_p$
       is characterized by a set of inequalities 
       $k_{\rm max} \ge k$ for any $k \in \cP \setminus {k_{\rm max}}$, 
       the weight in \eq{eq:gc3},
       i.e.,
       $k_{\max}$,
       is also characterized by quadratic functions and inequalities.
       
 \item $p=S, q \in \cB^{\rm se}$
       \begin{align}
	\label{eq:gc4}
	w_{(S,q)} = 0.
       \end{align}

 \item $p \in \cP \setminus (\cO^{\rm se} \cup \cB^{\rm se}), q=T$
       \begin{align}
	\label{eq:gc5}
	w_{(p,T)} = \lambda \log(2 \pi \sigma_{\rm bg}^2) + \frac{(x_q - m_{\rm bg}^{\rm se})^2}{2 \sigma_{\rm bg}^2}
       \end{align}
       As done for \eq{eq:gc2}, 
       the weight in \eq{eq:gc5} can be written as
       $
       \bm x_j^\top A_j \bm x + c_j
       $
       with
       $
       A_j = \frac{\lambda}{2 \sigma_{\rm bg}^2}(\bm e_p - \bm e_{\cB^{\rm se}}/|\cB^{\rm se}|)(\bm e_p - \bm e_{\cB^{\rm se}}/|\cB^{\rm se}|)^\top,
       c_j = \log(2 \pi \sigma_{\rm bg}^2)
       $.
       
 \item $p \in \cO^{\rm se}, q=T$
       \begin{align}
	\label{eq:gc6}
	w_{(p,T)} = 1 + \max_{q \in \cP} \sum_{r:(r,q) \in \cN} w_{(r,q)}.
       \end{align}
       As done for \eq{eq:gc3}, 
       the weight in \eq{eq:gc6} 
       can be characterized by quadratic functions and inequalities.
       
 \item $p \in \cB^{\rm se}, q=T$
       \begin{align}
	\label{eq:gc7}
	w_{(p,T)} = 0.
       \end{align}
\end{itemize}
%

%%%%%%%%%%%%%%%%%%%%%%%%%
\section{Additional experimental results} \label{App:D}
\renewcommand{\thesection}{\Alph{section}}
\setcounter{figure}{0}

\subsection{Segmentation results for pathological images with fibrous regions}
\begin{figure}[H]
\begin{center}
 \begin{minipage}[b]{0.15\hsize}
  \centering
  \includegraphics[width = \hsize]{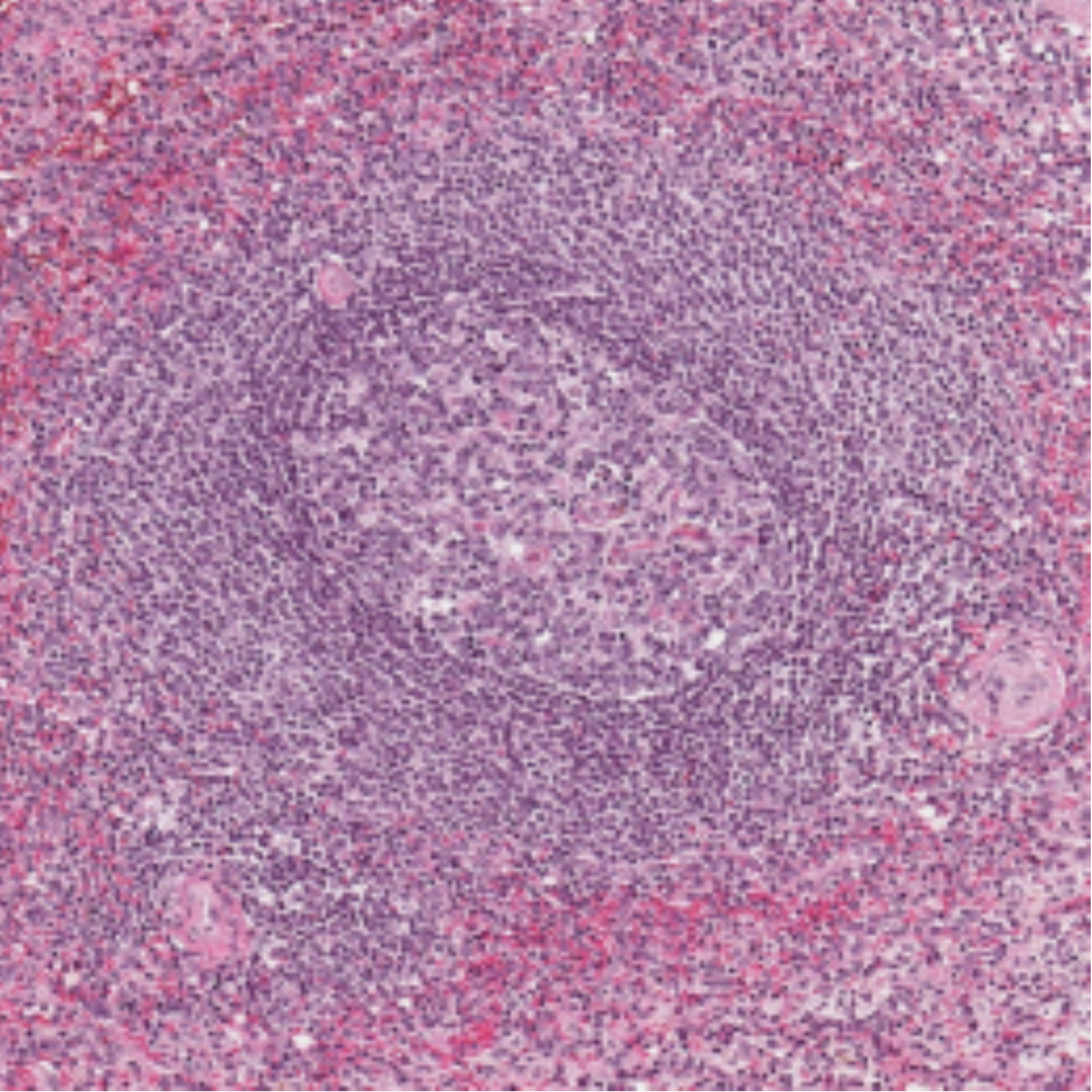}
  \subcaption{Original} \label{fig: 1}
 \end{minipage}
 \begin{minipage}[b]{0.15\hsize}
  \centering
  \includegraphics[width = \hsize]{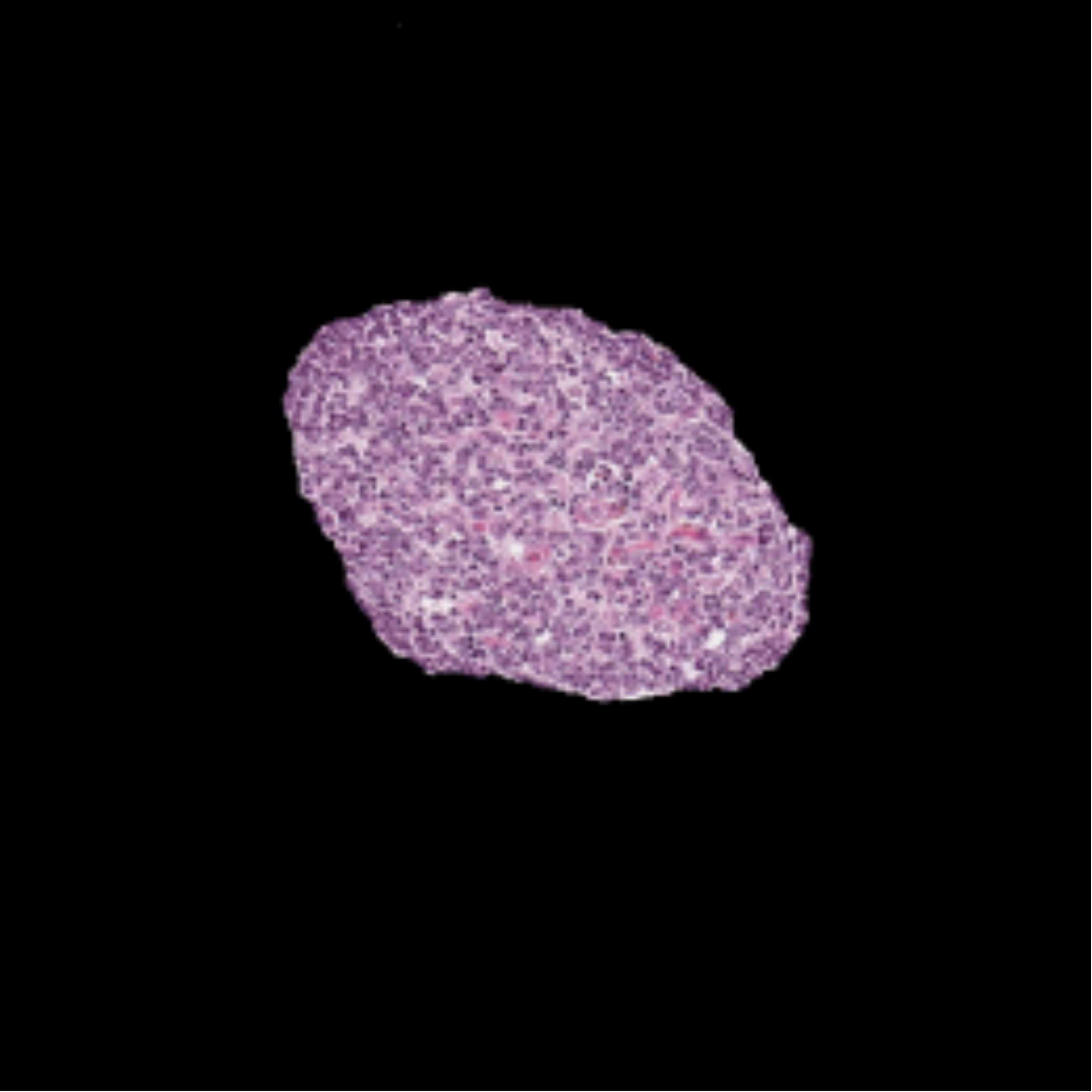}
  \subcaption{Object} \label{fig: 1_obj}
 \end{minipage}
 \begin{minipage}[b]{0.15\hsize}
  \centering
  \includegraphics[width = \hsize]{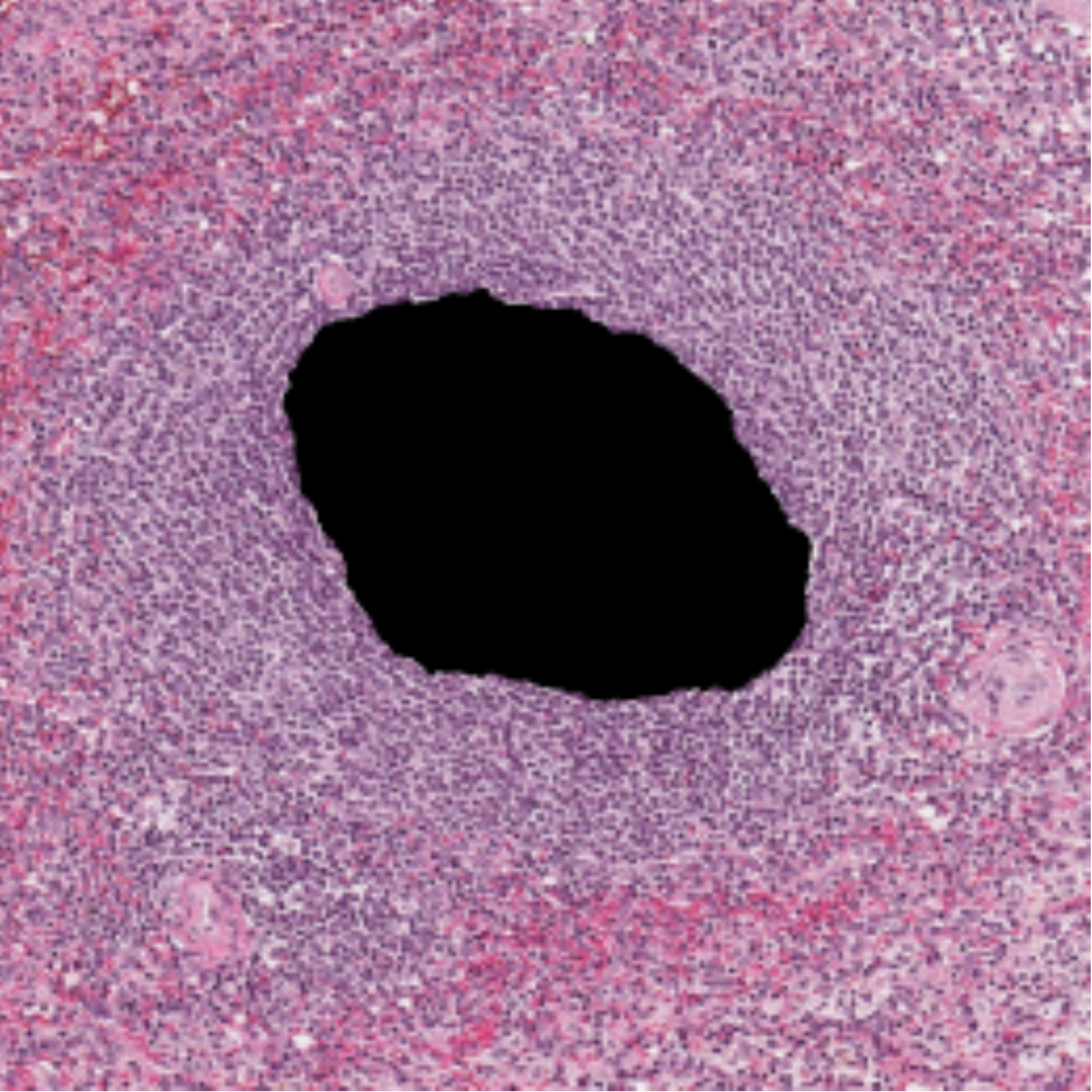}
  \subcaption{Background} \label{fig1_bkg}
 \end{minipage}
 \begin{minipage}[b]{0.5\hsize}
 \end{minipage}
% \begin{center}
%  (naive-$p$ = {\bf 0.00} and selective-$p$ = {\bf 0.00})
% \end{center}   
\begin{minipage}[b]{0.03\hsize}  
\end{minipage}
 \begin{minipage}[b]{0.15\hsize}
  \centering
  \includegraphics[width = \hsize]{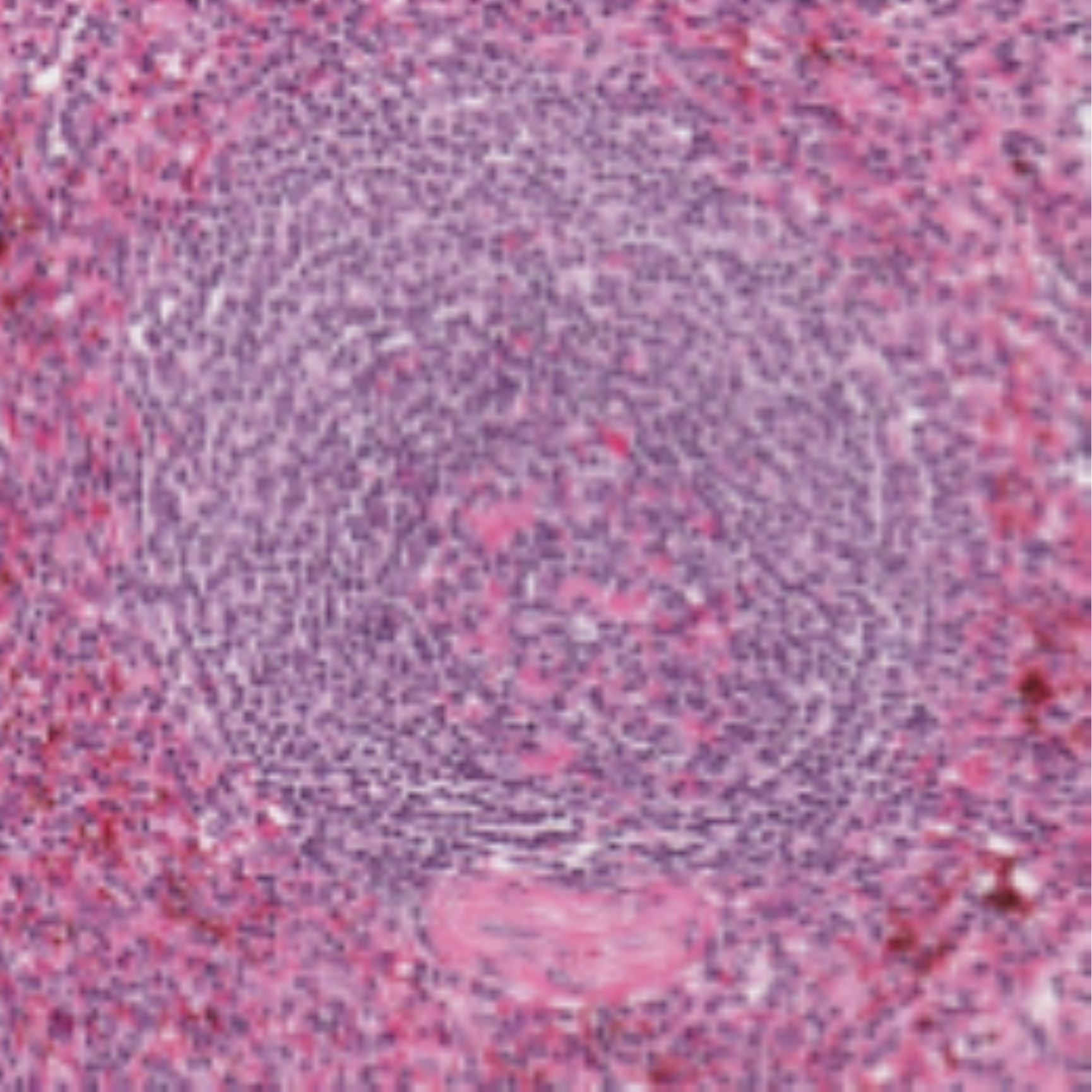}
  \subcaption{Original} \label{fig: 2}
 \end{minipage}
 \begin{minipage}[b]{0.15\hsize}
  \centering
  \includegraphics[width = \hsize]{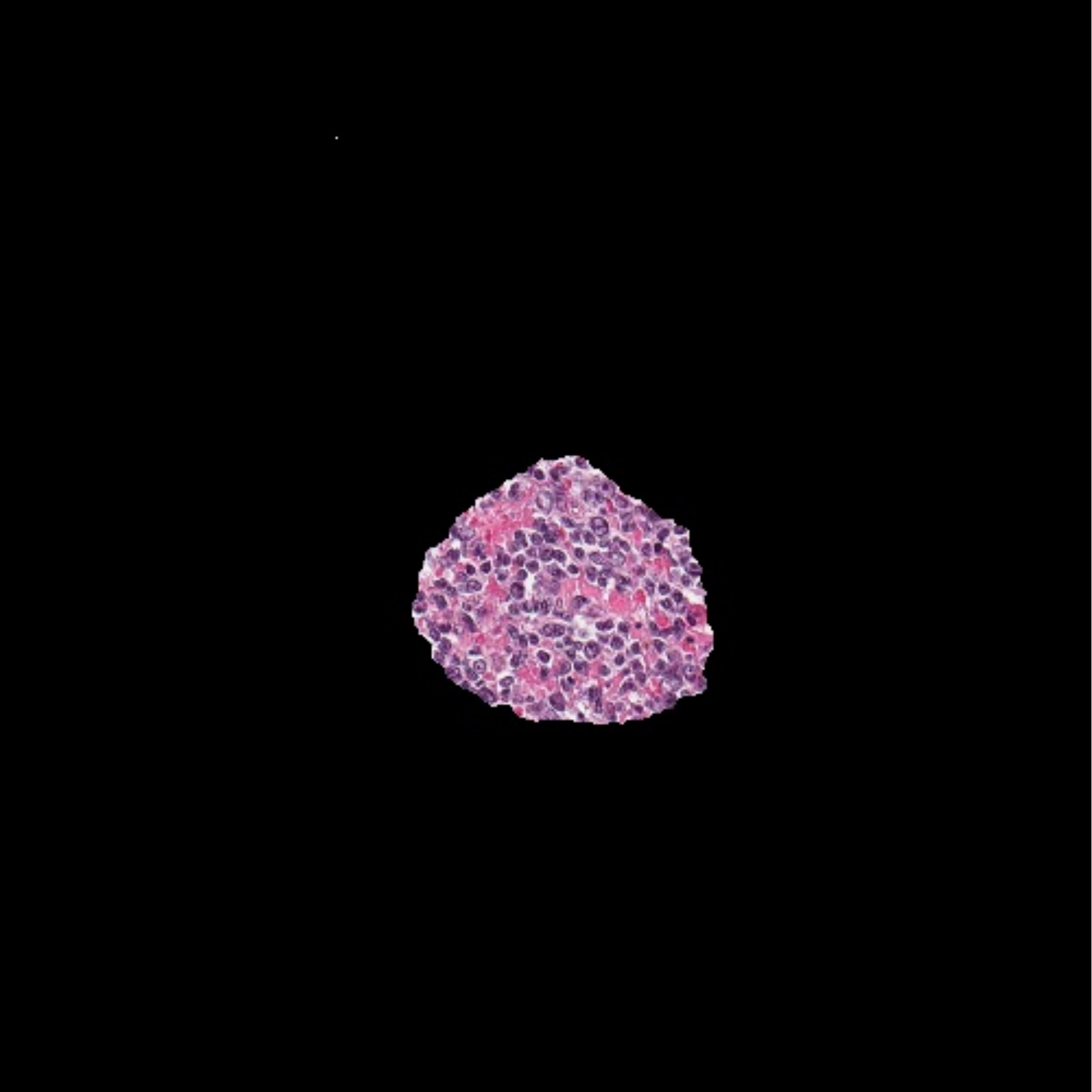}
  \subcaption{Object} \label{fig: 2_obj}
 \end{minipage}
 \begin{minipage}[b]{0.15\hsize}
  \centering
  \includegraphics[width = \hsize]{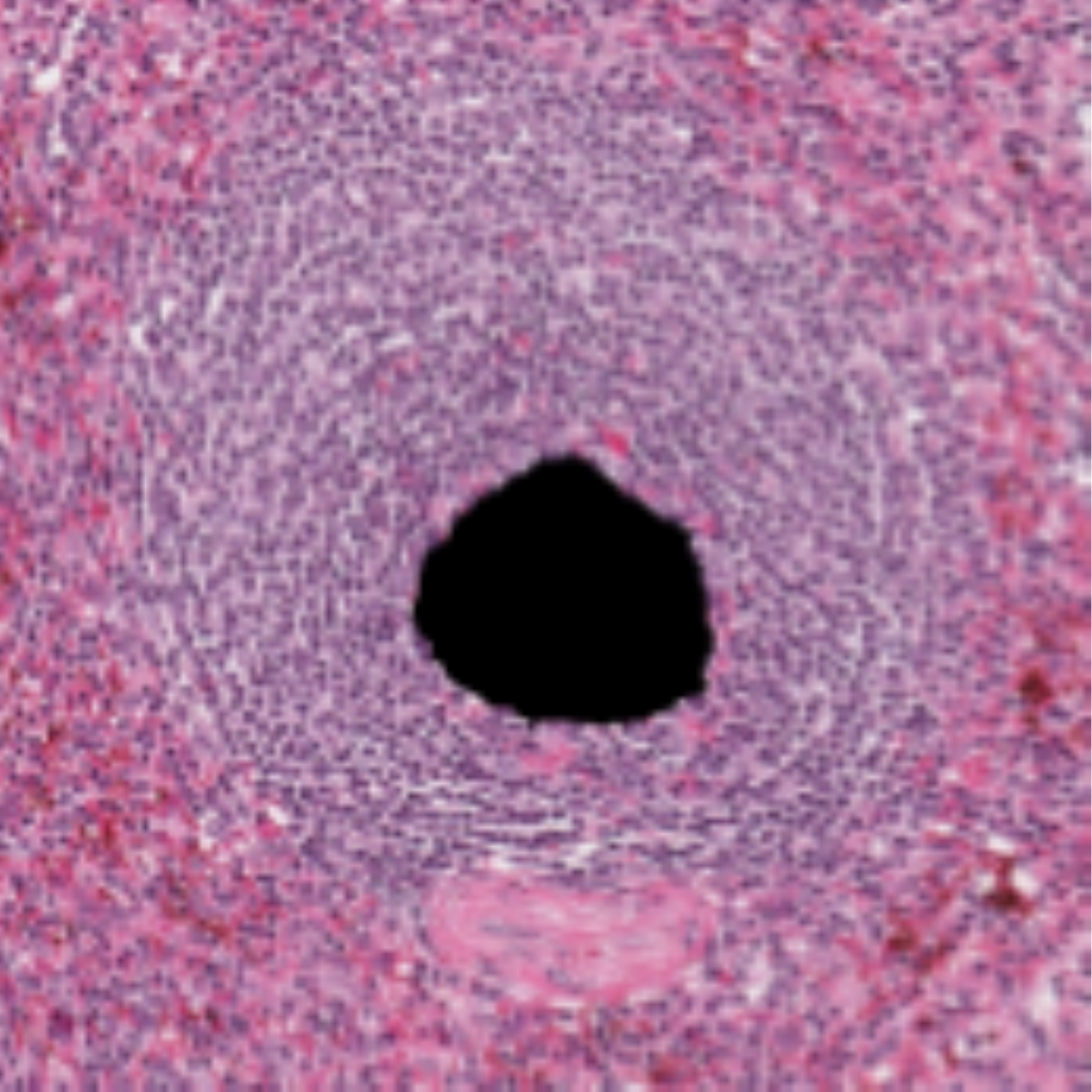}
  \subcaption{Background} \label{fig: 2_bkg}
 \end{minipage}
 \vspace{-3mm}
 \begin{center}
  ~~(naive-$p$ = {\bf 0.00}, selective-$p$ = {\bf 0.00})~~~~~~~~~~~~~~~~~~~~~~~~~~~~~~~~(naive-$p$ = {\bf 0.00}, selective-$p$ = {\bf 0.00})
 \end{center}
% \begin{flushright}
%  (naive-$p$ = {\bf 0.00}, selective-$p$ = {\bf 0.00})
% \end{flushright}
 \begin{minipage}[b]{0.15\hsize}
  \centering
  \includegraphics[width = \hsize]{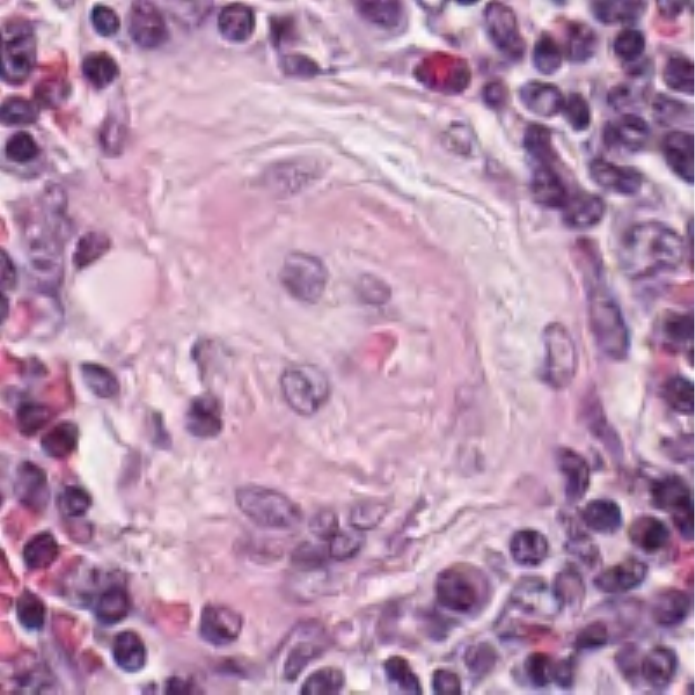}
  \subcaption{Original} \label{fig: 3}
 \end{minipage}
 \begin{minipage}[b]{0.15\hsize}
  \centering
  \includegraphics[width = \hsize]{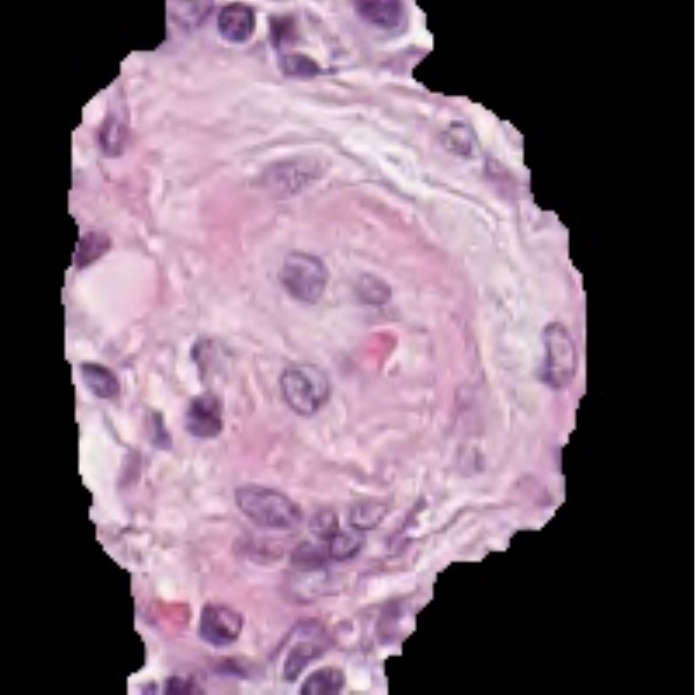}
  \subcaption{Object} \label{fig: 3_obj}
 \end{minipage}
 \begin{minipage}[b]{0.15\hsize}
  \centering
  \includegraphics[width = \hsize]{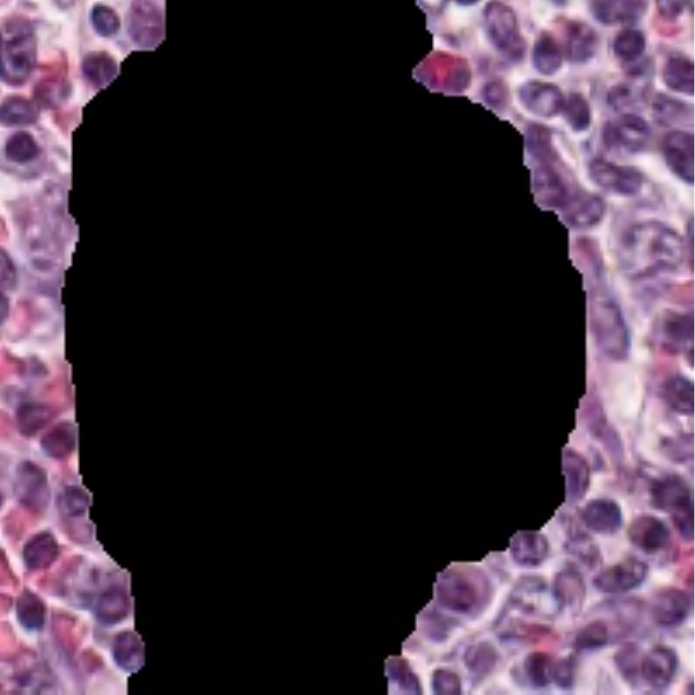}
  \subcaption{Background} \label{fig3_bkg}
 \end{minipage}
 \begin{minipage}[b]{0.5\hsize}
 \end{minipage}
% \begin{center}
%  (naive-$p$ = {\bf 0.00} and selective-$p$ = {\bf 0.00})
% \end{center}   
\begin{minipage}[b]{0.03\hsize}  
\end{minipage}
 \begin{minipage}[b]{0.15\hsize}
  \centering
  \includegraphics[width = \hsize]{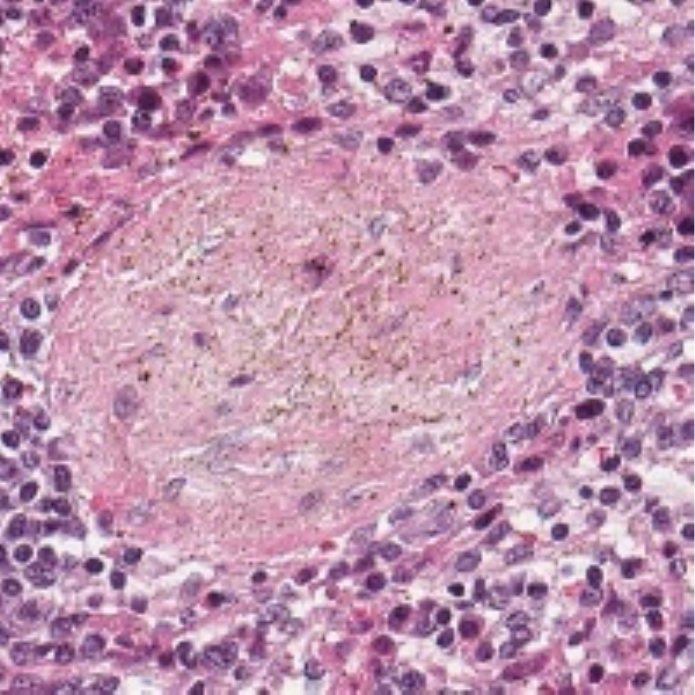}
  \subcaption{Original} \label{fig: 4}
 \end{minipage}
 \begin{minipage}[b]{0.15\hsize}
  \centering
  \includegraphics[width = \hsize]{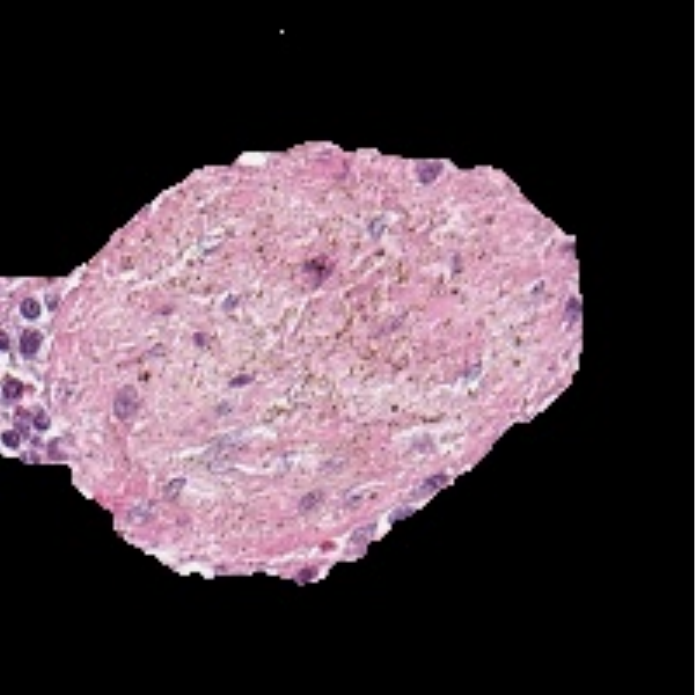}
  \subcaption{Object} \label{fig: 4_obj}
 \end{minipage}
 \begin{minipage}[b]{0.15\hsize}
  \centering
  \includegraphics[width = \hsize]{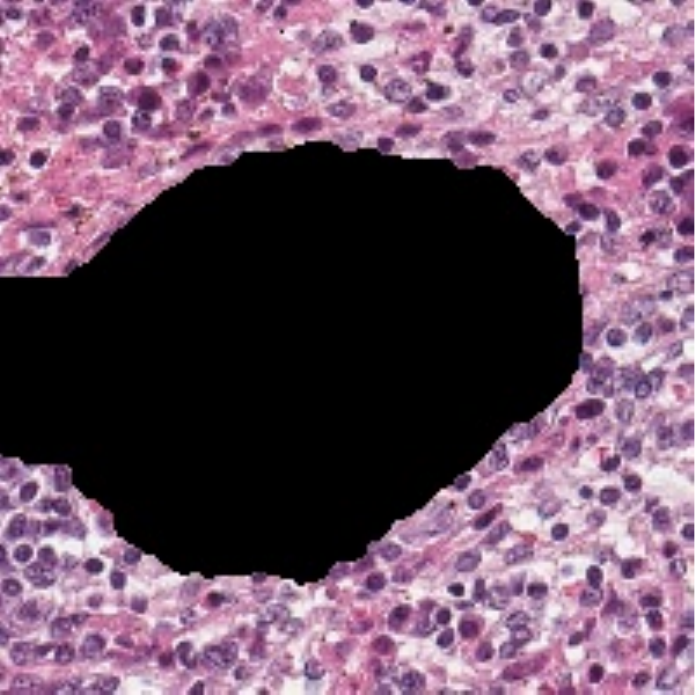}
  \subcaption{Background} \label{fig: 4_bkg}
 \end{minipage}
 \vspace{-3mm}
 \begin{center}
  ~~(naive-$p$ = {\bf 0.00}, selective-$p$ = 0.93)~~~~~~~~~~~~~~~~~~~~~~~~~~~~~~~~(naive-$p$ = {\bf 0.00}, selective-$p$ = 0.72)
 \end{center}
 
  \begin{minipage}[b]{0.15\hsize}
  \centering
  \includegraphics[width = \hsize]{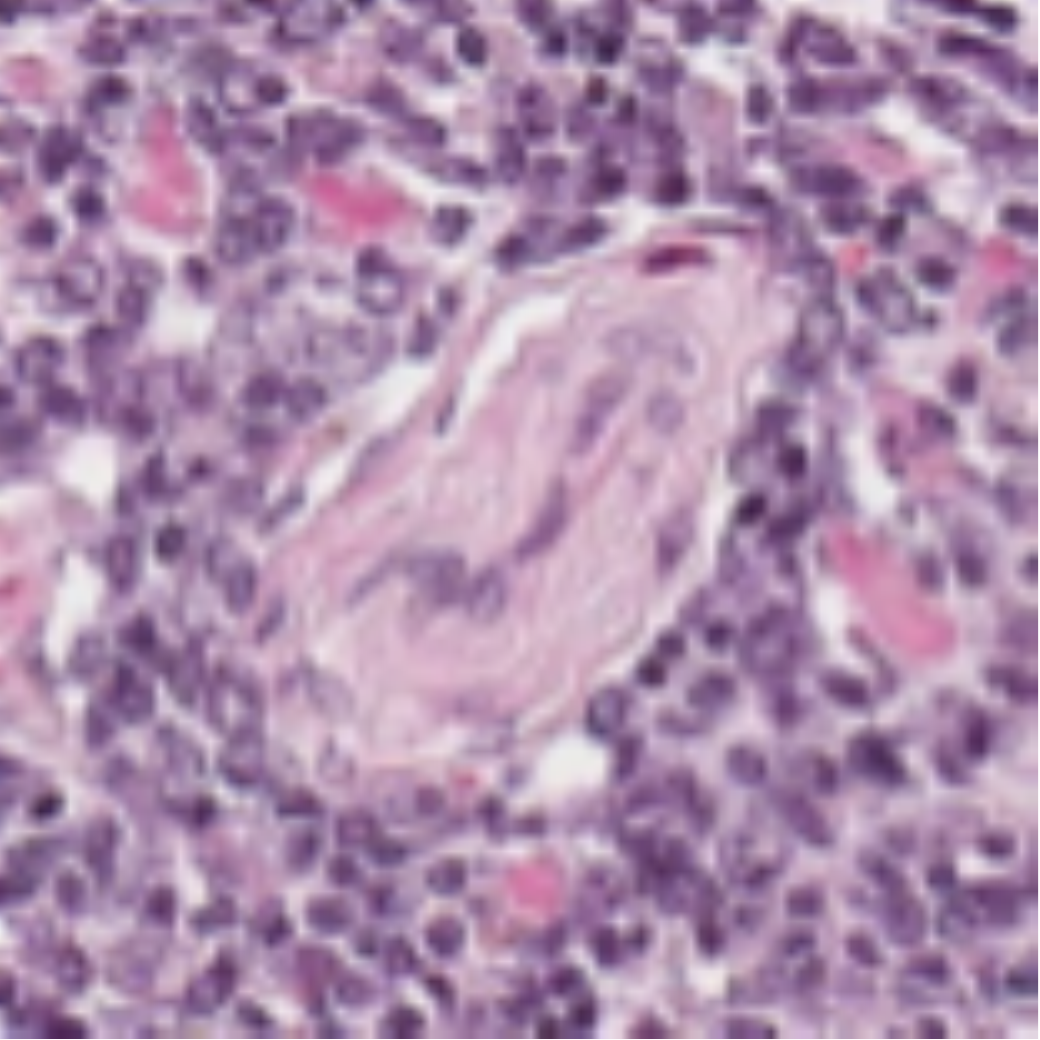}
  \subcaption{Original} \label{fig: 5}
 \end{minipage}
 \begin{minipage}[b]{0.15\hsize}
  \centering
  \includegraphics[width = \hsize]{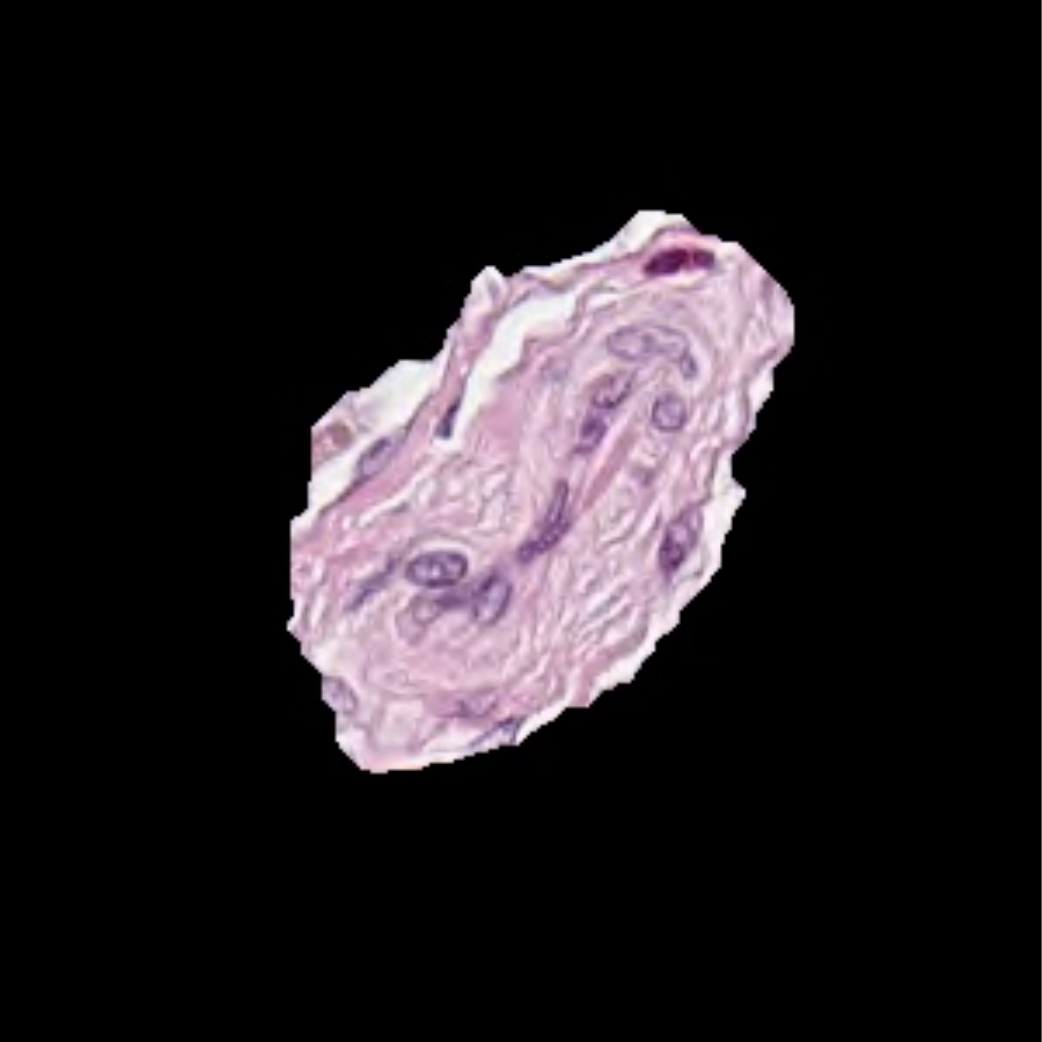}
  \subcaption{Object} \label{fig: 5_obj}
 \end{minipage}
 \begin{minipage}[b]{0.15\hsize}
  \centering
  \includegraphics[width = \hsize]{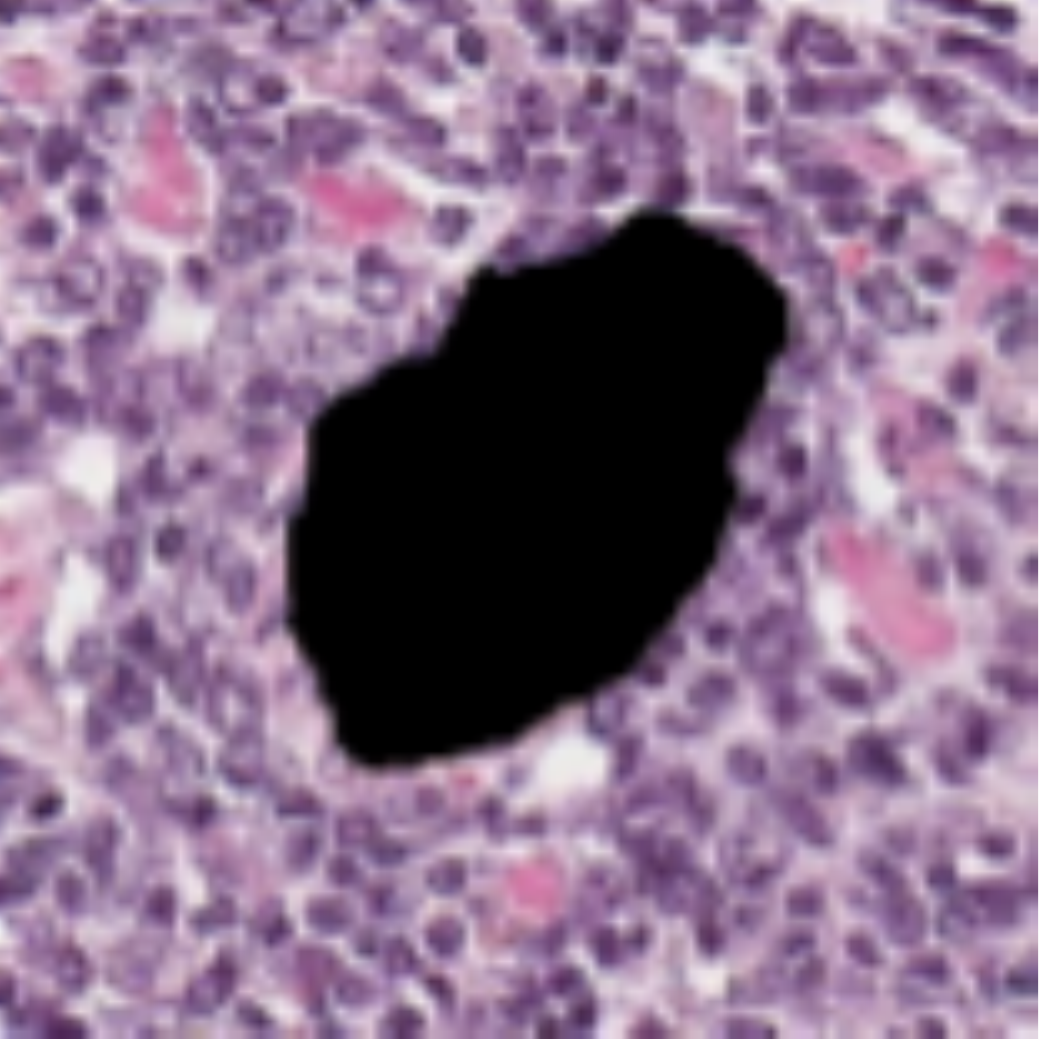}
  \subcaption{Background} \label{fig5_bkg}
 \end{minipage}
 \begin{minipage}[b]{0.5\hsize}
 \end{minipage}
% \begin{center}
%  (naive-$p$ = {\bf 0.00} and selective-$p$ = {\bf 0.00})
% \end{center}   
\begin{minipage}[b]{0.03\hsize}  
\end{minipage}
 \begin{minipage}[b]{0.15\hsize}
  \centering
  \includegraphics[width = \hsize]{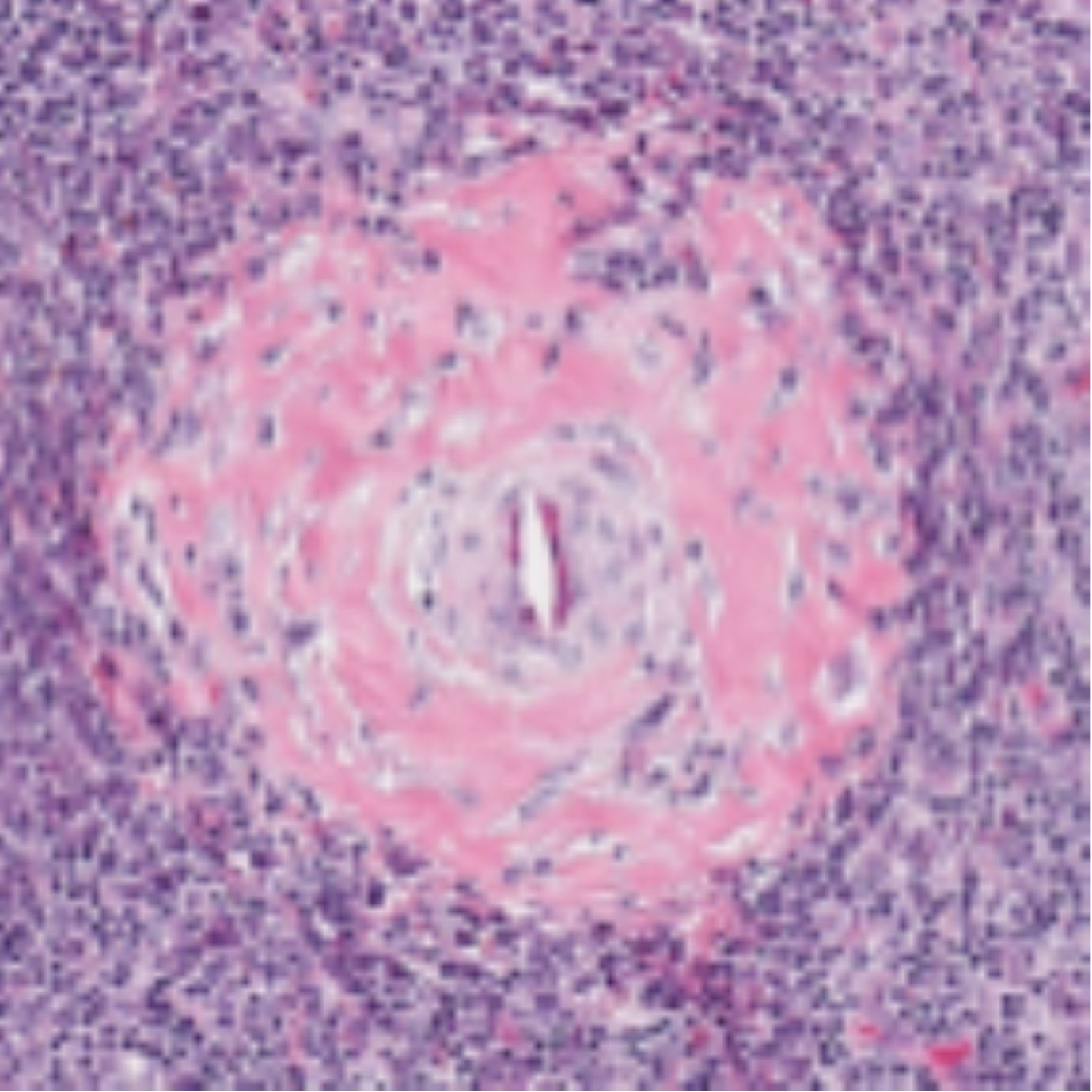}
  \subcaption{Original} \label{fig: 6}
 \end{minipage}
 \begin{minipage}[b]{0.15\hsize}
  \centering
  \includegraphics[width = \hsize]{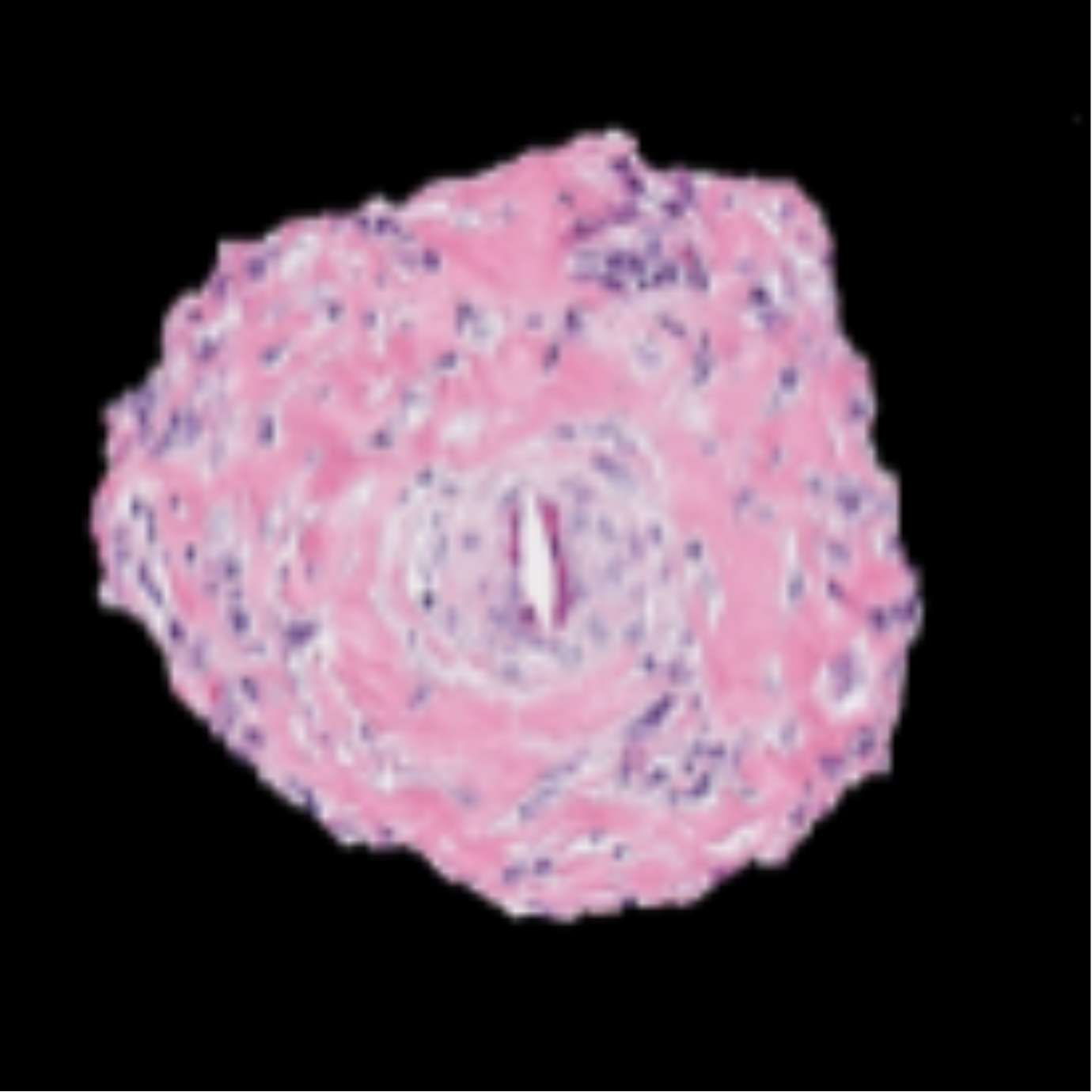}
  \subcaption{Object} \label{fig: 6_obj}
 \end{minipage}
 \begin{minipage}[b]{0.15\hsize}
  \centering
  \includegraphics[width = \hsize]{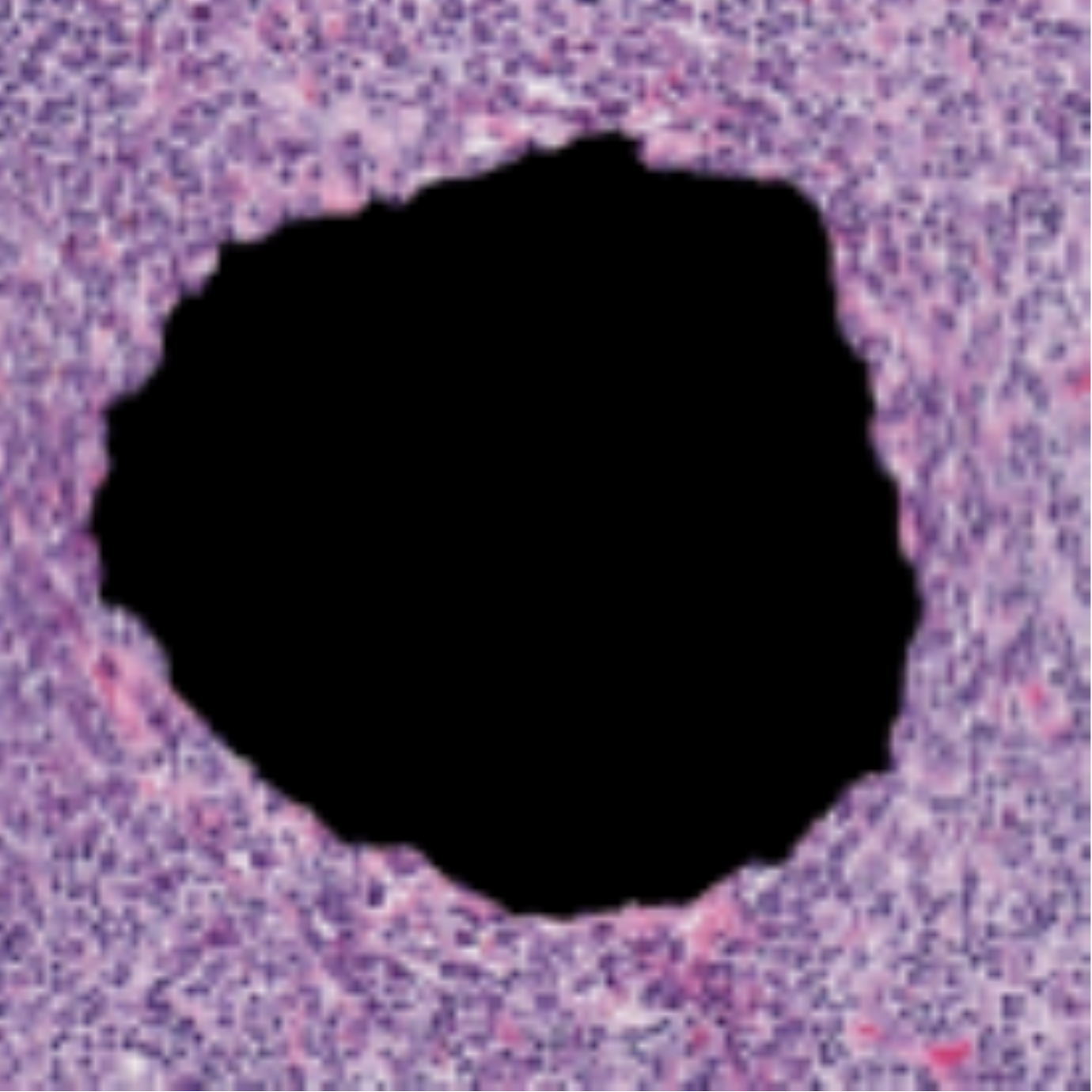}
  \subcaption{Background} \label{fig: 6_bkg}
 \end{minipage}
 \vspace{-3mm}
 \begin{center}
  ~~(naive-$p$ = {\bf 0.00}, selective-$p$ = {\bf 0.00})~~~~~~~~~~~~~~~~~~~~~~~~~~~~~~~~(naive-$p$ = {\bf 0.00}, selective-$p$ = {\bf 0.00})
 \end{center}

  \begin{minipage}[b]{0.15\hsize}
  \centering
  \includegraphics[width = \hsize]{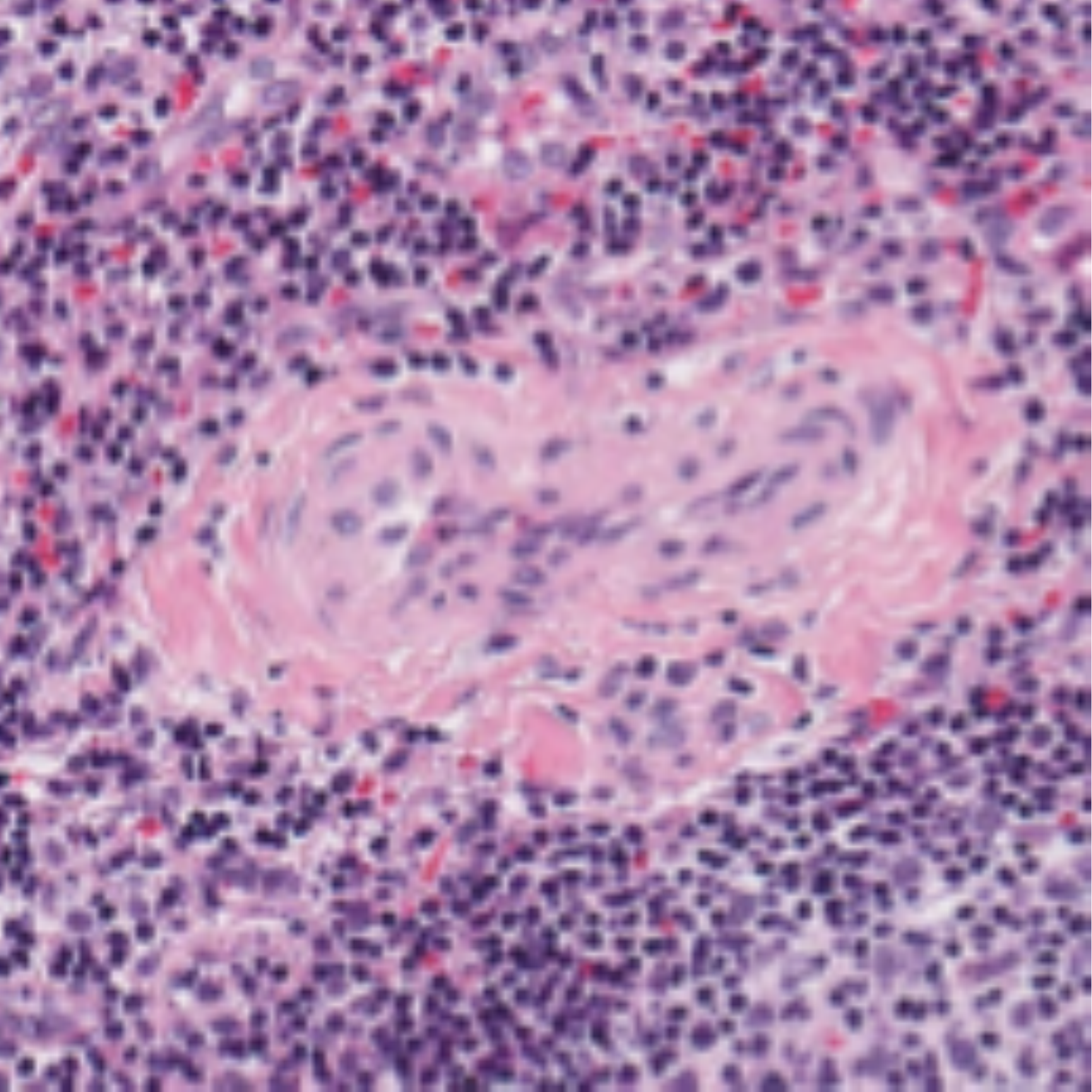}
  \subcaption{Original} \label{fig: 7}
 \end{minipage}
 \begin{minipage}[b]{0.15\hsize}
  \centering
  \includegraphics[width = \hsize]{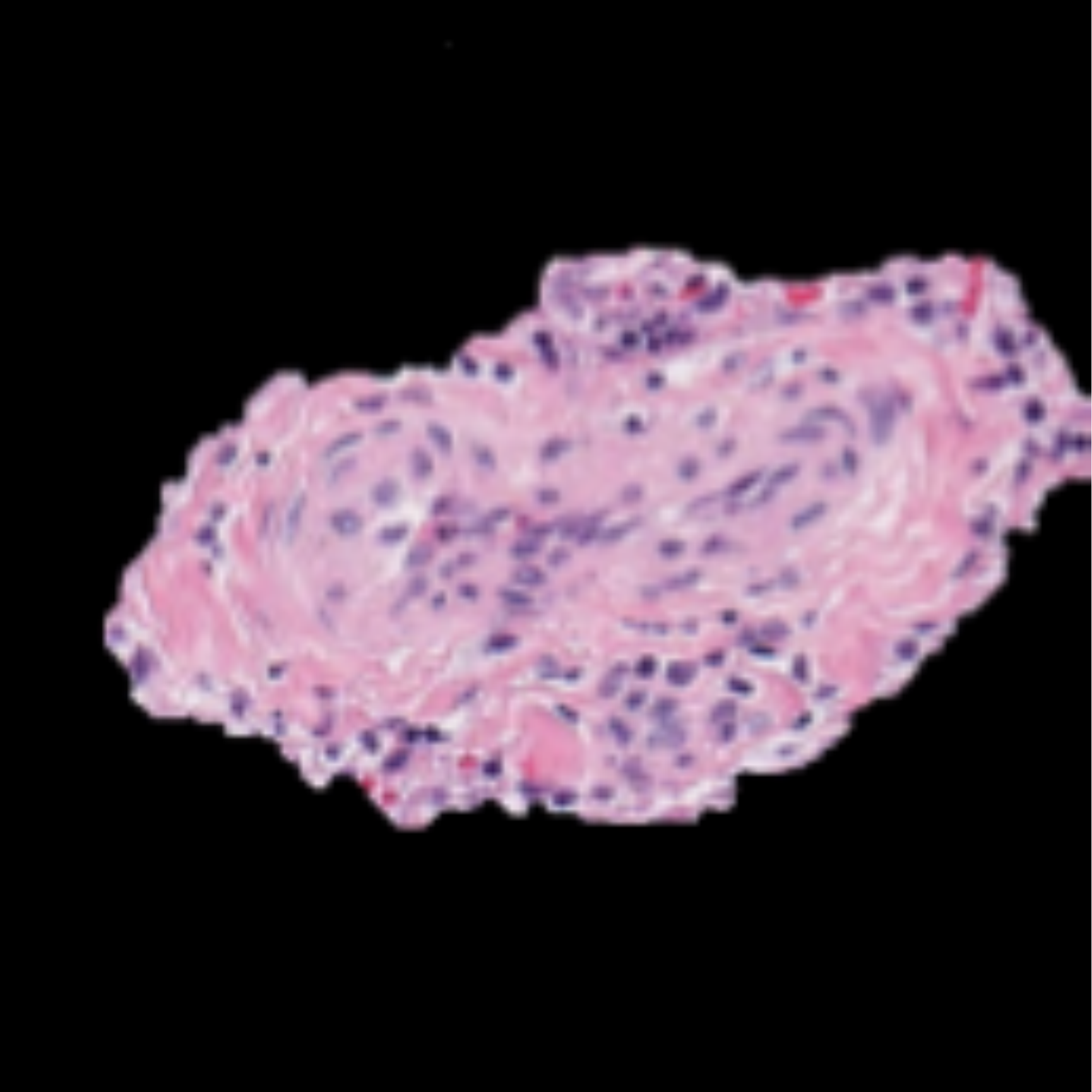}
  \subcaption{Object} \label{fig: 7_obj}
 \end{minipage}
 \begin{minipage}[b]{0.15\hsize}
  \centering
  \includegraphics[width = \hsize]{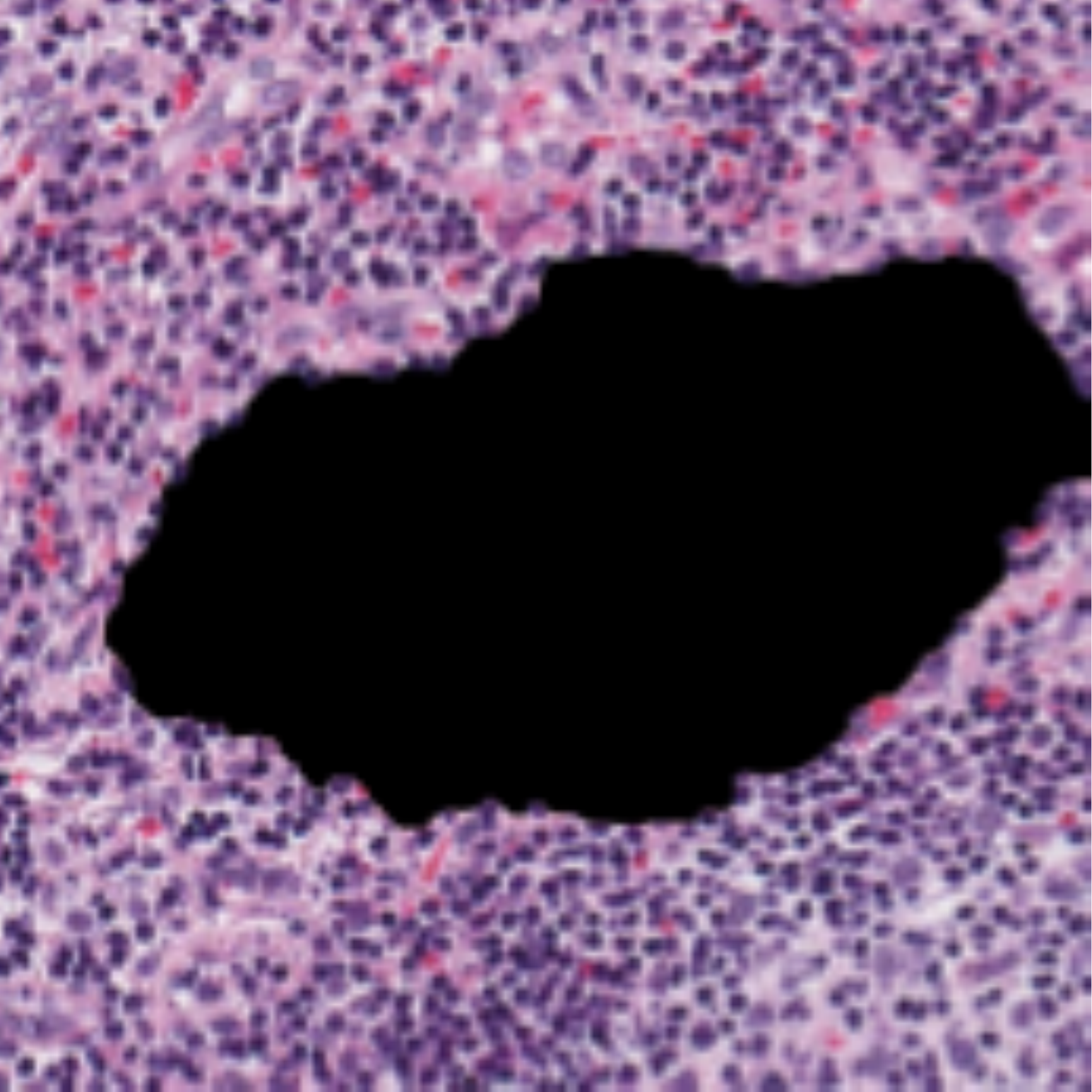}
  \subcaption{Background} \label{fig7_bkg}
 \end{minipage}
 \begin{minipage}[b]{0.5\hsize}
 \end{minipage}
% \begin{center}
%  (naive-$p$ = {\bf 0.00} and selective-$p$ = {\bf 0.00})
% \end{center}   
\begin{minipage}[b]{0.03\hsize}  
\end{minipage}
 \begin{minipage}[b]{0.15\hsize}
  \centering
  \includegraphics[width = \hsize]{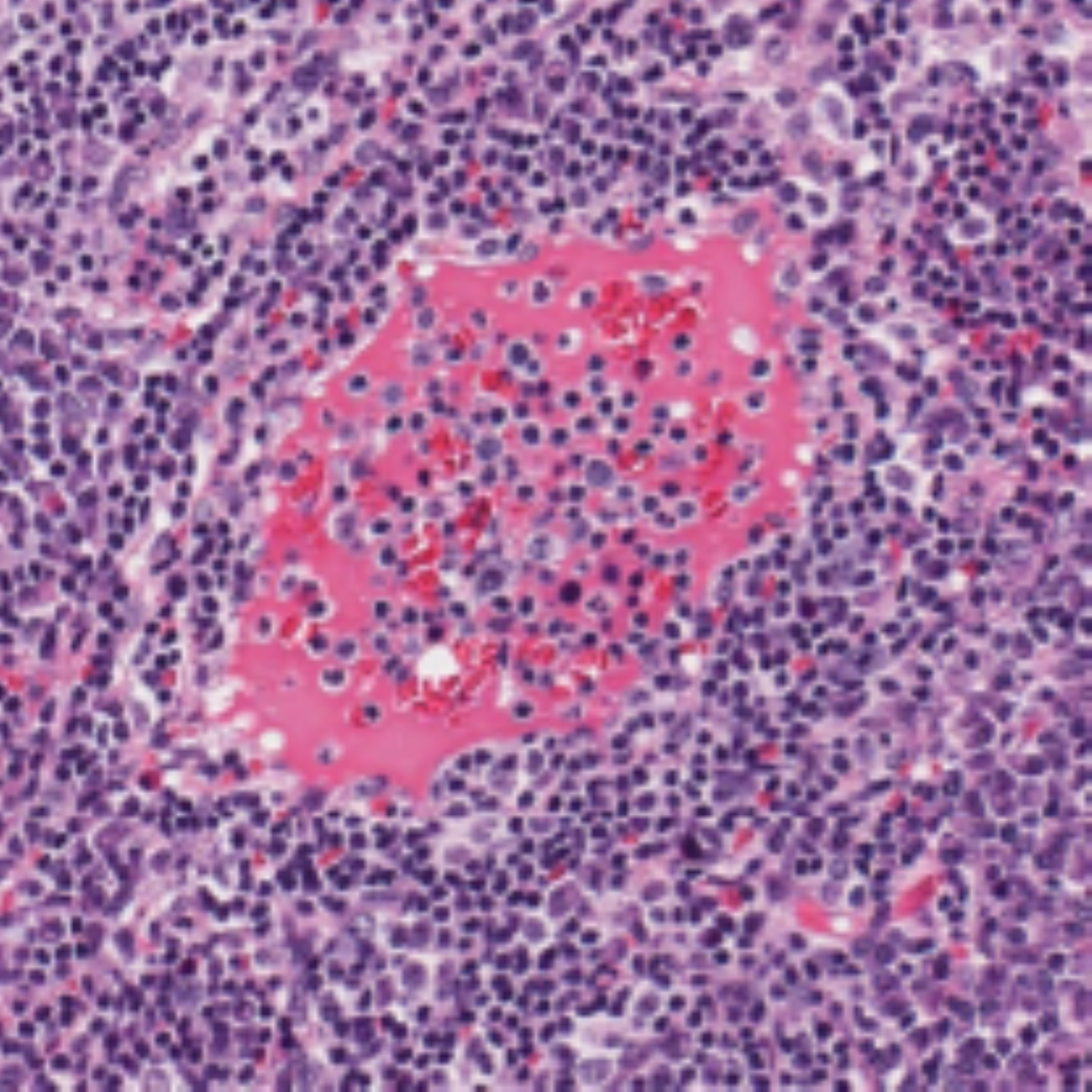}
  \subcaption{Original} \label{fig: 8}
 \end{minipage}
 \begin{minipage}[b]{0.15\hsize}
  \centering
  \includegraphics[width = \hsize]{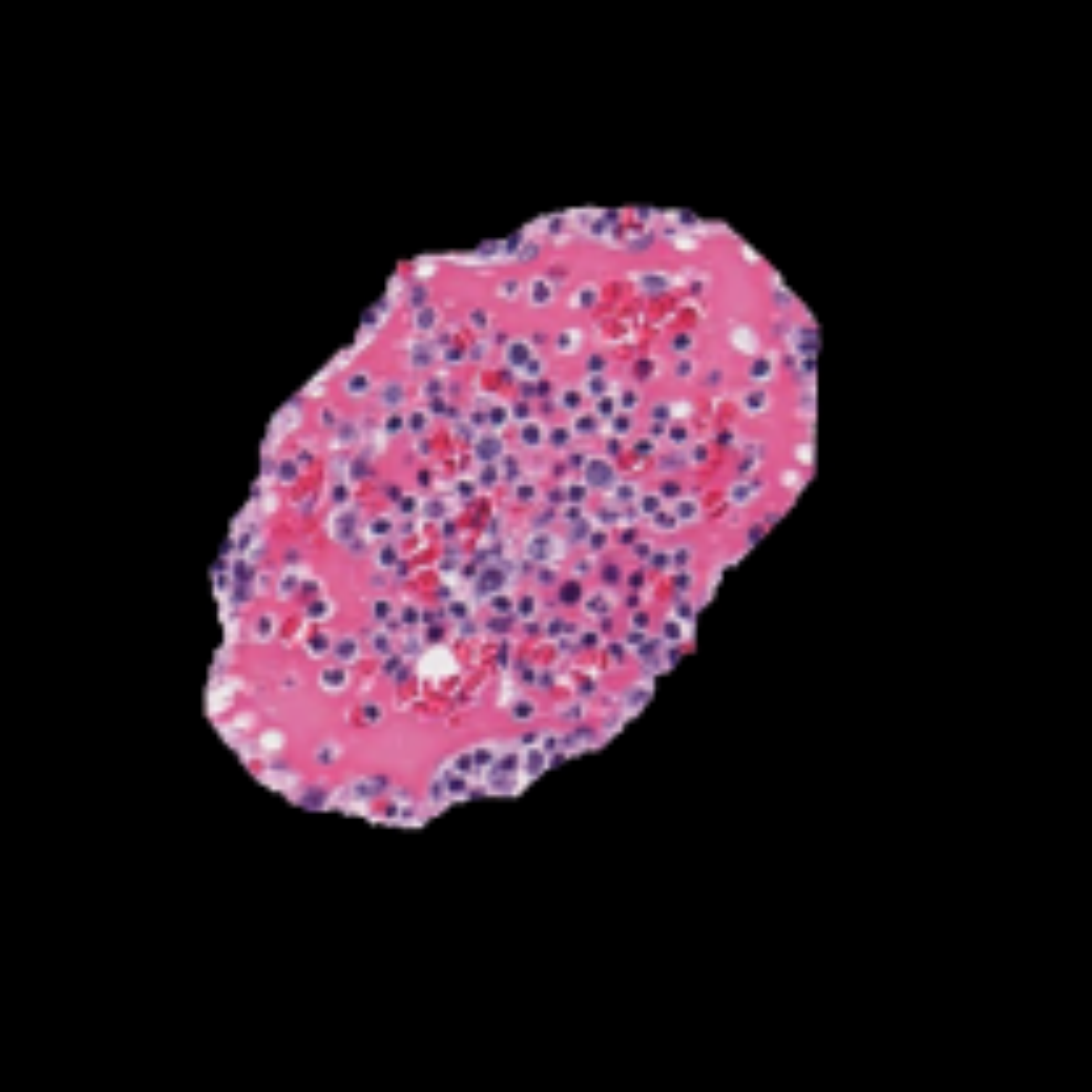}
  \subcaption{Object} \label{fig: 8_obj}
 \end{minipage}
 \begin{minipage}[b]{0.15\hsize}
  \centering
  \includegraphics[width = \hsize]{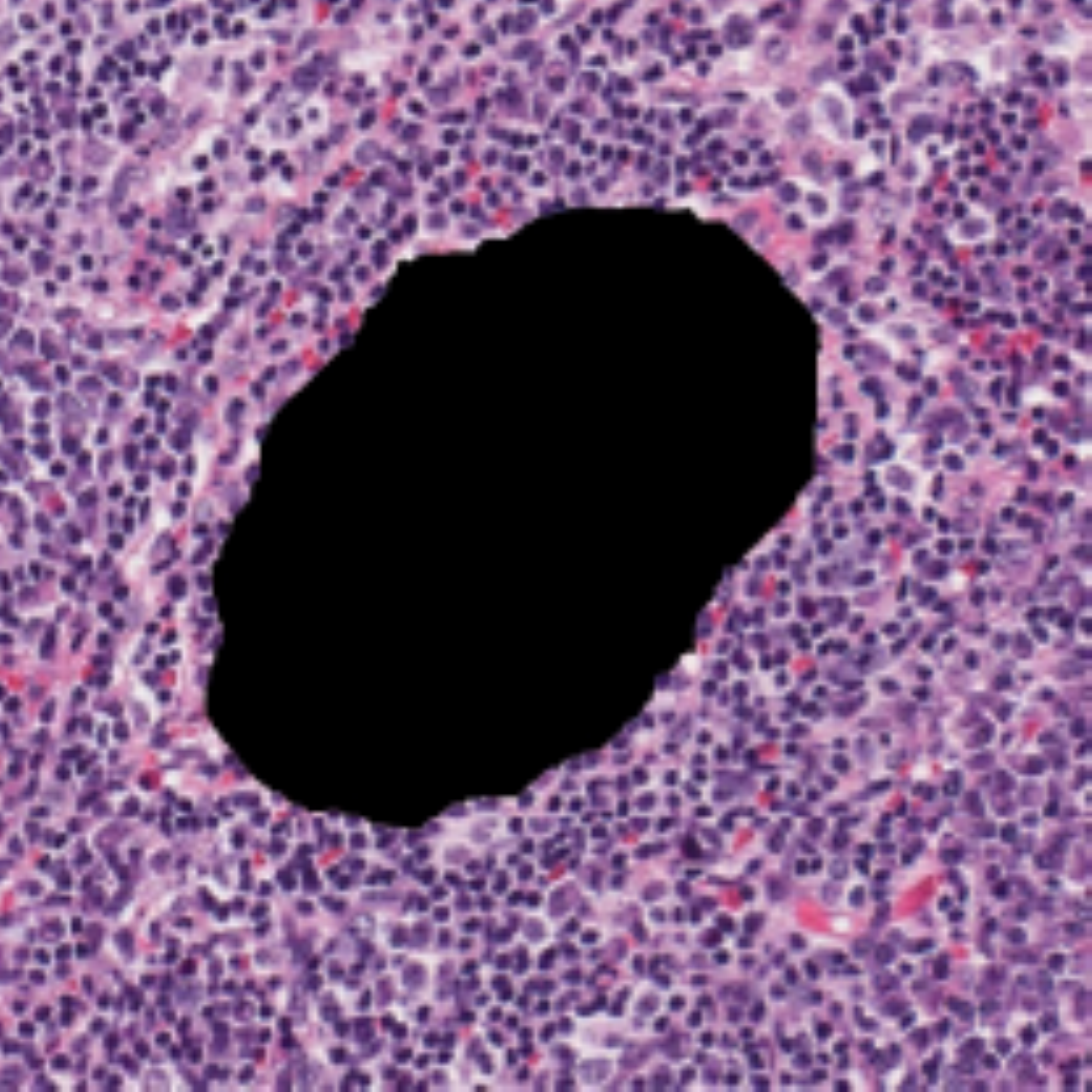}
  \subcaption{Background} \label{fig: 8_bkg}
 \end{minipage}
 \vspace{-3mm}
 \begin{center}
  ~~(naive-$p$ = {\bf 0.00}, selective-$p$ = {\bf 0.00})~~~~~~~~~~~~~~~~~~~~~~~~~~~~~~~~(naive-$p$ = {\bf 0.00}, selective-$p$ = 0.36)
 \end{center}
 \caption{
 Segmentation results for pathological images with fibrous regions
  }
 \label{fig: real}
\end{center}
\end{figure}

\subsection{Segmentation results for pathological images without fibrous regions}
\begin{figure}[H]
\begin{center}
 \begin{minipage}[b]{0.15\hsize}
  \centering
  \includegraphics[width = \hsize]{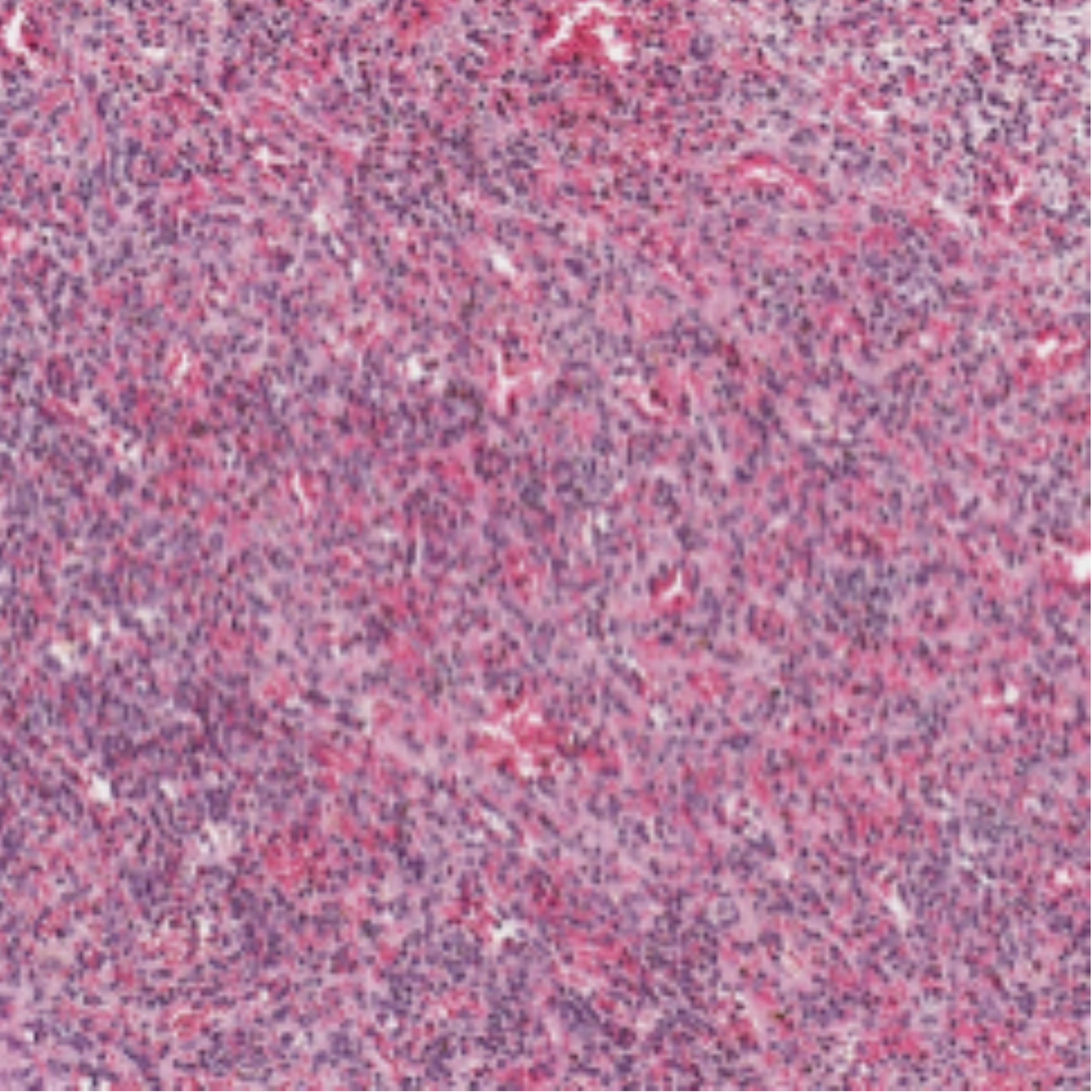}
  \subcaption{Original} \label{fig: 9}
 \end{minipage}
 \begin{minipage}[b]{0.15\hsize}
  \centering
  \includegraphics[width = \hsize]{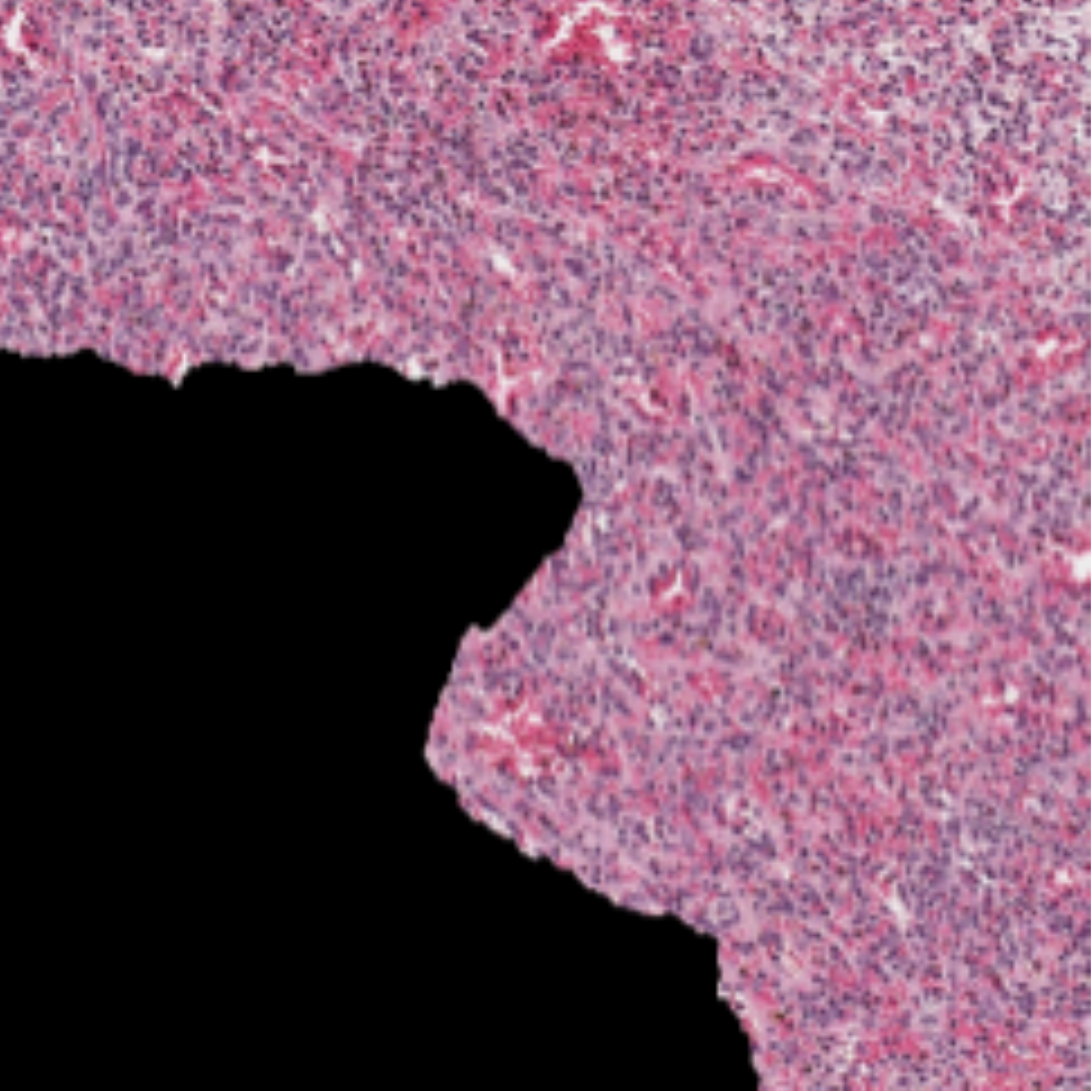}
  \subcaption{Object} \label{fig: 9_obj}
 \end{minipage}
 \begin{minipage}[b]{0.15\hsize}
  \centering
  \includegraphics[width = \hsize]{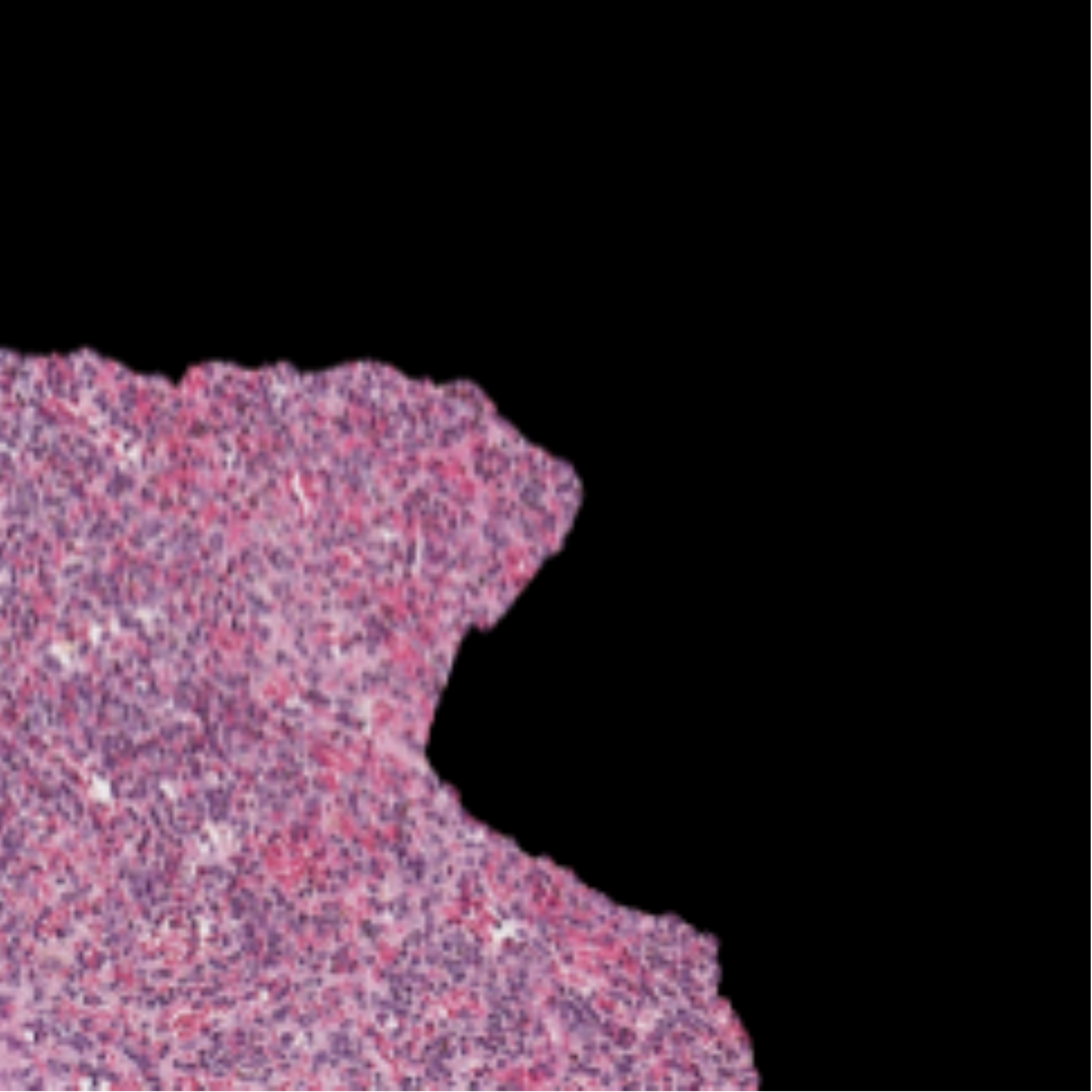}
  \subcaption{Background} \label{fig9_bkg}
 \end{minipage}
 \begin{minipage}[b]{0.5\hsize}
 \end{minipage}
% \begin{center}
%  (naive-$p$ = {\bf 0.00} and selective-$p$ = {\bf 0.00})
% \end{center}   
\begin{minipage}[b]{0.03\hsize}  
\end{minipage}
 \begin{minipage}[b]{0.15\hsize}
  \centering
  \includegraphics[width = \hsize]{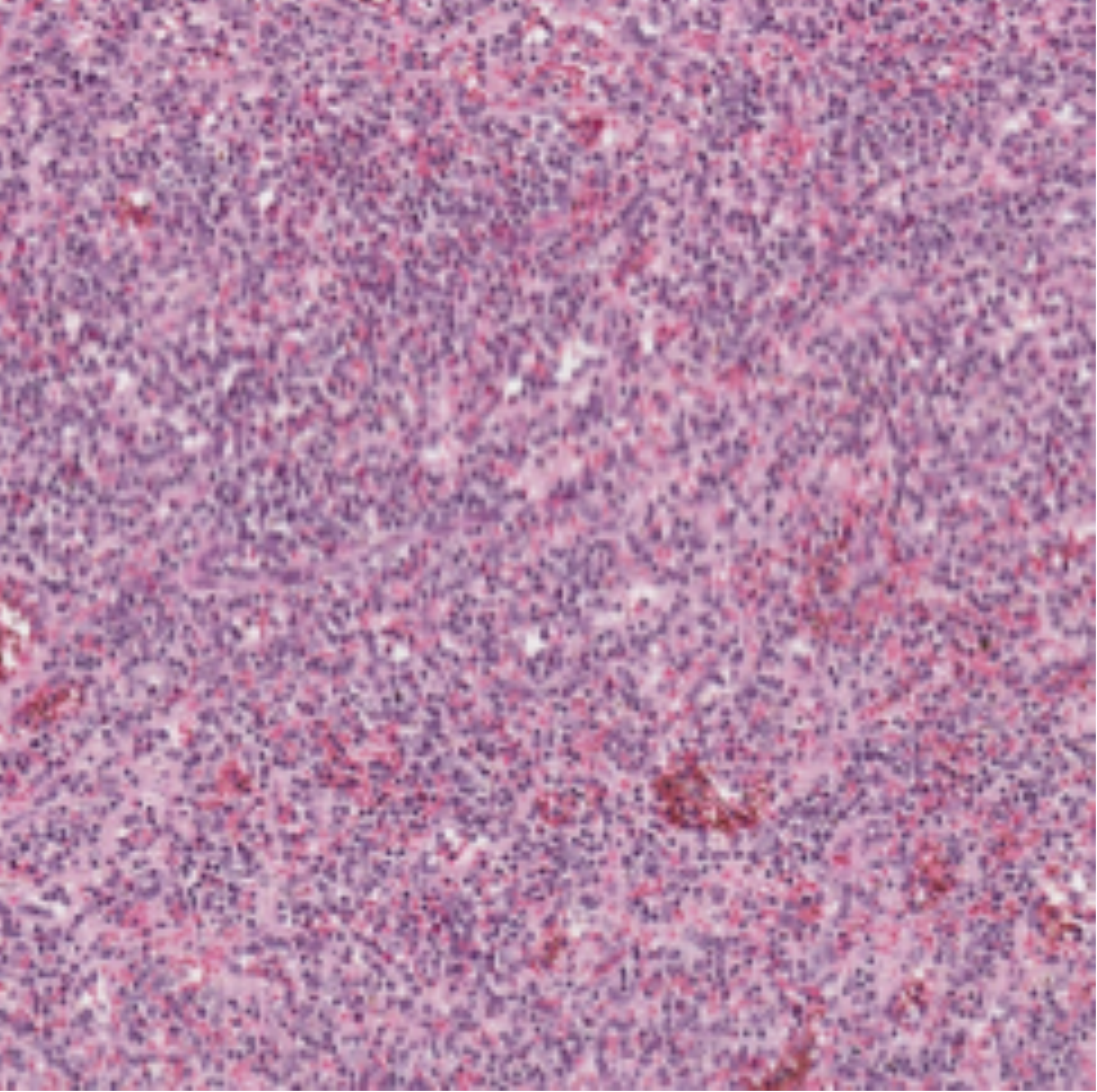}
  \subcaption{Original} \label{fig: 10}
 \end{minipage}
 \begin{minipage}[b]{0.15\hsize}
  \centering
  \includegraphics[width = \hsize]{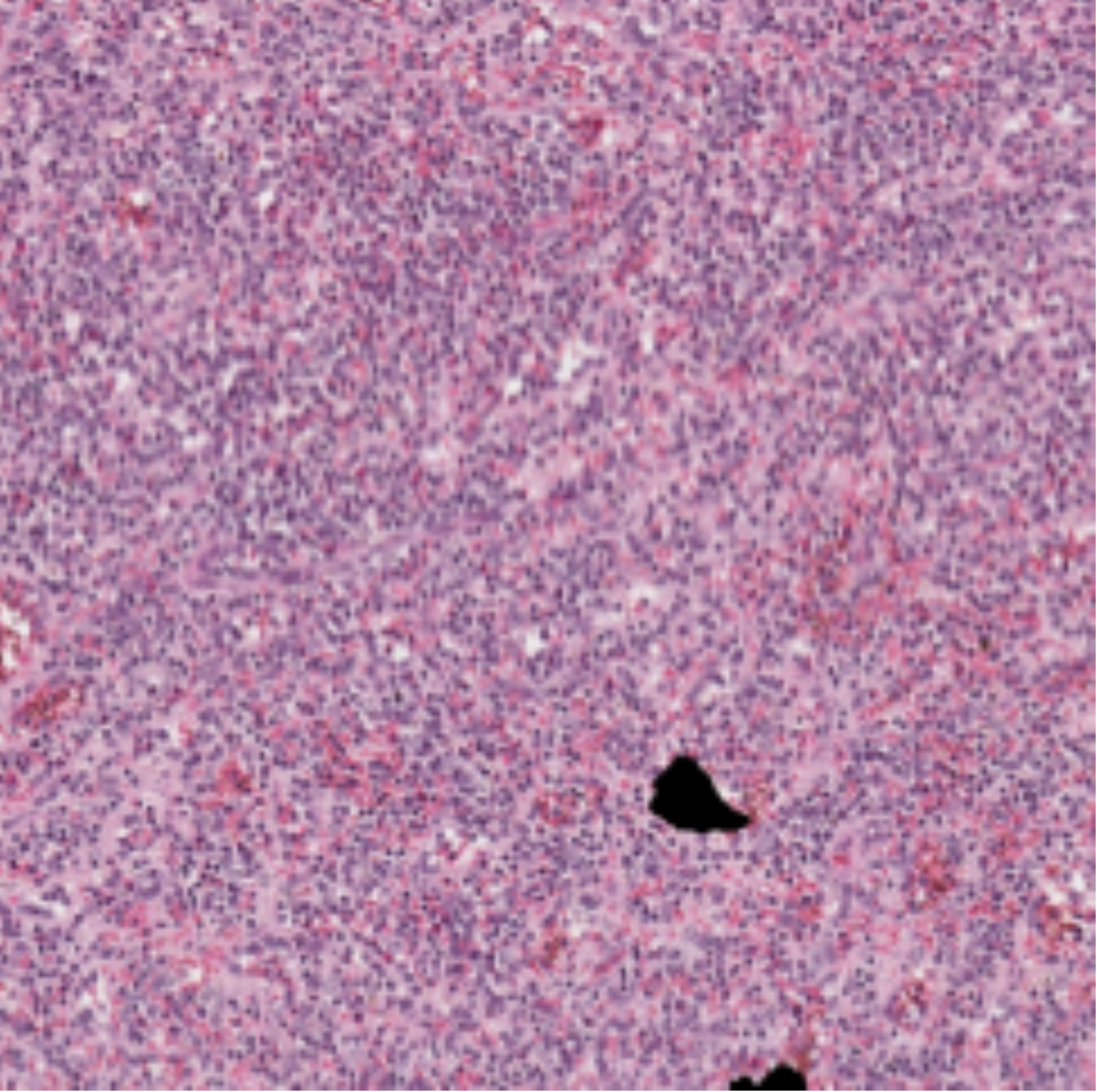}
  \subcaption{Object} \label{fig: 10_obj}
 \end{minipage}
 \begin{minipage}[b]{0.15\hsize}
  \centering
  \includegraphics[width = \hsize]{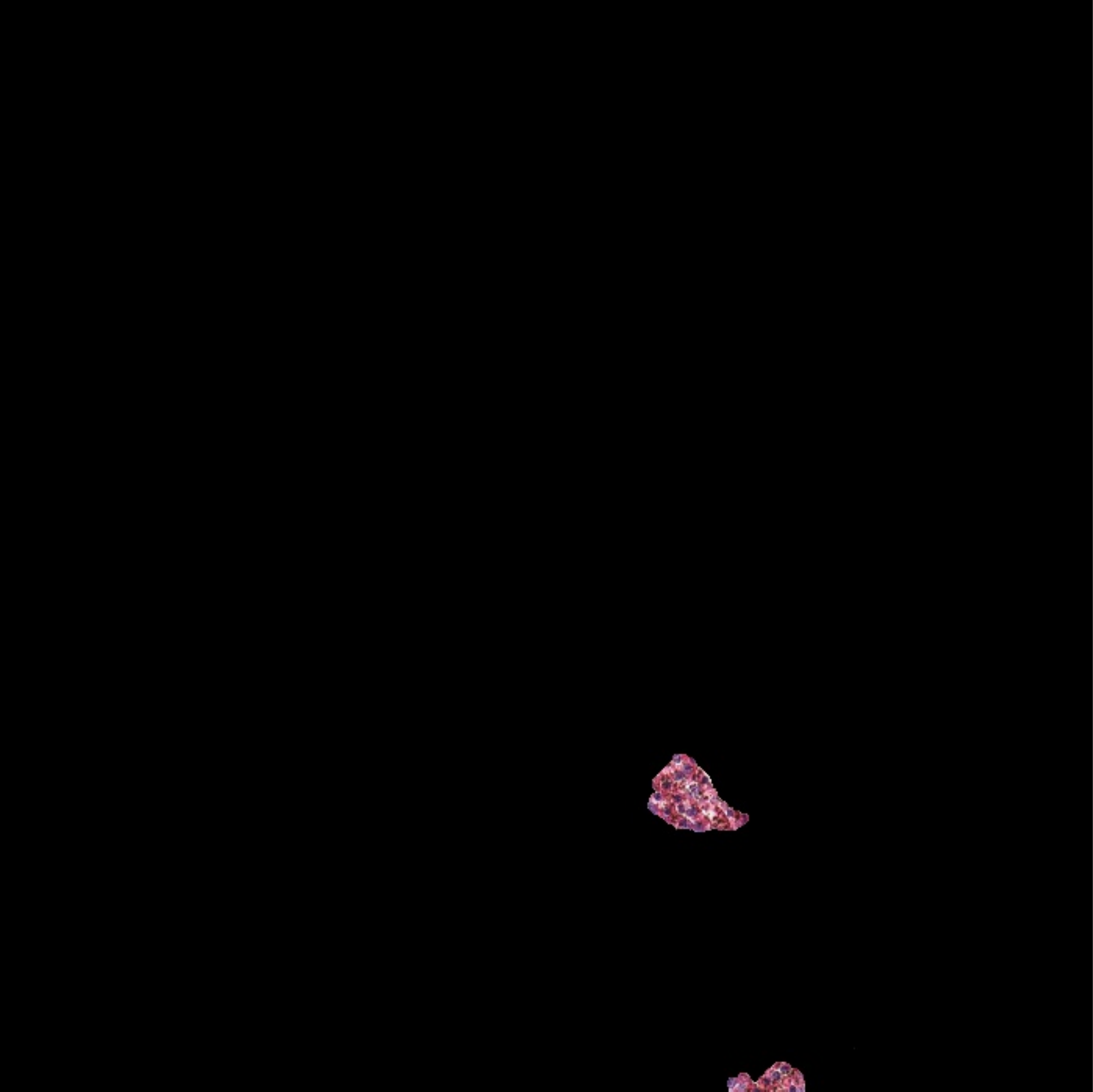}
  \subcaption{Background} \label{fig: 10_bkg}
 \end{minipage}
 \vspace{-3mm}
 \begin{center}
  ~~(naive-$p$ = {\bf 0.00}, selective-$p$ = 0.24)~~~~~~~~~~~~~~~~~~~~~~~~~~~~~~~~(naive-$p$ = {\bf 0.00}, selective-$p$ = 0.08)
 \end{center}
% \begin{flushright}
%  (naive-$p$ = {\bf 0.00}, selective-$p$ = {\bf 0.00})
% \end{flushright}
 \begin{minipage}[b]{0.15\hsize}
  \centering
  \includegraphics[width = \hsize]{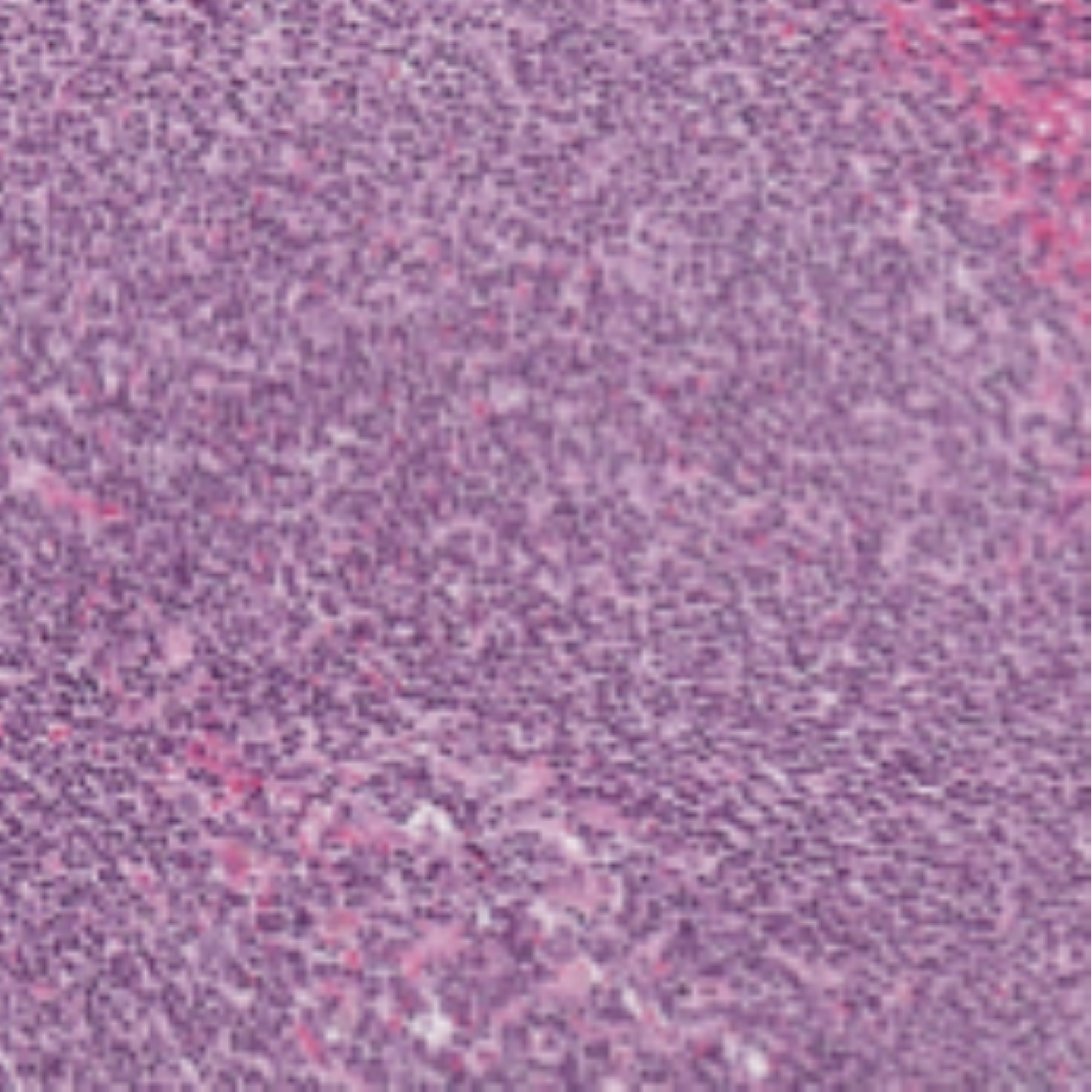}
  \subcaption{Original} \label{fig: 11}
 \end{minipage}
 \begin{minipage}[b]{0.15\hsize}
  \centering
  \includegraphics[width = \hsize]{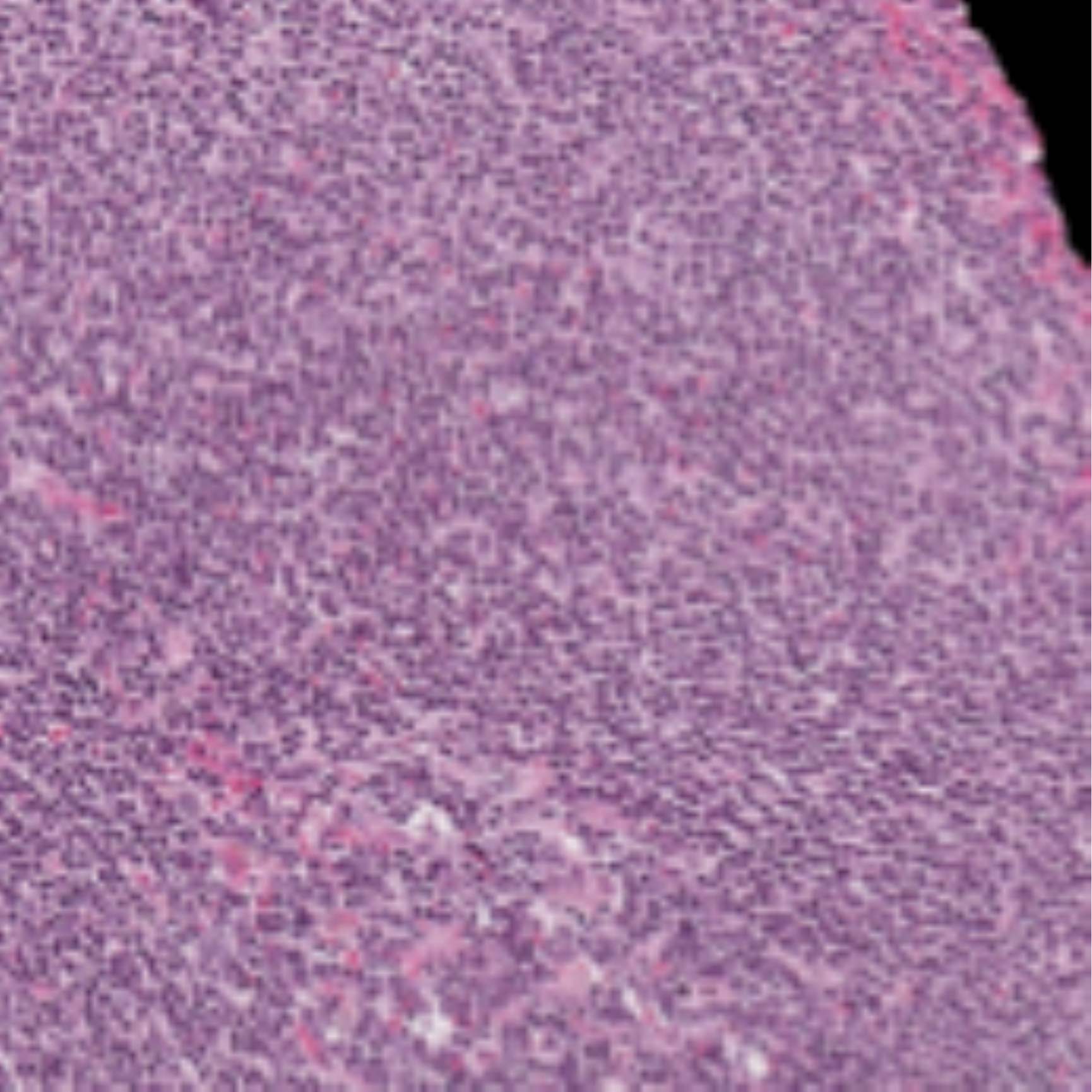}
  \subcaption{Object} \label{fig: 11_obj}
 \end{minipage}
 \begin{minipage}[b]{0.15\hsize}
  \centering
  \includegraphics[width = \hsize]{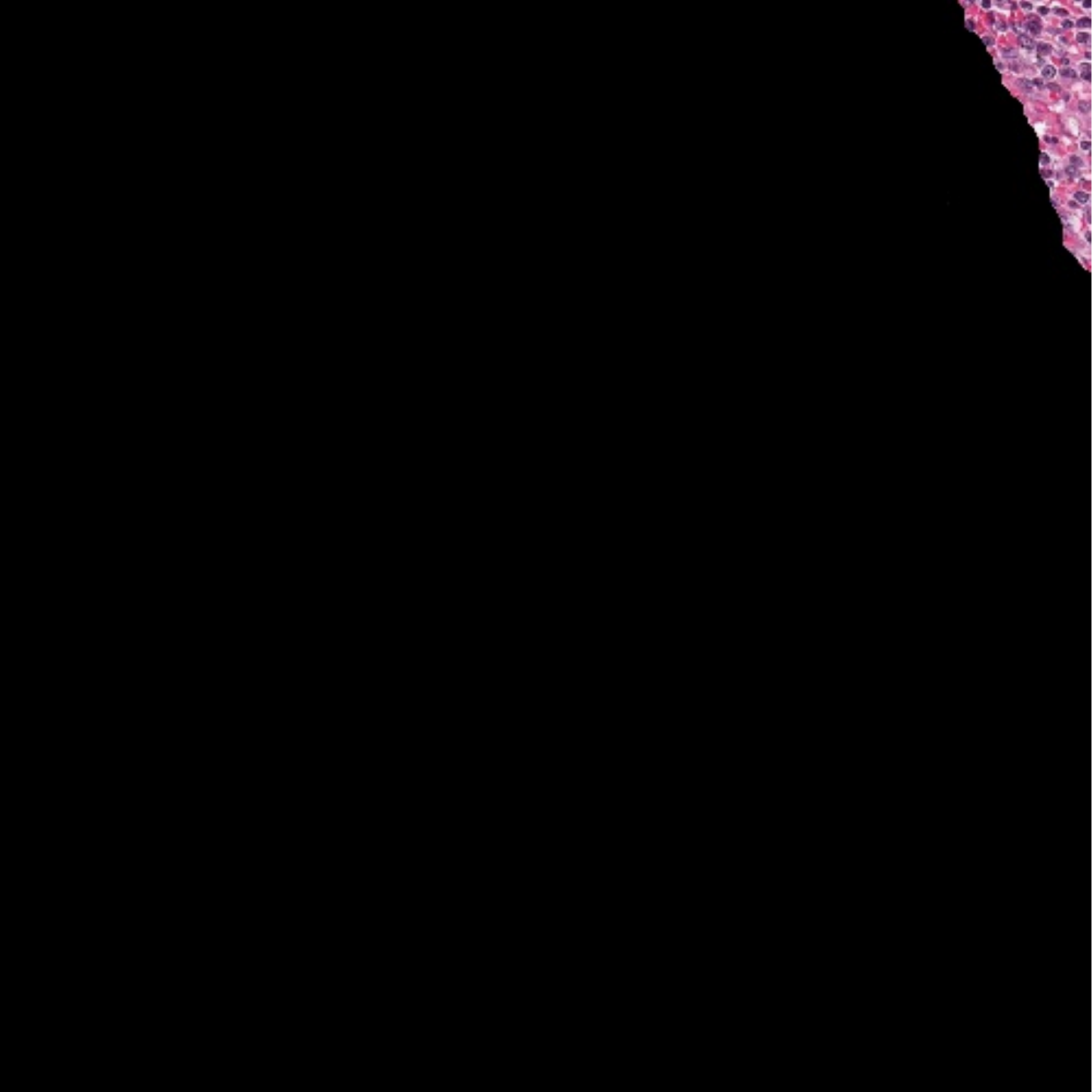}
  \subcaption{Background} \label{fig11_bkg}
 \end{minipage}
 \begin{minipage}[b]{0.5\hsize}
 \end{minipage}
% \begin{center}
%  (naive-$p$ = {\bf 0.00} and selective-$p$ = {\bf 0.00})
% \end{center}   
\begin{minipage}[b]{0.03\hsize}  
\end{minipage}
 \begin{minipage}[b]{0.15\hsize}
  \centering
  \includegraphics[width = \hsize]{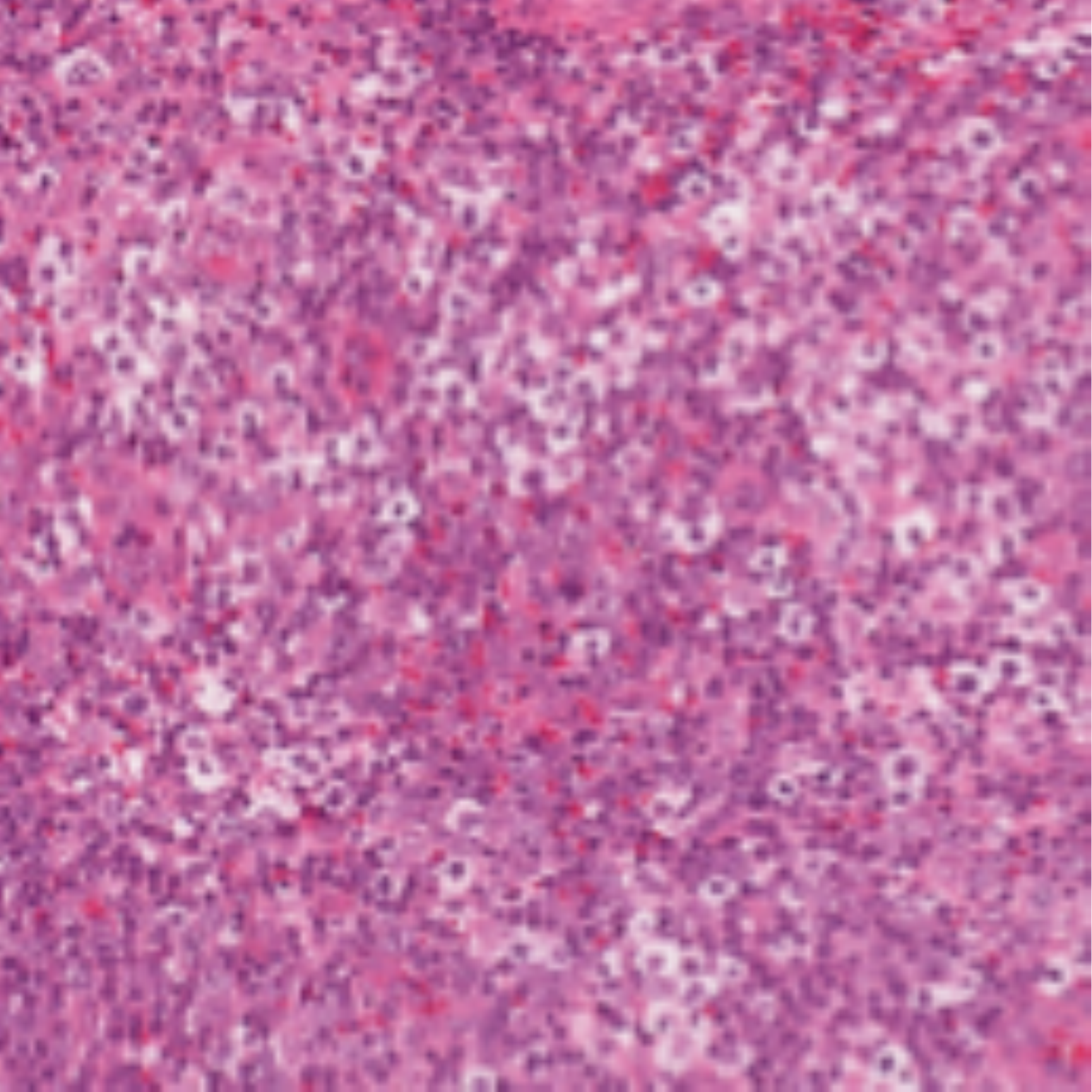}
  \subcaption{Original} \label{fig: 12}
 \end{minipage}
 \begin{minipage}[b]{0.15\hsize}
  \centering
  \includegraphics[width = \hsize]{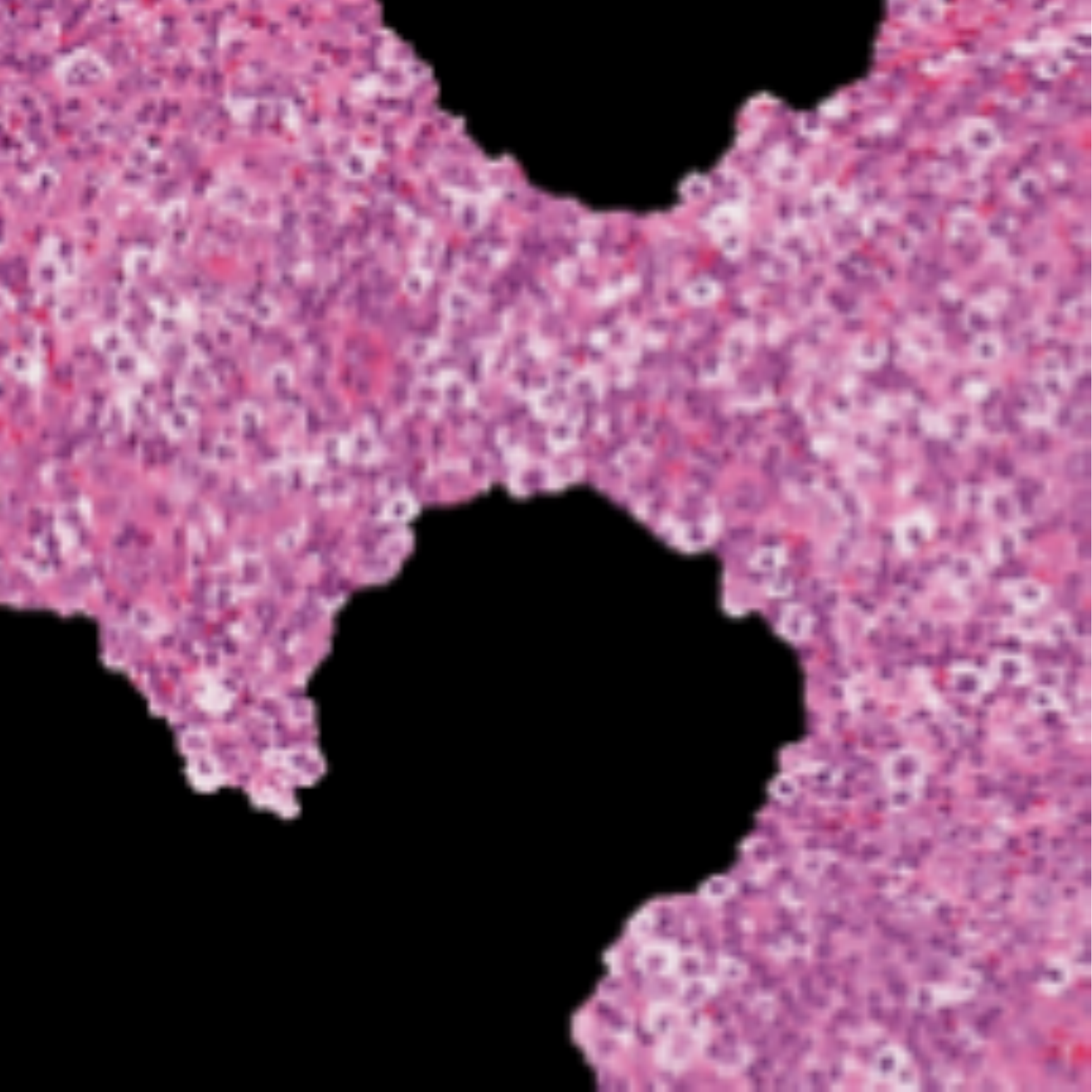}
  \subcaption{Object} \label{fig: 12_obj}
 \end{minipage}
 \begin{minipage}[b]{0.15\hsize}
  \centering
  \includegraphics[width = \hsize]{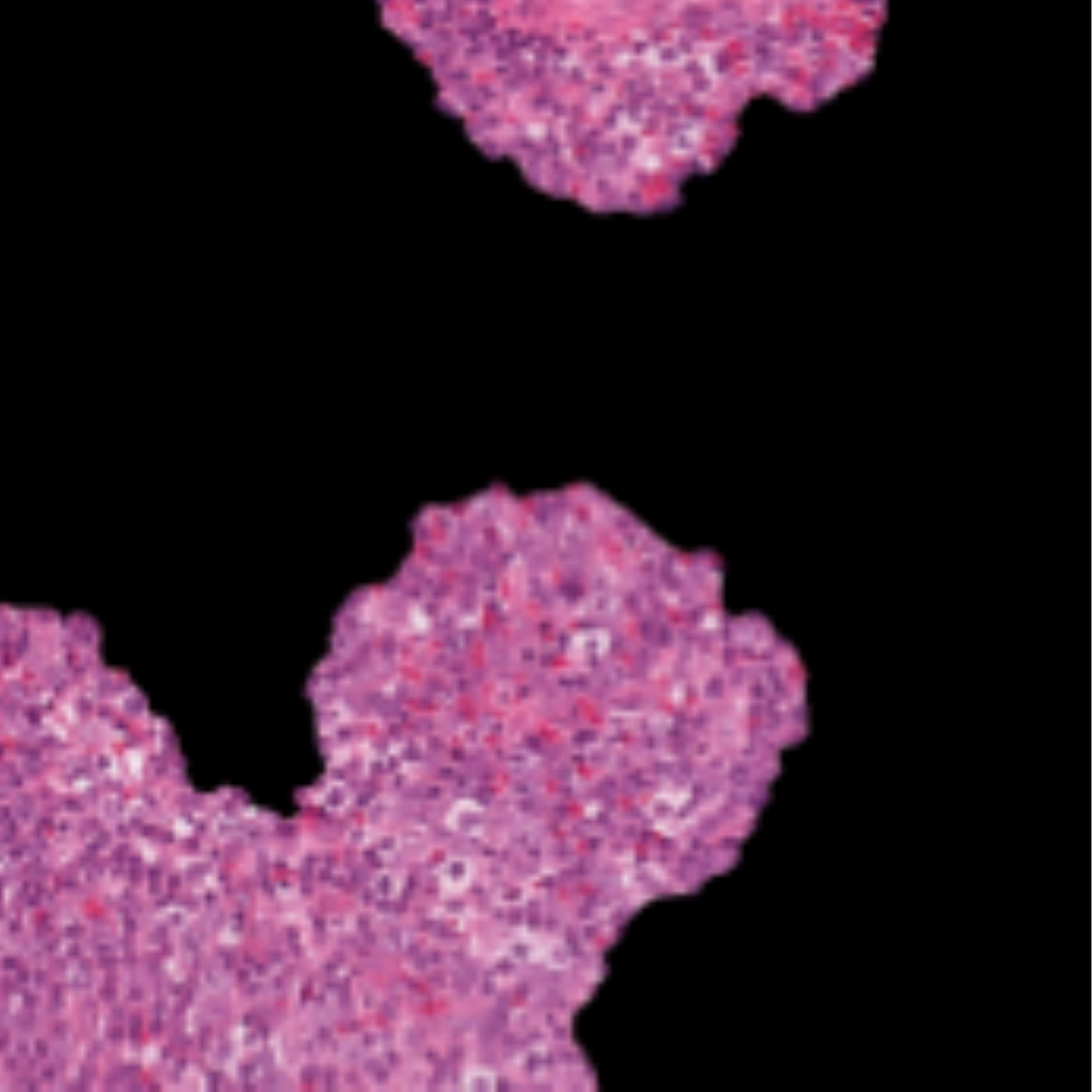}
  \subcaption{Background} \label{fig: 12_bkg}
 \end{minipage}
 \vspace{-3mm}
 \begin{center}
  ~~(naive-$p$ = {\bf 0.00}, selective-$p$ = 0.35)~~~~~~~~~~~~~~~~~~~~~~~~~~~~~~~~(naive-$p$ = {\bf 0.00}, selective-$p$ = 0.86)
 \end{center}
 
  \begin{minipage}[b]{0.15\hsize}
  \centering
  \includegraphics[width = \hsize]{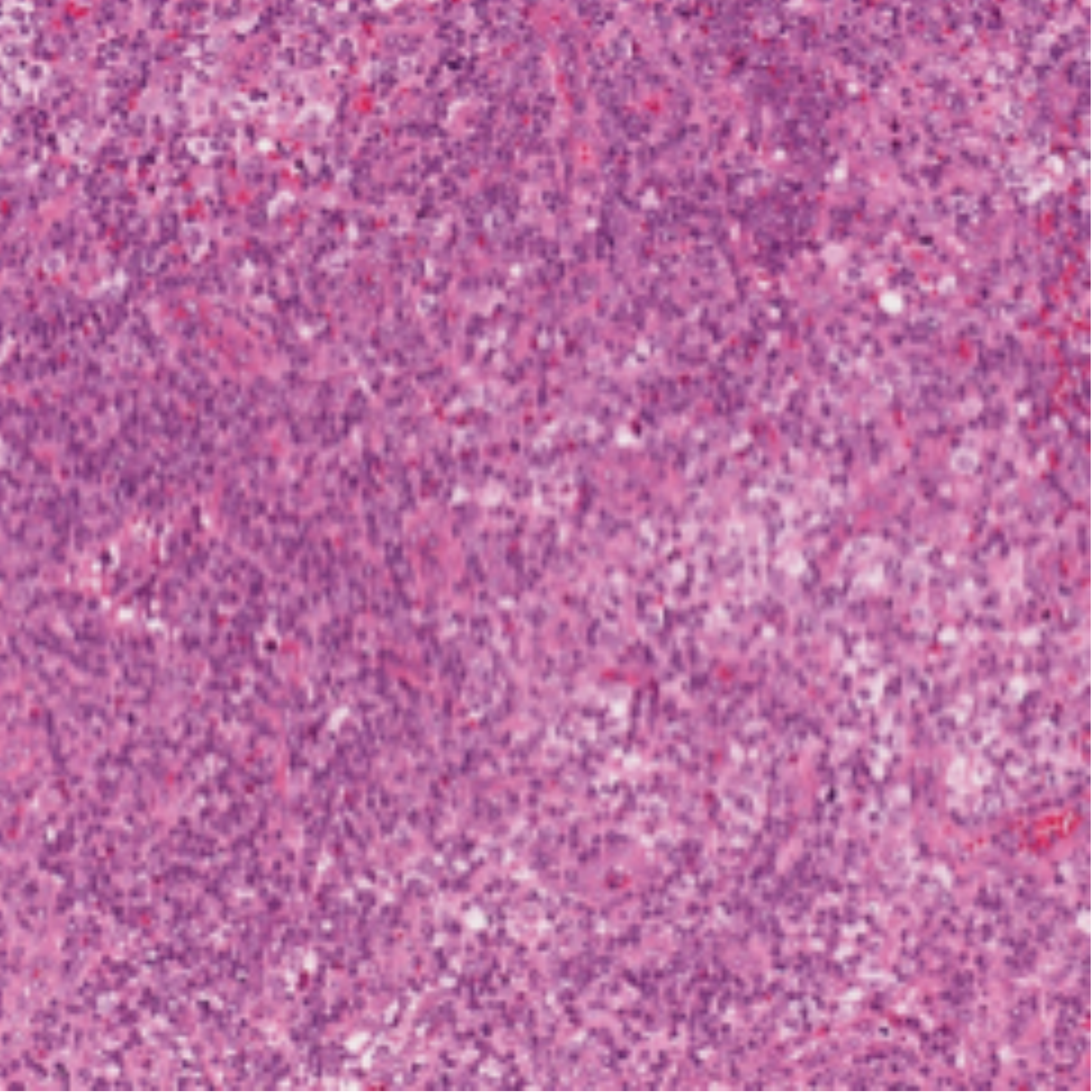}
  \subcaption{Original} \label{fig: 13}
 \end{minipage}
 \begin{minipage}[b]{0.15\hsize}
  \centering
  \includegraphics[width = \hsize]{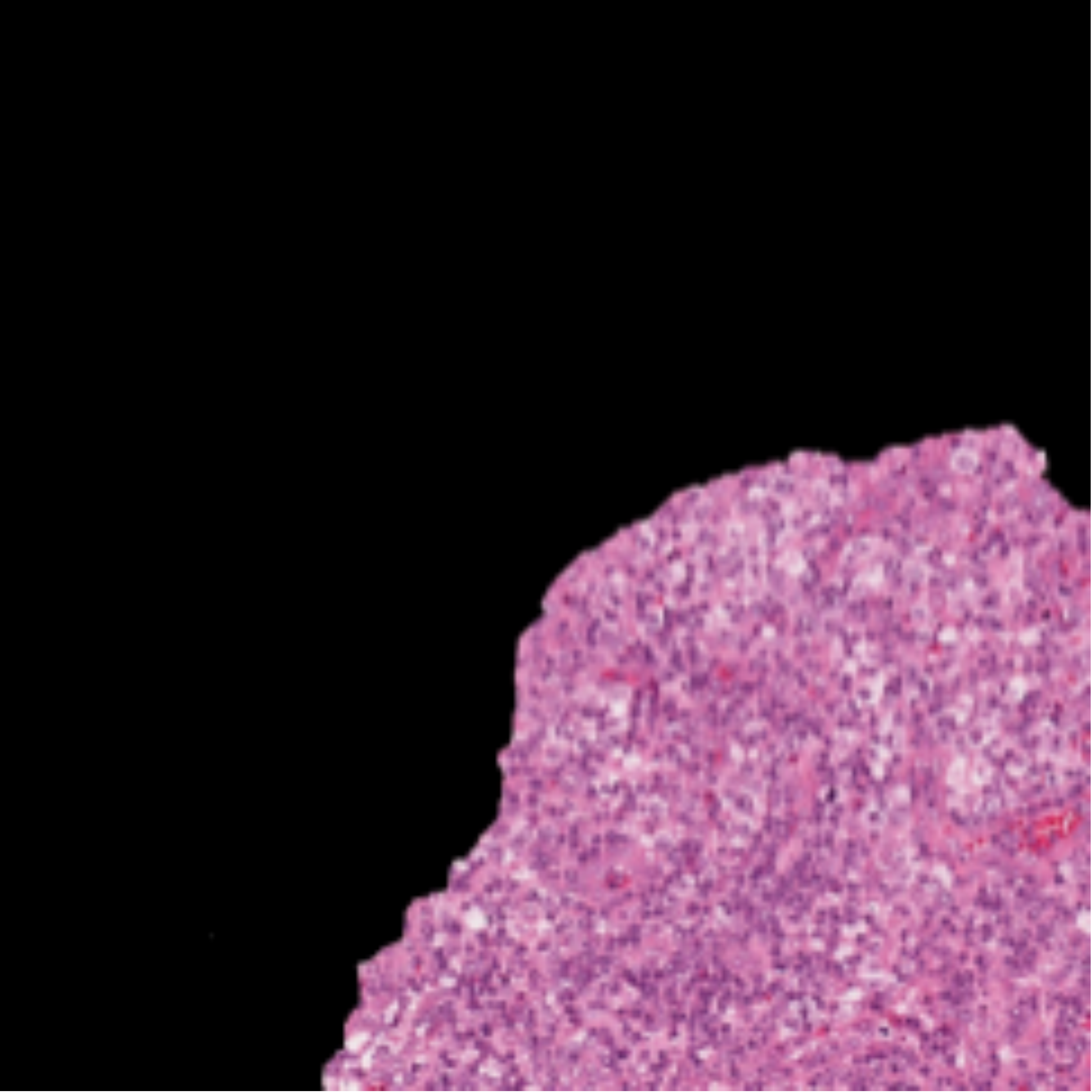}
  \subcaption{Object} \label{fig: 13_obj}
 \end{minipage}
 \begin{minipage}[b]{0.15\hsize}
  \centering
  \includegraphics[width = \hsize]{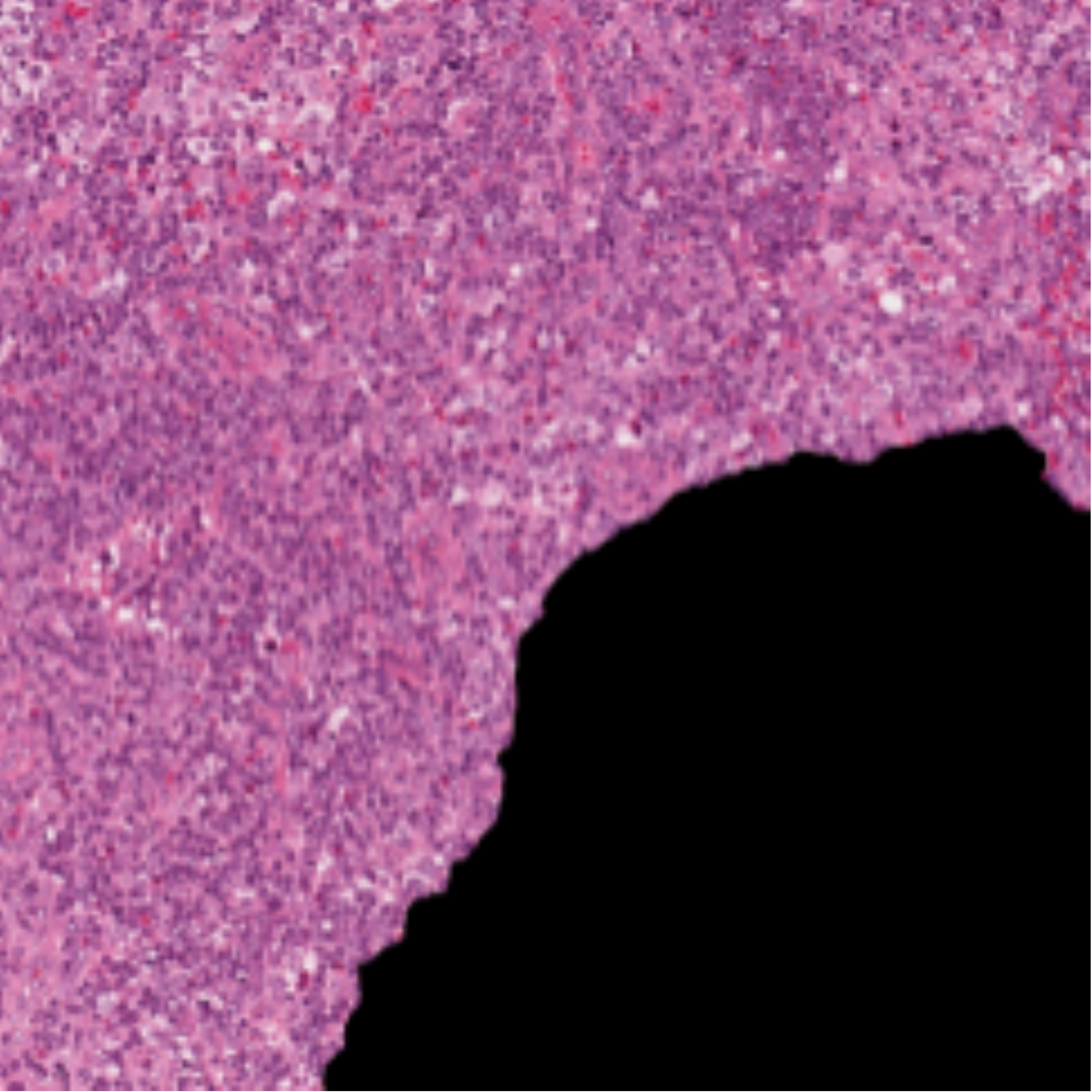}
  \subcaption{Background} \label{fig13_bkg}
 \end{minipage}
 \begin{minipage}[b]{0.5\hsize}
 \end{minipage}
% \begin{center}
%  (naive-$p$ = {\bf 0.00} and selective-$p$ = {\bf 0.00})
% \end{center}   
\begin{minipage}[b]{0.03\hsize}  
\end{minipage}
 \begin{minipage}[b]{0.15\hsize}
  \centering
  \includegraphics[width = \hsize]{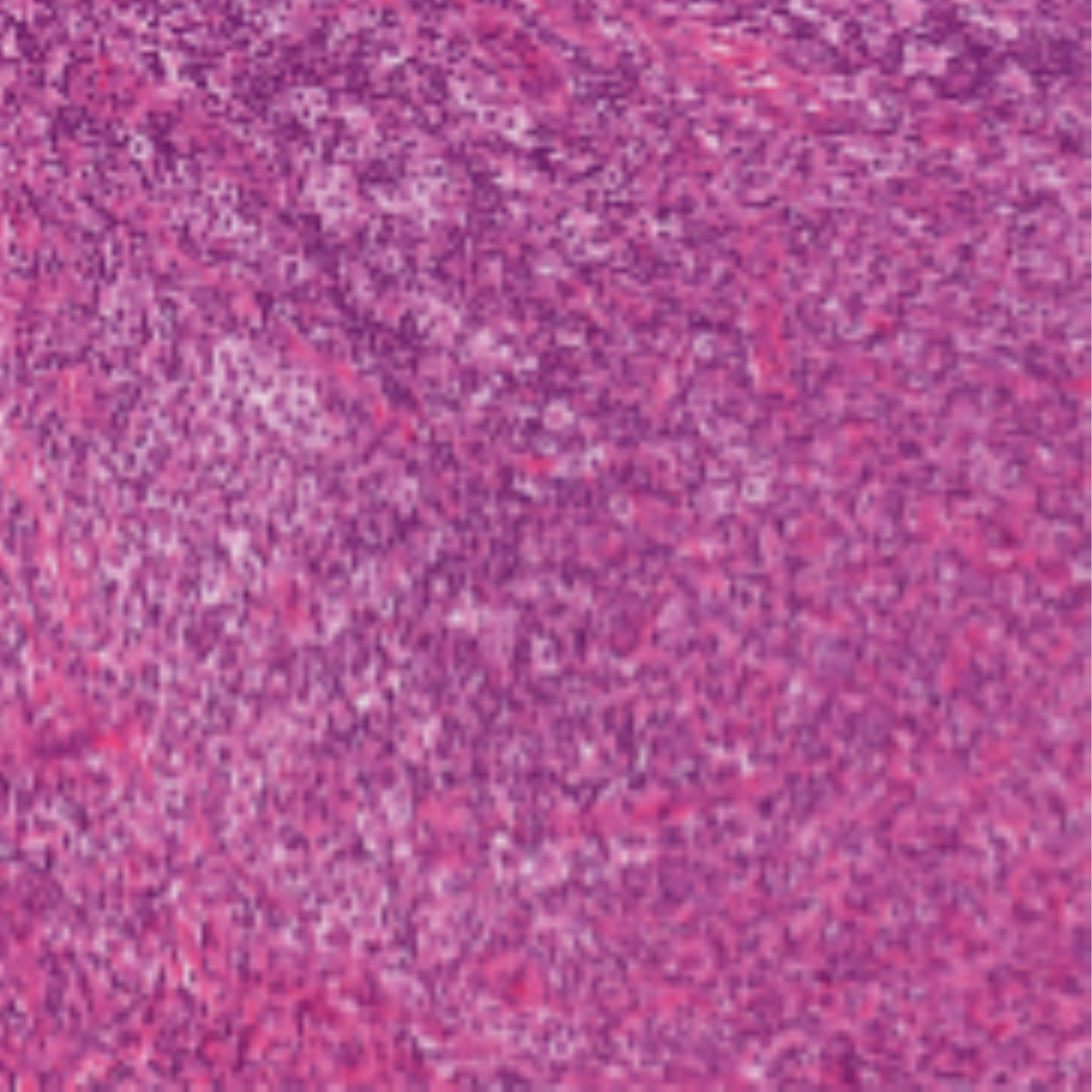}
  \subcaption{Original} \label{fig: 14}
 \end{minipage}
 \begin{minipage}[b]{0.15\hsize}
  \centering
  \includegraphics[width = \hsize]{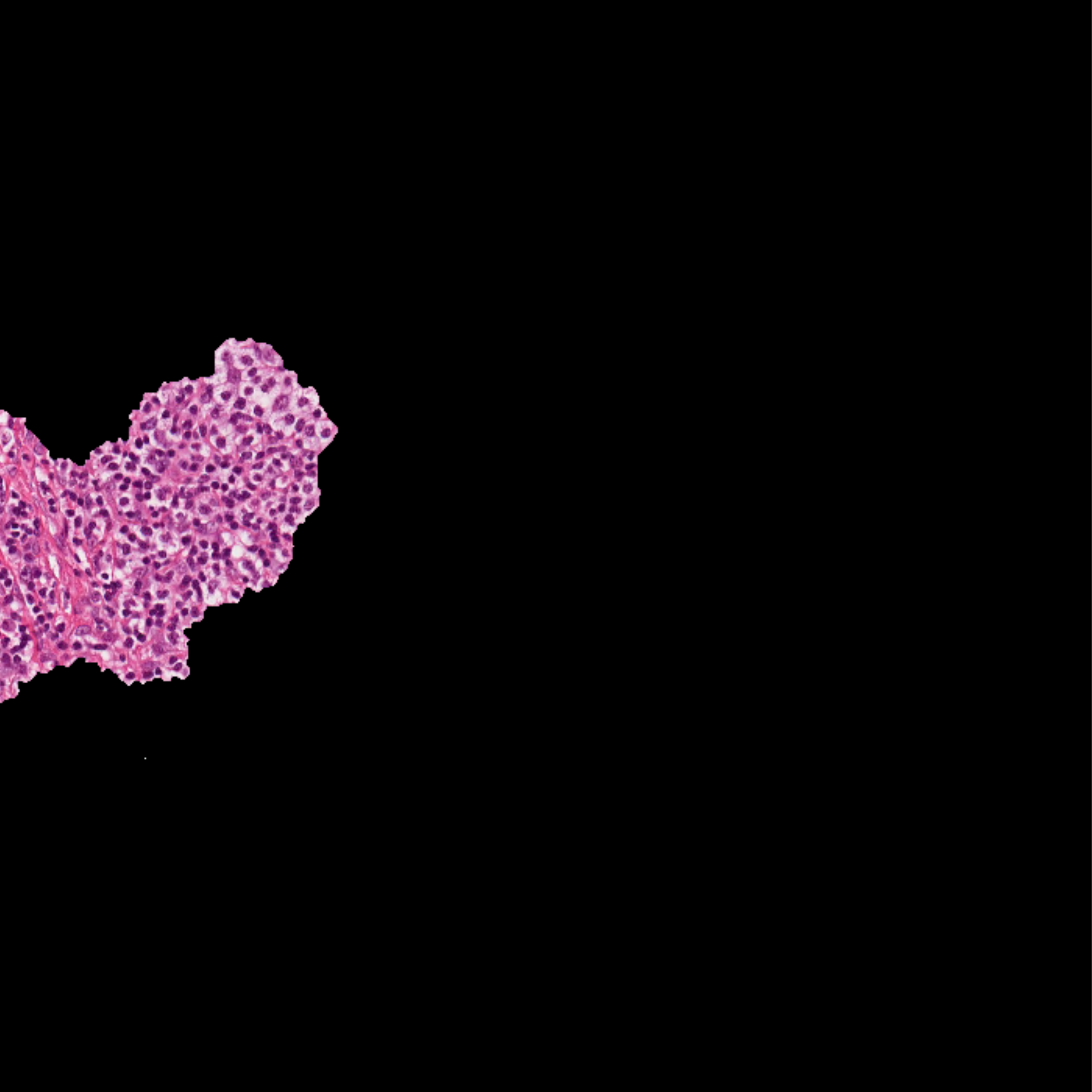}
  \subcaption{Object} \label{fig: 14_obj}
 \end{minipage}
 \begin{minipage}[b]{0.15\hsize}
  \centering
  \includegraphics[width = \hsize]{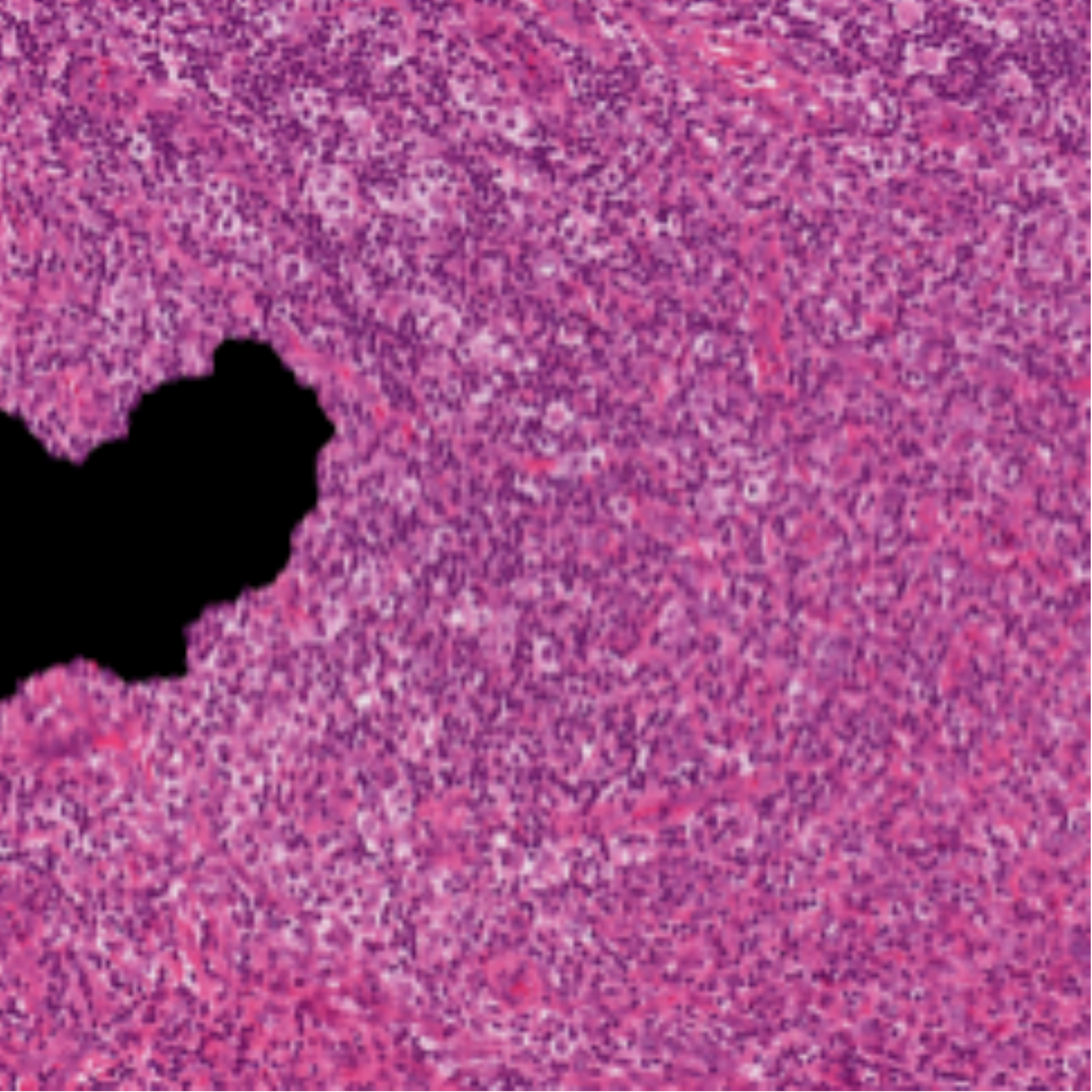}
  \subcaption{Background} \label{fig: 14_bkg}
 \end{minipage}
 \vspace{-3mm}
 \begin{center}
  ~~(naive-$p$ = {\bf 0.00}, selective-$p$ = 0.95)~~~~~~~~~~~~~~~~~~~~~~~~~~~~~~~~(naive-$p$ = {\bf 0.00}, selective-$p$ = 0.67)
 \end{center}

  \begin{minipage}[b]{0.15\hsize}
  \centering
  \includegraphics[width = \hsize]{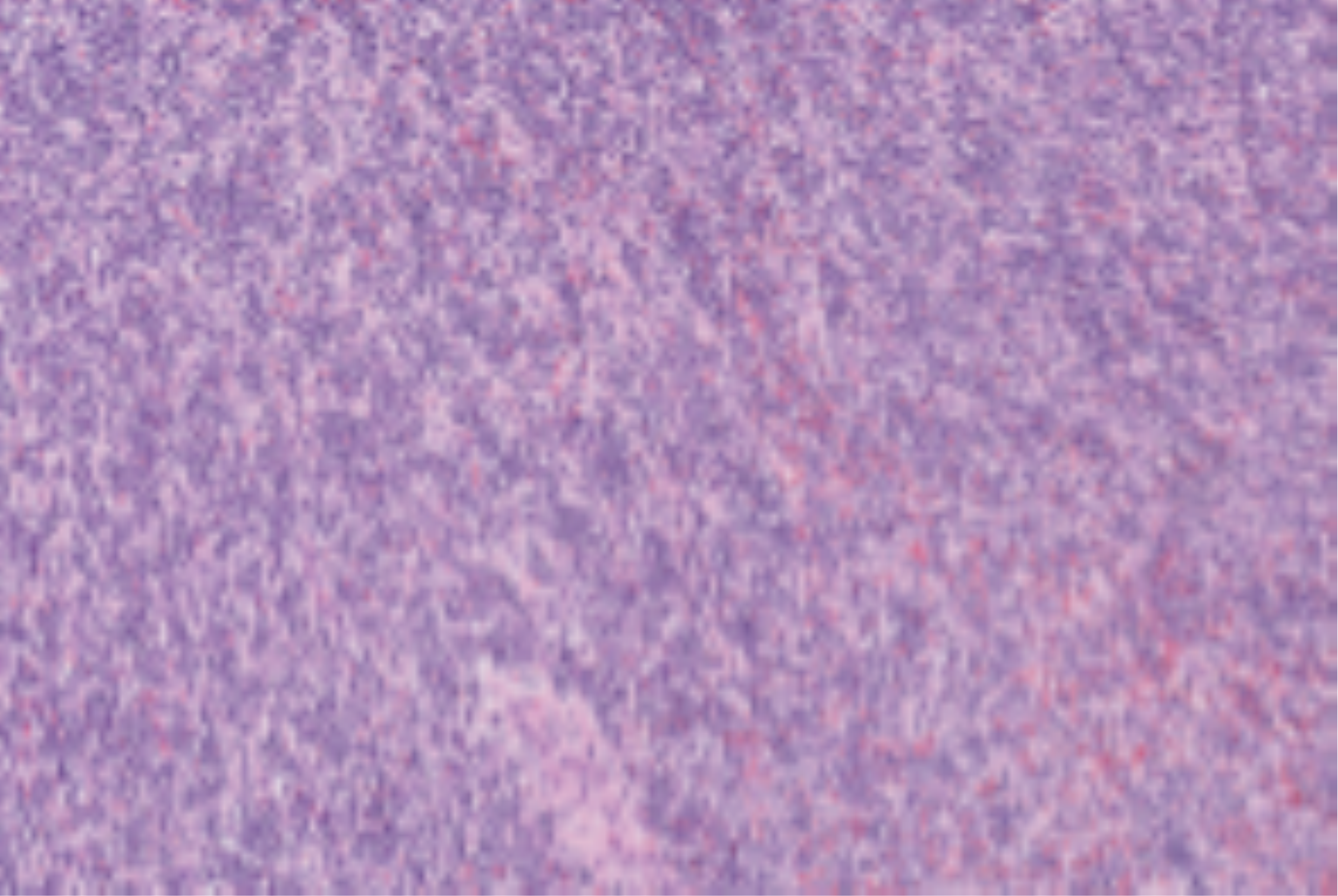}
  \subcaption{Original} \label{fig: 15}
 \end{minipage}
 \begin{minipage}[b]{0.15\hsize}
  \centering
  \includegraphics[width = \hsize]{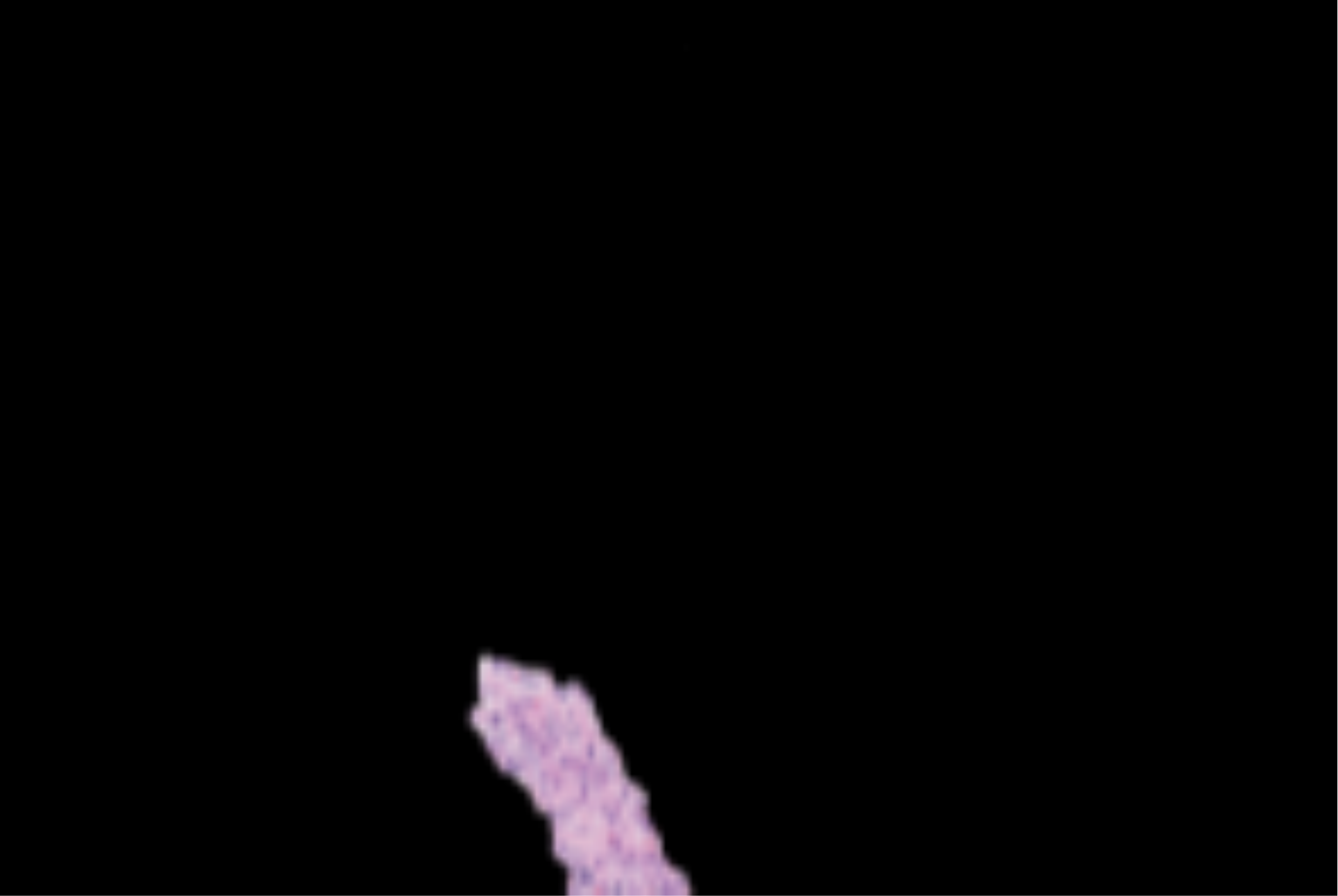}
  \subcaption{Object} \label{fig: 15_obj}
 \end{minipage}
 \begin{minipage}[b]{0.15\hsize}
  \centering
  \includegraphics[width = \hsize]{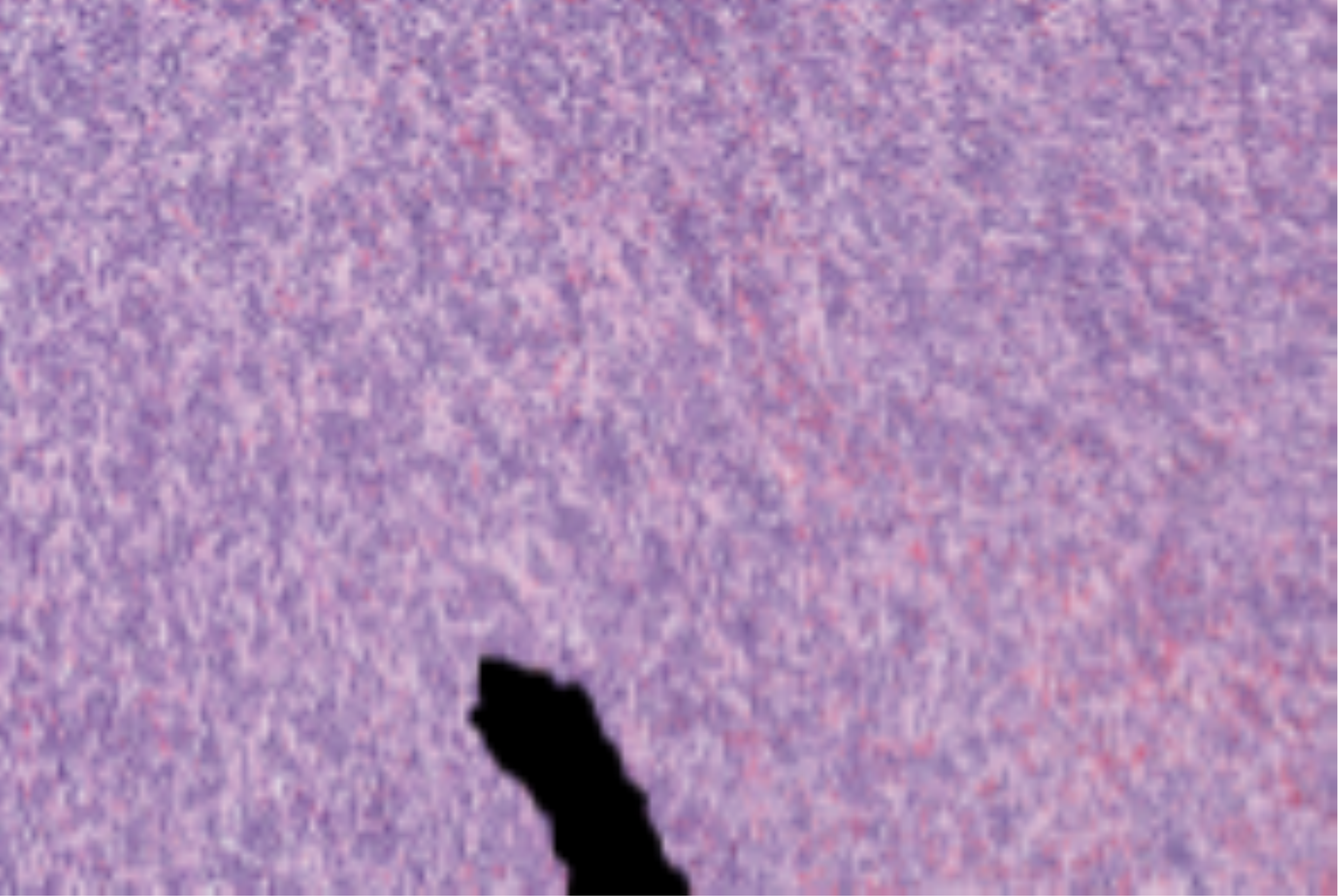}
  \subcaption{Background} \label{fig15_bkg}
 \end{minipage}
 \begin{minipage}[b]{0.5\hsize}
 \end{minipage}
% \begin{center}
%  (naive-$p$ = {\bf 0.00} and selective-$p$ = {\bf 0.00})
% \end{center}   
\begin{minipage}[b]{0.03\hsize}  
\end{minipage}
 \begin{minipage}[b]{0.15\hsize}
  \centering
  \includegraphics[width = \hsize]{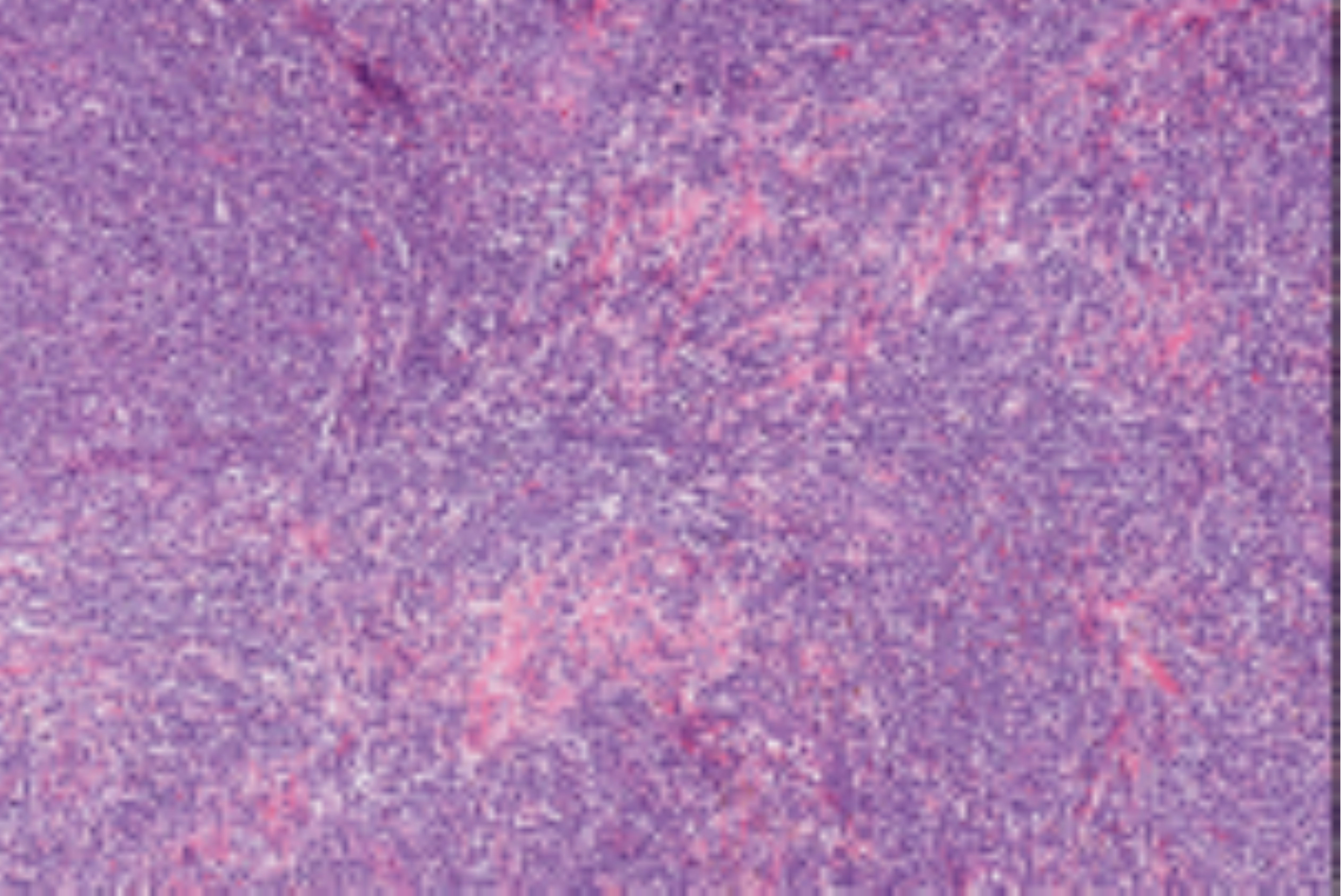}
  \subcaption{Original} \label{fig: 16}
 \end{minipage}
 \begin{minipage}[b]{0.15\hsize}
  \centering
  \includegraphics[width = \hsize]{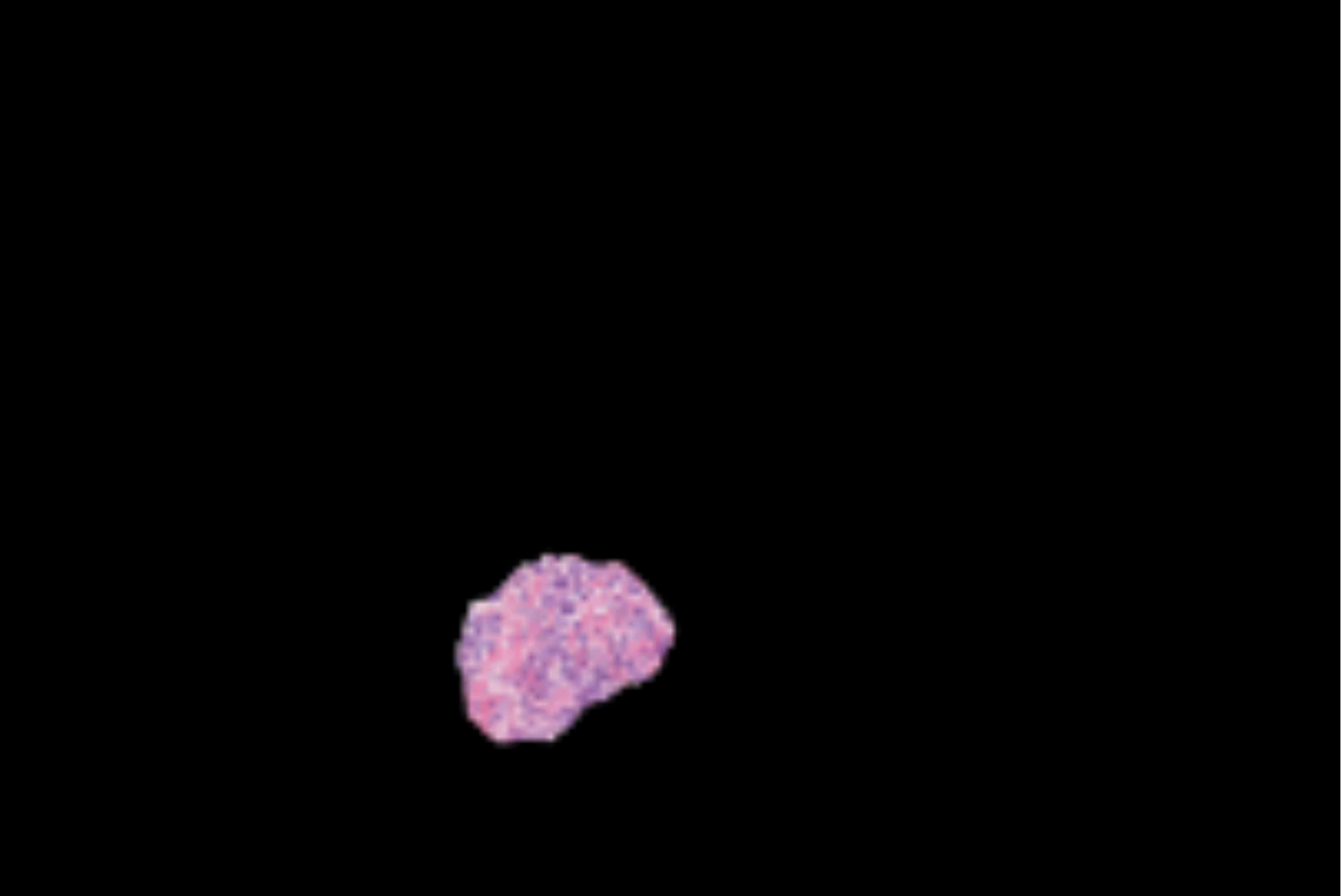}
  \subcaption{Object} \label{fig: 16_obj}
 \end{minipage}
 \begin{minipage}[b]{0.15\hsize}
  \centering
  \includegraphics[width = \hsize]{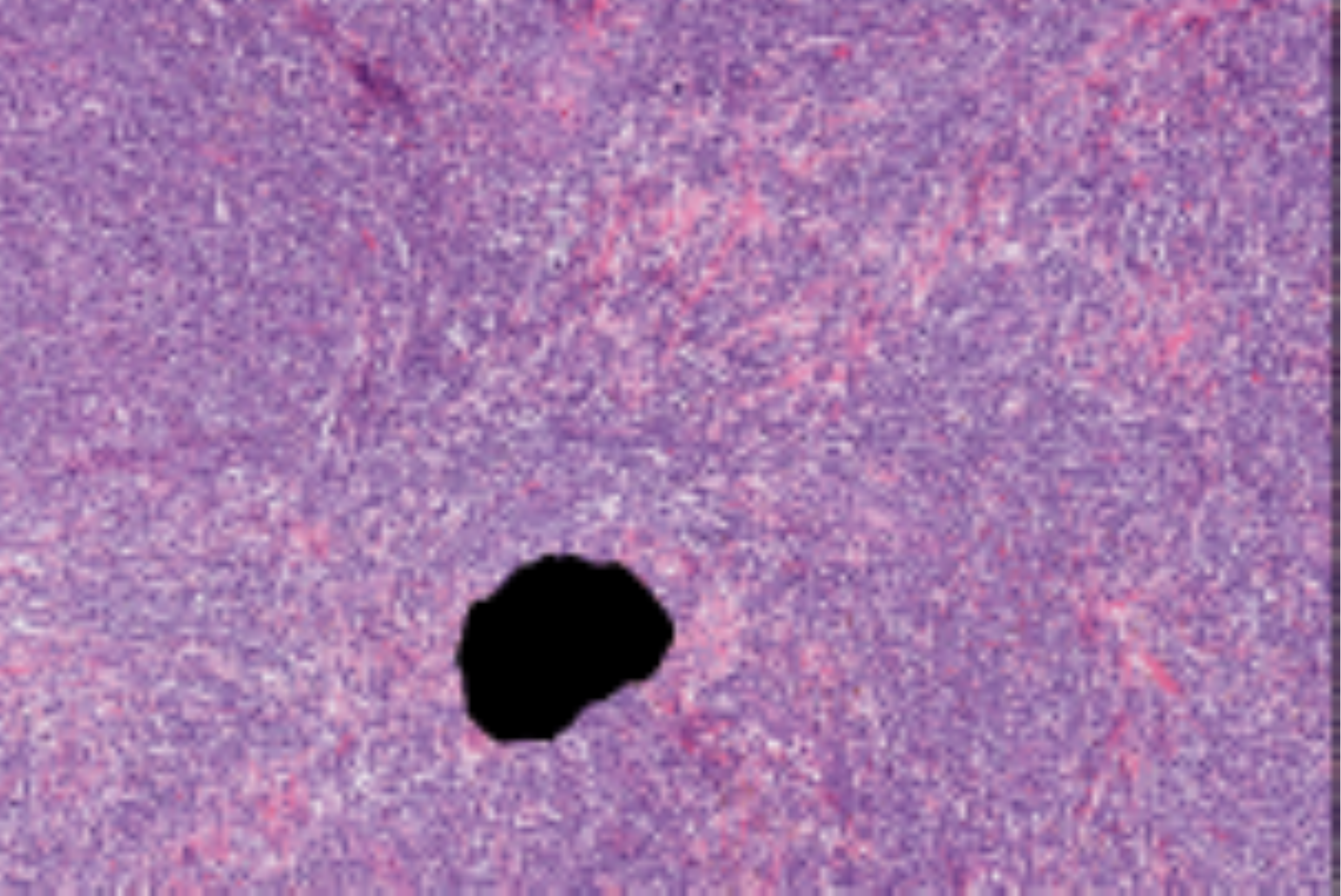}
  \subcaption{Background} \label{fig: 16_bkg}
 \end{minipage}
 \vspace{-3mm}
 \begin{center}
  ~~(naive-$p$ = {\bf 0.00}, selective-$p$ = 0.82)~~~~~~~~~~~~~~~~~~~~~~~~~~~~~~~~(naive-$p$ = {\bf 0.00}, selective-$p$ = 0.94)
 \end{center}
 \caption{
 Segmentation results for pathological images without fibrous regions
  }
 \label{fig: real2}
\end{center}
\end{figure}

\subsection{Segmentation results for CT images with tumor regions}
\begin{figure}[H]
\begin{center}
 \begin{minipage}[b]{0.15\hsize}
  \centering
  \includegraphics[width = \hsize]{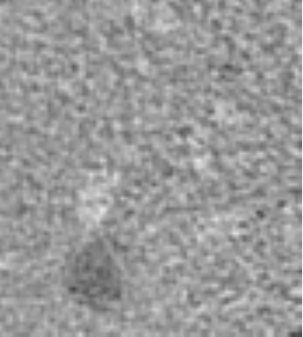}
  \subcaption{Original} \label{fig: 17}
 \end{minipage}
 \begin{minipage}[b]{0.15\hsize}
  \centering
  \includegraphics[width = \hsize]{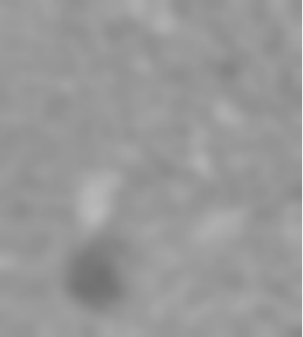}
  \subcaption{Blurred} \label{fig: 17_g}
 \end{minipage}
 \begin{minipage}[b]{0.15\hsize}
  \centering
  \fbox{\includegraphics[width = 0.9\hsize]{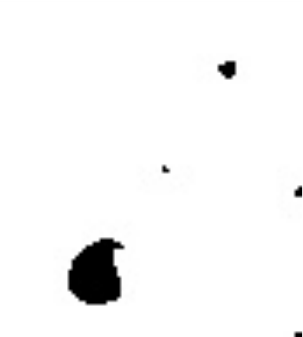}}
  \subcaption{Binarized} \label{fig: 17_b}
 \end{minipage}
 \begin{minipage}[b]{0.5\hsize}
 \end{minipage}
% \begin{center}
%  (naive-$p$ = {\bf 0.00} and selective-$p$ = {\bf 0.00})
% \end{center}   
\begin{minipage}[b]{0.03\hsize}  
\end{minipage}
  \begin{minipage}[b]{0.15\hsize}
  \centering
  \includegraphics[width = \hsize]{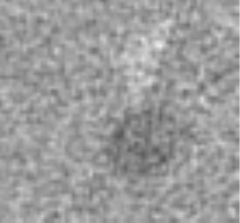}
  \subcaption{Original} \label{fig: 18}
 \end{minipage}
 \begin{minipage}[b]{0.15\hsize}
  \centering
  \includegraphics[width = \hsize]{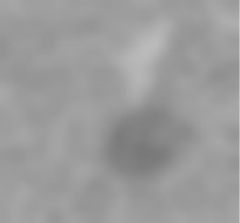}
  \subcaption{Blurred} \label{fig: 18_g}
 \end{minipage}
 \begin{minipage}[b]{0.15\hsize}
  \centering
  \fbox{\includegraphics[width = 0.9\hsize]{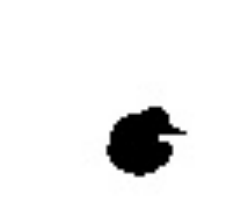}}
  \subcaption{Binarized} \label{fig: 18_b}
 \end{minipage}
 \vspace{-3mm}
 \begin{center}
  ~~(naive-$p$ = {\bf 0.00}, selective-$p$ = 0.30)~~~~~~~~~~~~~~~~~~~~~~~~~~~~~~~~(naive-$p$ = {\bf 0.00}, selective-$p$ = {\bf 0.00})
 \end{center}
 
  \begin{minipage}[b]{0.15\hsize}
  \centering
  \includegraphics[width = \hsize]{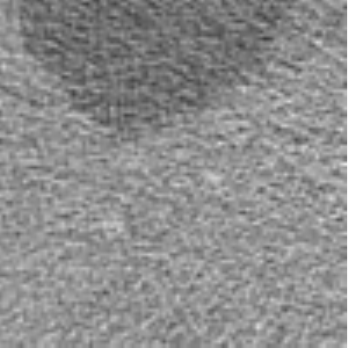}
  \subcaption{Original} \label{fig: 19}
 \end{minipage}
 \begin{minipage}[b]{0.15\hsize}
  \centering
  \includegraphics[width = \hsize]{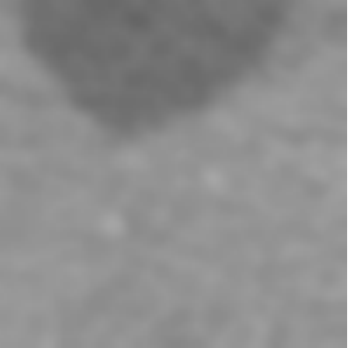}
  \subcaption{Blurred} \label{fig: 19_g}
 \end{minipage}
 \begin{minipage}[b]{0.15\hsize}
  \centering
  \fbox{\includegraphics[width = 0.9\hsize]{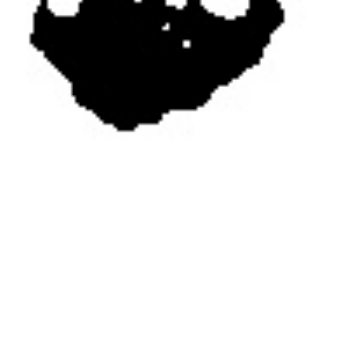}}
  \subcaption{Binarized} \label{fig: 19_b}
 \end{minipage}
 \begin{minipage}[b]{0.5\hsize}
 \end{minipage}
% \begin{center}
%  (naive-$p$ = {\bf 0.00} and selective-$p$ = {\bf 0.00})
% \end{center}   
\begin{minipage}[b]{0.03\hsize}  
\end{minipage}
  \begin{minipage}[b]{0.15\hsize}
  \centering
  \includegraphics[width = \hsize]{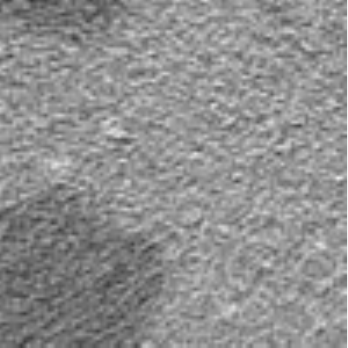}
  \subcaption{Original} \label{fig: 20}
 \end{minipage}
 \begin{minipage}[b]{0.15\hsize}
  \centering
  \includegraphics[width = \hsize]{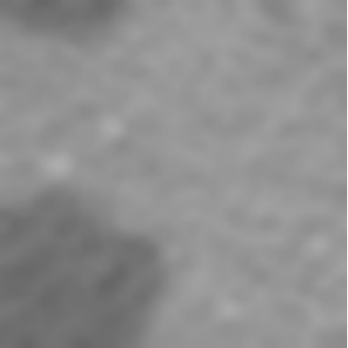}
  \subcaption{Blurred} \label{fig: 20_g}
 \end{minipage}
 \begin{minipage}[b]{0.15\hsize}
  \centering
  \fbox{\includegraphics[width = 0.9\hsize]{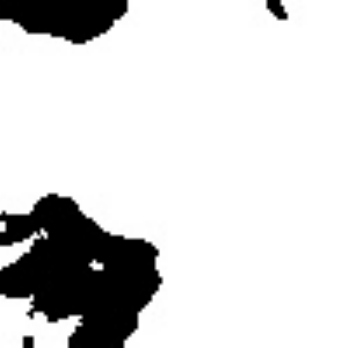}}
  \subcaption{Binarized} \label{fig: 20_b}
 \end{minipage}
 \vspace{-3mm}
 \begin{center}
  ~~(naive-$p$ = {\bf 0.00}, selective-$p$ = {\bf 0.00})~~~~~~~~~~~~~~~~~~~~~~~~~~~~~~~~(naive-$p$ = {\bf 0.00}, selective-$p$ = {\bf 0.00})
 \end{center}
 
  \begin{minipage}[b]{0.15\hsize}
  \centering
  \includegraphics[width = \hsize]{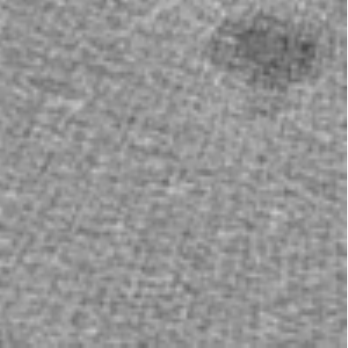}
  \subcaption{Original} \label{fig: 21}
 \end{minipage}
 \begin{minipage}[b]{0.15\hsize}
  \centering
  \includegraphics[width = \hsize]{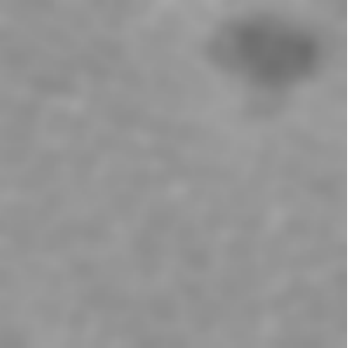}
  \subcaption{Blurred} \label{fig: 21_g}
 \end{minipage}
 \begin{minipage}[b]{0.15\hsize}
  \centering
  \fbox{\includegraphics[width = 0.9\hsize]{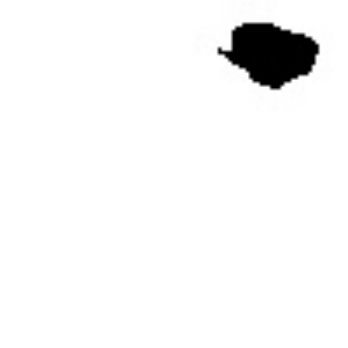}}
  \subcaption{Binarized} \label{fig: 21_b}
 \end{minipage}
 \begin{minipage}[b]{0.5\hsize}
 \end{minipage}
% \begin{center}
%  (naive-$p$ = {\bf 0.00} and selective-$p$ = {\bf 0.00})
% \end{center}   
\begin{minipage}[b]{0.03\hsize}  
\end{minipage}
  \begin{minipage}[b]{0.15\hsize}
  \centering
  \includegraphics[width = \hsize]{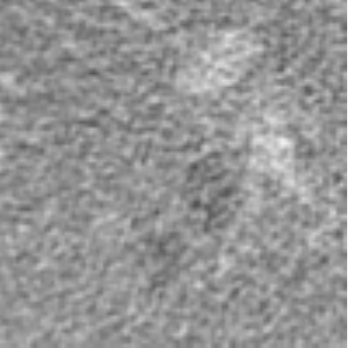}
  \subcaption{Original} \label{fig: 22}
 \end{minipage}
 \begin{minipage}[b]{0.15\hsize}
  \centering
  \includegraphics[width = \hsize]{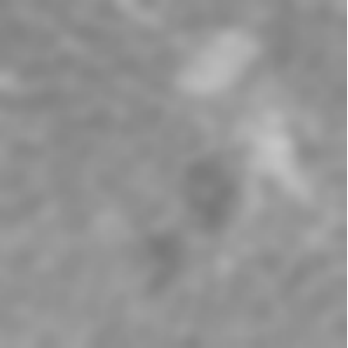}
  \subcaption{Blurred} \label{fig: 22_g}
 \end{minipage}
 \begin{minipage}[b]{0.15\hsize}
  \centering
  \fbox{\includegraphics[width = 0.9\hsize]{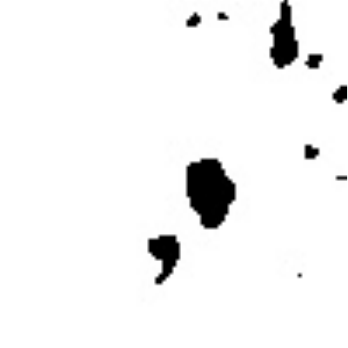}}
  \subcaption{Binarized} \label{fig: 22_b}
 \end{minipage}
 \vspace{-3mm}
 \begin{center}
  ~~(naive-$p$ = {\bf 0.00}, selective-$p$ = {\bf 0.00})~~~~~~~~~~~~~~~~~~~~~~~~~~~~~~~~(naive-$p$ = {\bf 0.00}, selective-$p$ = 0.38)
 \end{center}

 \caption{
 Segmentation results for CT images with tumor regions
 }
 \label{fig: real3}
\end{center}
\end{figure}

\subsection{Segmentation results for CT images without tumor regions}
\begin{figure}[H]
\begin{center}
 \begin{minipage}[b]{0.15\hsize}
  \centering
  \includegraphics[width = \hsize]{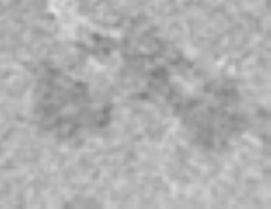}
  \subcaption{Original} \label{fig: 23}
 \end{minipage}
 \begin{minipage}[b]{0.15\hsize}
  \centering
  \includegraphics[width = \hsize]{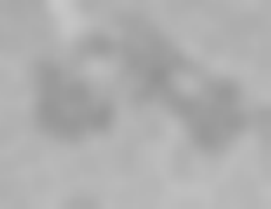}
  \subcaption{Blurred} \label{fig: 23_g}
 \end{minipage}
 \begin{minipage}[b]{0.15\hsize}
  \centering
  \fbox{\includegraphics[width = 0.9\hsize]{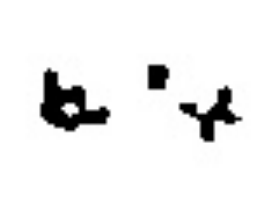}}
  \subcaption{Binarized} \label{fig: 23_b}
 \end{minipage}
 \begin{minipage}[b]{0.5\hsize}
 \end{minipage}
% \begin{center}
%  (naive-$p$ = {\bf 0.00} and selective-$p$ = {\bf 0.00})
% \end{center}   
\begin{minipage}[b]{0.03\hsize}  
\end{minipage}
  \begin{minipage}[b]{0.15\hsize}
  \centering
  \includegraphics[width = \hsize]{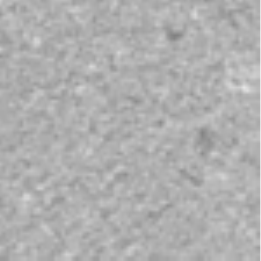}
  \subcaption{Original} \label{fig: 24}
 \end{minipage}
 \begin{minipage}[b]{0.15\hsize}
  \centering
  \includegraphics[width = \hsize]{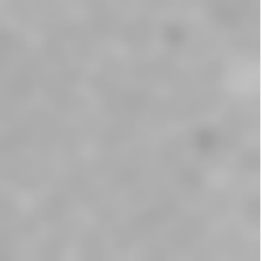}
  \subcaption{Blurred} \label{fig: 24_g}
 \end{minipage}
 \begin{minipage}[b]{0.15\hsize}
  \centering
  \fbox{\includegraphics[width = 0.9\hsize]{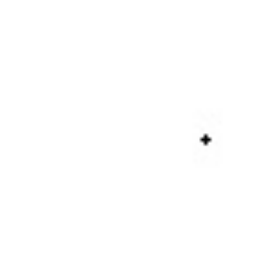}}
  \subcaption{Binarized} \label{fig: 24_b}
 \end{minipage}
 \vspace{-3mm}
 \begin{center}
  ~~(naive-$p$ = {\bf 0.00}, selective-$p$ = 0.96)~~~~~~~~~~~~~~~~~~~~~~~~~~~~~~~~(naive-$p$ = {\bf 0.00}, selective-$p$ = 0.54)
 \end{center}
 
  \begin{minipage}[b]{0.15\hsize}
  \centering
  \includegraphics[width = \hsize]{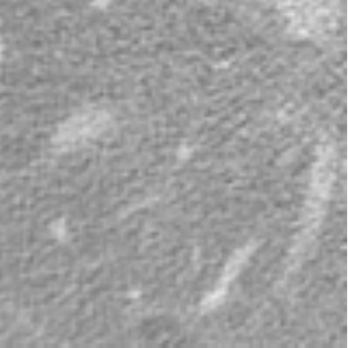}
  \subcaption{Original} \label{fig: 25}
 \end{minipage}
 \begin{minipage}[b]{0.15\hsize}
  \centering
  \includegraphics[width = \hsize]{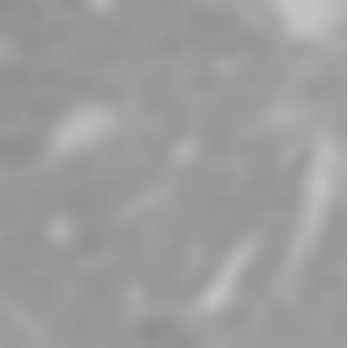}
  \subcaption{Blurred} \label{fig: 25_g}
 \end{minipage}
 \begin{minipage}[b]{0.15\hsize}
  \centering
  \fbox{\includegraphics[width = 0.9\hsize]{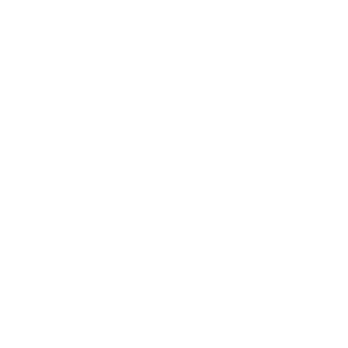}}
  \subcaption{Binarized} \label{fig: 25_b}
 \end{minipage}
 \begin{minipage}[b]{0.5\hsize}
 \end{minipage}
% \begin{center}
%  (naive-$p$ = {\bf 0.00} and selective-$p$ = {\bf 0.00})
% \end{center}   
\begin{minipage}[b]{0.03\hsize}  
\end{minipage}
  \begin{minipage}[b]{0.15\hsize}
  \centering
  \includegraphics[width = \hsize]{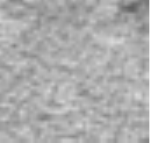}
  \subcaption{Original} \label{fig: 26}
 \end{minipage}
 \begin{minipage}[b]{0.15\hsize}
  \centering
  \includegraphics[width = \hsize]{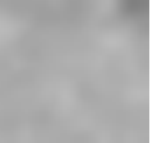}
  \subcaption{Blurred} \label{fig: 26_g}
 \end{minipage}
 \begin{minipage}[b]{0.15\hsize}
  \centering
  \fbox{\includegraphics[width = 0.9\hsize]{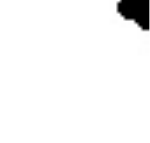}}
  \subcaption{Binarized} \label{fig: 26_b}
 \end{minipage}
 \vspace{-3mm}
 \begin{center}
  ~~(naive-$p$ = 0.06, selective-$p$ = 0.89)~~~~~~~~~~~~~~~~~~~~~~~~~~~~~~~~(naive-$p$ = {\bf 0.00}, selective-$p$ = 0.06)
 \end{center}

 \caption{
 Segmentation results for CT images without tumor regions
 }
 \label{fig: real4}
\end{center}
\end{figure}

\end{document}